\documentclass[11pt,a4paper]{article}
\usepackage{times,latexsym}
\usepackage{url}
\usepackage[T1]{fontenc}

\usepackage[acceptedWithA]{tacl2021v1}

\usepackage{xspace,mfirstuc,tabulary}

\newif\iftaclinstructions
\taclinstructionsfalse 
\iftaclinstructions
\renewcommand{\confidential}{}
\renewcommand{\anonsubtext}{(No author info supplied here, for consistency with
TACL-submission anonymization requirements)}
\newcommand{\instr}
\fi

\iftaclpubformat 

\else

\fi

\usepackage{graphicx}
\usepackage{amsmath}
\usepackage{amssymb}
\usepackage{float}

\usepackage{pgfplots} 
\pgfplotsset{compat=1.18}
\usepackage[caption=false]{subfig} 

\usepackage{booktabs} 
\usepackage{multirow} 
\usepackage{booktabs} 
\usepackage{makecell} 
\usepackage{colortbl} 
\usepackage{xltabular} 

\definecolor{color1}{HTML}{DE7833} 
\definecolor{color2}{HTML}{912C2C} 
\definecolor{color3}{HTML}{F2BB6B} 
\definecolor{color4}{HTML}{C2ABC8} 
\definecolor{color5}{HTML}{329845} 
\definecolor{color6}{HTML}{AED185} 
\definecolor{color7}{HTML}{276C9E} 
\definecolor{color8}{HTML}{A3C9D5} 

\title{Cross-layer Attention Sharing for Pre-trained Large Language Models}

\author{\\
    {\bf Yongyu Mu\textsuperscript{1}\thanks{\xspace\xspace Work was done when Yongyu Mu was interning at Pattern Recognition Center, WeChat AI, Tencent Inc.}, Yuzhang Wu\textsuperscript{1}, Yuchun Fan\textsuperscript{1}, Chenglong Wang\textsuperscript{1}, Hengyu Li\textsuperscript{1}, Jiali Zeng\textsuperscript{2}, Qiaozhi He, } \\
    {\bf Murun Yang\textsuperscript{1}, Fandong Meng\textsuperscript{2}, Jie Zhou\textsuperscript{2}, Tong Xiao\textsuperscript{1}\thanks{\xspace\xspace Corresponding author.} \and Jingbo Zhu\textsuperscript{1}}\\
    \textsuperscript{1}NLP Lab, School of Computer Science and Engineering, Northeastern University, Shenyang, China\\
    \textsuperscript{2}Pattern Recognition Center, WeChat AI, Tencent Inc, China\\
    \texttt{lixiaoyumu9@gmail.com} \\
    \texttt{\{lemonzeng,fandongmeng,withtomzhou\}@tencent.com} \\
    \texttt{\{xiaotong,zhujingbo\}@mail.neu.edu.cn}
}

\date{}

\begin{document}
\maketitle
\begin{abstract}
To enhance the efficiency of the attention mechanism within large language models (LLMs), previous works primarily compress the KV cache or group attention heads, while largely overlooking redundancy between layers. Our comprehensive analyses across various LLMs show that highly similar attention patterns persist within most layers. It's intuitive to reduce the redundancy by sharing attention weights across layers. However, further analysis reveals two challenges: (1) Directly sharing the weight matrix without carefully rearranging the attention heads proves to be ineffective; (2) Shallow layers are vulnerable to small deviations in attention weights. 

Driven by these insights, we introduce \textsc{LiSA}, a lightweight substitute for self-attention in well-trained LLMs. \textsc{LiSA} employs tiny feed-forward networks to align attention heads between adjacent layers and low-rank matrices to approximate differences in layer-wise attention weights. Evaluations encompassing 13 typical benchmarks demonstrate that \textsc{LiSA} maintains high response quality in terms of accuracy and perplexity while reducing redundant attention calculations within $53\%-84\%$ of the total layers. Our implementations of \textsc{LiSA} achieve a $6\times$ compression of $Q$ and $K$ matrices within the attention mechanism, with maximum throughput improvements $19.5\%$, $32.3\%$, and $40.1\%$ for LLaMA3-8B, LLaMA2-7B, and LLaMA2-13B, respectively. Our code is available at \url{https://github.com/takagi97/lisa}.
\end{abstract}

\section{Introduction}
Many transformer models are over-parameterized, leading to significant redundancy across various model components, including attention mechanisms \cite{DBLP:journals/csur/TayDBM23}, feed-forward networks \cite{DBLP:conf/wmt/PiresLAS23}, layers \cite{DBLP:journals/csur/MatsubaraLR23}, and others \cite{DBLP:conf/iclr/LanCGGSS20,DBLP:conf/iclr/JaegleBADIDKZBS22,DBLP:conf/interspeech/HanZZYCQGPW20}. When entering the era of large language models (LLMs), the parameters have extremely expanded. For example, comparing open-source pre-trained models between BERT$_{BASE}$ \cite{DBLP:conf/naacl/DevlinCLT19} and Bloom-176B \cite{DBLP:journals/corr/abs-2211-05100}, the number of parameters has grown nearly $1600\times$, let alone the commercial closed-source ones. Consequently, the redundancy of these models also increases at a gallop.

One of the typical instances is that though the self-attention mechanism consumes unbearably massive memory and computation when tackling long sequences in LLMs, its crucial weight matrix is extremely sparse \cite{DBLP:conf/nips/LiuDLWXXKS23,DBLP:conf/nips/Zhang00CZC0TRBW23,DBLP:conf/iclr/KitaevKL20}, which means substantial computational resources predominantly contribute to marginal effects. Thus, recently, reducing the redundancy within the self-attention of LLMs has become a continually appealing focus. One line of work along this research is reducing the KV cache by cutting down useless tokens \cite{DBLP:conf/nips/LiuDLWXXKS23,DBLP:conf/nips/Zhang00CZC0TRBW23,DBLP:conf/iclr/XiaoTCHL24} or compressing the representation of KV cache \cite{DBLP:journals/corr/abs-2405-04434,DBLP:journals/corr/abs-2403-05527}. Others attempt to prune the attention heads via clustering \cite{DBLP:conf/icml/AgarwalAHE0VPW24} or sparsity predictor \cite{DBLP:conf/icml/LiuWDZY0S0TRC23}.

Indeed, most previous works focus on reducing intra-layer redundancy within LLMs' attention mechanisms. However, inter-layer redundancy—specifically whether it's necessary to calculate attention at every layer—has been overlooked. Efforts contributing to this area are non-trivial, as scaling LLMs leads to more stacked layers, which might sharply increase inter-layer redundancy. In this work, we aim to answer the questions: \textit{To what extent does the redundancy of attention exist across layers in LLMs, and what hinders us from reducing this redundancy?}

We start with a pioneer similarity analysis of each sub-module of the attention mechanism in LLMs. A widespread observation is that the attention weights of most layers are highly similar, especially in adjacent layers of large models. Inspired by the efforts of sharing similar parameters or activation \cite{DBLP:conf/emnlp/AinslieLJZLS23,DBLP:conf/ijcai/XiaoLZ0L19,DBLP:conf/nips/GomezRUG17}, a natural next step is to reuse the attention weight matrices calculated by shallow layers and share them with others. Yet, our analysis shows that this naïve approach inherently faces two main challenges:
\begin{itemize}
    \item Directly sharing the weight matrix without carefully rearranging attention heads is ineffective. Since heads lack positional relationships, directly sharing them is akin to random permutation, adversely impacting similarity. Indeed, most heads can be aligned with a highly similar one in the shared matrix, making it crucial to align them before sharing.
    \item Shallow layers are sensitive to attention weights. Even small deviations can cause performance collapse. Therefore, a remedy for differences is necessary.
\end{itemize}

To address these challenges, we take a further step by presenting a simple, lightweight, and \textbf{L}earnable \textbf{S}haring \textbf{A}ttention mechanism (\textsc{LiSA}) for existing well-trained LLMs. \textsc{LiSA} involves two key components. The first is the \textit{attention heads alignment} module, wherein we align the attention heads in the shared matrix with ones of the current layer to reuse the weights from the most similar heads. The second is the \textit{difference compensation} module, which can approximate the differences of attention weight matrices in two layers, thus preventing performance loss caused by tiny deviations. Experimental results on 13 typical benchmarks show that applying \textsc{LiSA} to more than half of the total layers achieves performance comparable to the original model, even on challenging tasks like mathematical reasoning, while requiring only $0.46\%$ to $1.64\%$ of the parameters to be trained. In terms of efficiency, \textsc{LiSA} significantly reduces redundant attention calculations within $53\%-84\%$ of the total layers via compressing both $Q$ and $K$ matrices by $6\times$. Consequently, \textsc{LiSA} achieves throughput improvements of $19.5\%$ for LLaMA3-8B, $32.3\%$ for LLaMA2-7B, and $40.1\%$ for the larger 13B, with the latter two underscoring \textsc{LiSA}'s scaling benefits.


\section{Background and Related Work}
Most methods that enhance the efficiency of transformer models generally reduce redundancy in parameters, structures, and other aspects. These methods include knowledge distillation \cite{DBLP:conf/emnlp/JiaoYSJCL0L20,DBLP:conf/acl/SunYSLYZ20,DBLP:journals/mta/LinWCS21,DBLP:conf/acl/SunYSLYZ20}, pruning \cite{DBLP:conf/acl/VoitaTMST19,DBLP:conf/iclr/FanGJ20,DBLP:conf/rep4nlp/GordonDA20,DBLP:conf/coling/MaoWWZWZYTB20,DBLP:conf/nips/Sanh0R20}, quantization \cite{DBLP:conf/aaai/ShenDYMYGMK20,DBLP:conf/nips/DettmersLBZ22,DBLP:conf/icml/KimGYMK21}, neural architecture search \cite{DBLP:conf/acl/WangWLCZGH20,DBLP:conf/kdd/Xu0LS0QL21,DBLP:conf/nips/XuMLDWZAG22}, and hardware-aware optimization \cite{DBLP:conf/nips/DaoFERR22,DBLP:conf/iclr/Dao24,DBLP:conf/hpca/HamJKOPSPLPLJ20,DBLP:journals/tvlsi/FangZW22}. In this work, we focus on the redundancy within the attention mechanism. We first review the efficient attention methods used in previous transformer models and then summarize those specifically designed for LLMs.

\begin{figure*}
\centering
\includegraphics[width=0.9\textwidth]{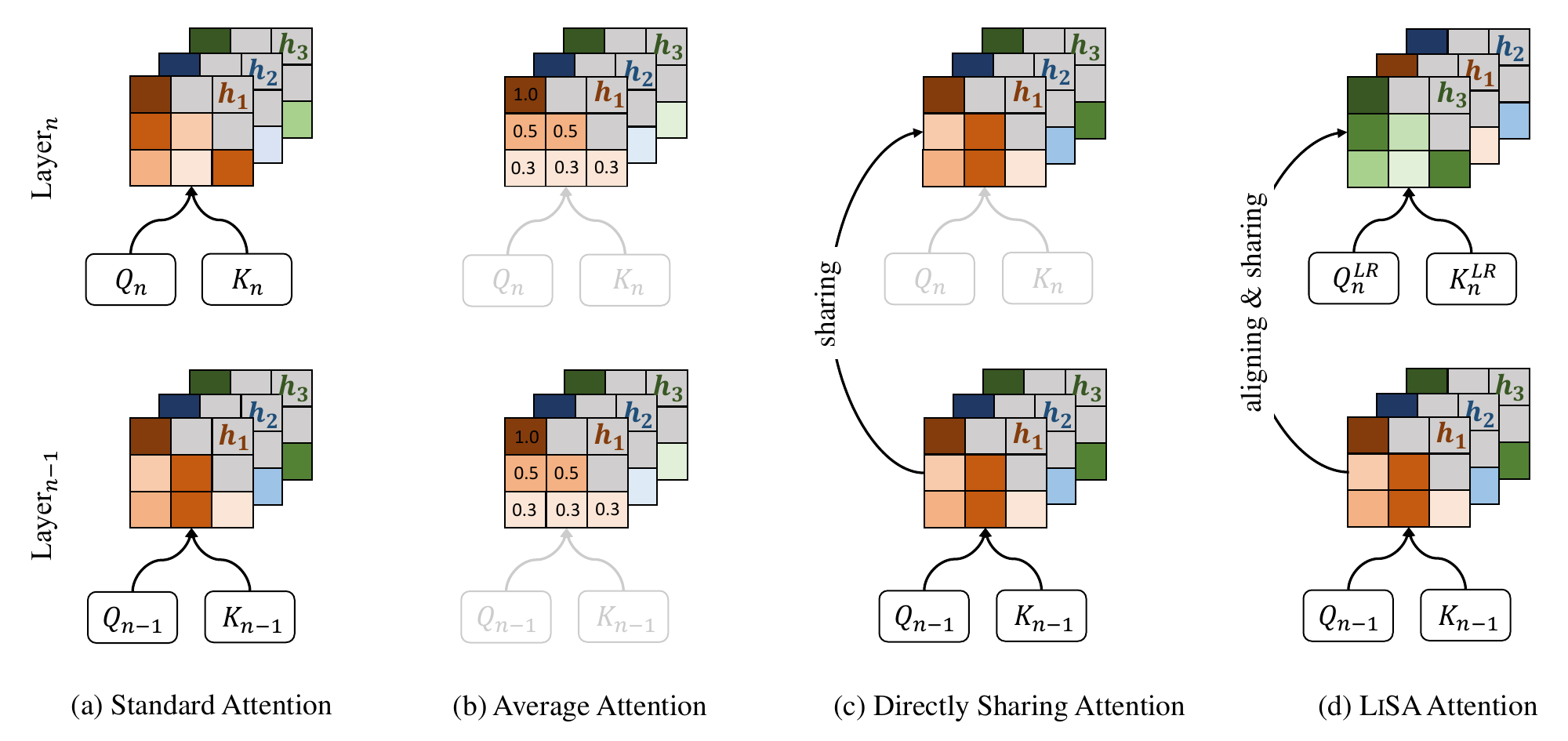}
\caption{Comparison of different attention models. Layer$_n$ stands for a Transformer layer while $h_1$, $h_2$, and $h_3$ represent three attention heads. \textit{Standard attention} individually calculates attention scores at each layer by employing $Q_n$ and $K_n$ matrices. \textit{Average attention} assigns uniform weights across all token positions, thus eliminating $Q$ and $K$ matrices. \textit{Directly sharing attention} reuses the raw weight matrix from the front layer but overlooks varied head weights across different layers. Our method, \textsc{LiSA} attention, not only aligns attention heads but also compensates for layer-wise weight differences leveraging low-rank $Q_n^{LR}$ and $K_n^{LR}$ matrices, thus maximally preserving the original performance while introducing only a few additional training parameters.}
\label{fig:various_attentions}
\end{figure*}

\subsection{Standard Transformer Models}
Let $H \in \mathbb{R}^{l \times d}$ represent the hidden state, where $l$ is the sequence length and $d$ is the dimension of the hidden states. The scaled dot-product multi-head attention (MHA), utilizing $h$ attention heads in $d_k$ dimensions, is defined as follows:

{\footnotesize
\begin{align}
& \text{MHA}(H) = \mathrm{Concat}(P_1HW^V_1, \dots, P_hHW^V_h)W^O \\
& \text{where } P_i = \mathrm{Softmax} \underbrace{\left[\frac{HW^Q_i(HW^K_i)^T} {\sqrt{d_k}}\right]}_{A}
\end{align}}

\noindent where $P_i \in \mathbb{R}^{l \times l}$ is the attention weight matrix, $A$ is the intermediate result before $\text{Softmax}(\cdot)$, and three linear projections $W^Q_i, W^K_i, W^V_i \in \mathbb{R}^{d \times hd_k}$ process the representations into $Q$, $K$, and $V$ matrices. Finally, the output linear projection $W^O \in \mathbb{R}^{hd_k \times d}$ integrates representations from different heads into a single output.

Numerous studies have focused on identifying and reducing redundancy within the components of the attention mechanism, including sparse attention activation $P$ \cite{DBLP:conf/emnlp/LuongPM15,DBLP:conf/interspeech/SperberNNSW18,DBLP:conf/icml/ParmarVUKSKT18,DBLP:conf/emnlp/AinslieOACFPRSW20,DBLP:journals/tacl/RoySVG21,DBLP:conf/iclr/KitaevKL20}, pruning and grouping attention heads \cite{DBLP:conf/nips/MichelLN19,DBLP:conf/acl/VoitaTMST19}, compressing representations $Q,K, \text{and}~V$ \cite{DBLP:conf/iclr/LiuSPGSKS18,DBLP:conf/icml/KatharopoulosV020,DBLP:journals/corr/abs-2006-04768}. In addition to these intra-layer methods, some works aim to reduce layer-wise redundancy by reusing parameters \cite{DBLP:conf/wmt/PiresLAS23} or attention weights $P$ \cite{DBLP:conf/ijcai/XiaoLZ0L19}, and skipping unnecessary layers \cite{DBLP:conf/icpr/Teerapittayanon16}.

The most similar work to ours is SAN \cite{DBLP:conf/ijcai/XiaoLZ0L19}, as it leverages the similarity of attention weights $P$ across multiple layers and directly shares them in neural machine translation (NMT) models, which is shown in Figure \ref{fig:various_attentions} (c). However, the model size and capabilities have significantly evolved from NMT models to LLMs, making a comprehensive analysis of inter-layer redundancy in modern LLMs essential. Additionally, SAN requires re-training models from scratch with a complex training strategy to achieve lossless speedup, limiting its applicability to LLMs. In contrast, our method \textsc{LiSA}, as shown in Figure \ref{fig:various_attentions} (d), can be applied to any existing transformer-based LLMs by only training a few parameters.

\subsection{Large Language Models}

\begin{figure*}
\centering
\includegraphics[width=1.0\textwidth]{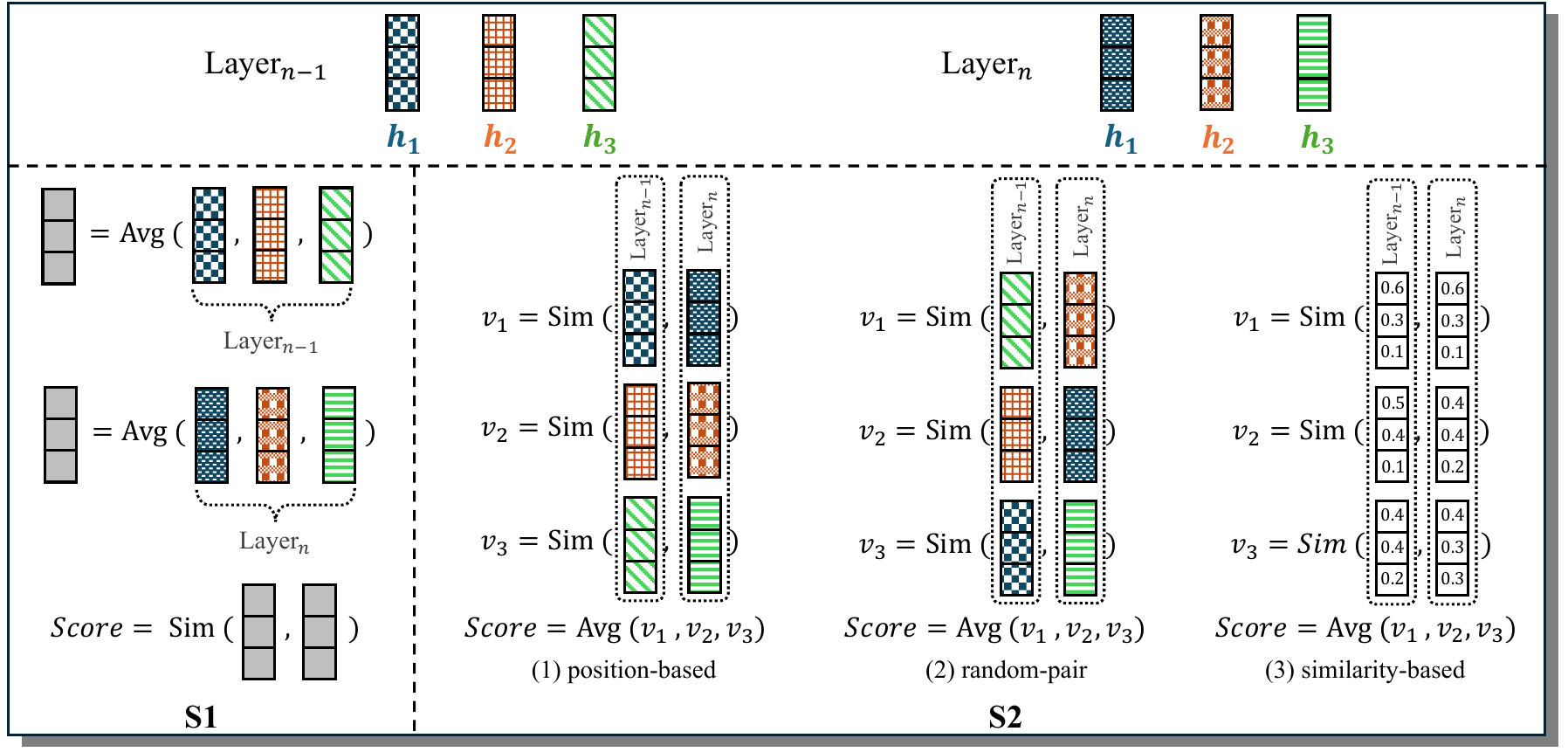}
\vspace{-4mm}
\caption{An illustration of strategies for measuring the similarity of attention weights across different layers. The attention mechanism in each layer is assumed to have three heads, represented by blue, red, and green colors, corresponding to their positions within the attention weight matrices. We propose two similarity calculation settings: \textbf{S1} computes the average attention weights across heads first and then calculates similarity via $\text{Sim}(\cdot)$, e.g., JS divergence; \textbf{S2} calculates pairwise similarity scores between aligned heads individually and then averages the scores. Specifically, three head-alignment strategies are considered: (1) \textit{Position-based} alignment matches heads according to their positional indices; (2) \textit{Random-pair} alignment randomly matches heads from two layers; (3) \textit{Similarity-based} alignment pairs each head with its most similar counterpart in the preceding layer, without enforcing a strict one-to-one correspondence.}
\label{fig:s1_s2}
\end{figure*}

For modern LLMs, KV cache, which stores history representations, has become an essential technique for accelerating inference. It involves two stages: (1) Prefilling, which initializes the KV cache for each layer; (2) Auto-regressive decoding, which updates the KV cache progressively. However, massive memory and computation consumption are still raised in the inference phase of LLMs \cite{DBLP:journals/corr/abs-2205-01068,DBLP:journals/corr/abs-2302-13971,LLaMA3modelcard}.

\subsubsection{Reducing Redundancy Within the Attention Mechanisms of LLMs}

\paragraph{Compressing KV cache.} It is commonly observed that the attention weight matrices are sparse, following a strong power law distribution \cite{DBLP:conf/iclr/KitaevKL20,DBLP:journals/corr/abs-2101-10277,DBLP:conf/iclr/ChoromanskiLDSG21}. This indicates that most tokens memorized in the KV cache are redundant. Some works show that only a few fixed tokens greatly catch attention, thus propose to identify and only store these ``important'' tokens \cite{DBLP:conf/nips/LiuDLWXXKS23,DBLP:conf/nips/Zhang00CZC0TRBW23,DBLP:conf/iclr/XiaoTCHL24,DBLP:conf/iclr/Ge0LZ0024}. Following works continually improve the identification algorithm to reduce performance loss \cite{DBLP:conf/mlsys/AdnanAJNSK24,devoto2024simple,guo2024attention}. Other studies either store the low-rank representation of tokens \cite{DBLP:journals/corr/abs-2405-04434} or quantize KV cache \cite{DBLP:journals/corr/abs-2403-05527}. Recently, \citet{DBLP:journals/corr/abs-2406-02069} and \citet{DBLP:conf/acl/YangHGHZ024} control the KV cache budget according to different layers' behavior. Other attempts implement the KV cache only at certain layers \cite{DBLP:conf/nips/Sun0ZHWMZ0W24,DBLP:conf/nips/LiuLPHHZ24,DBLP:conf/acl/WuT24,DBLP:conf/nips/BrandonMNPR24}.

\paragraph{Pruning attention heads.} Modern LLMs have plenty of attention heads, exacerbating the redundancy. To address this, several approaches have been proposed. Multi-query attention, for instance, shares keys and values among attention heads \cite{DBLP:conf/emnlp/AinslieLJZLS23,DBLP:journals/corr/abs-1911-02150}. Additionally, \citet{DBLP:conf/icml/LiuWDZY0S0TRC23} suggest using a contextual sparsity predictor to identify and dynamically prune unused heads during inference, while \citet{DBLP:conf/icml/AgarwalAHE0VPW24} propose combining heads based on their similar attention weights.

Indeed, the above methods mainly focus on reducing the redundancy within one component of the attention mechanism. However, analyzing the inter-layer redundancy of the attention mechanism in LLMs is overlooked. Furthermore, although several works of early existing and layer skipping reduce the layer-wise redundancy by pruning entire layers \cite{DBLP:conf/iclr/GromovTSGR25,DBLP:conf/ijcai/000100M0SSW25}, the after-pruned models struggle with challenging reasoning tasks \cite{DBLP:conf/acl/MenXZYWL0HC25}, leaving possibility of addressing the inter-layer redundancy within the attention mechanism.

\begin{figure*}
\captionsetup[subfloat]{labelfont=scriptsize, textfont=scriptsize}
\includegraphics[width = 0.24\textwidth]{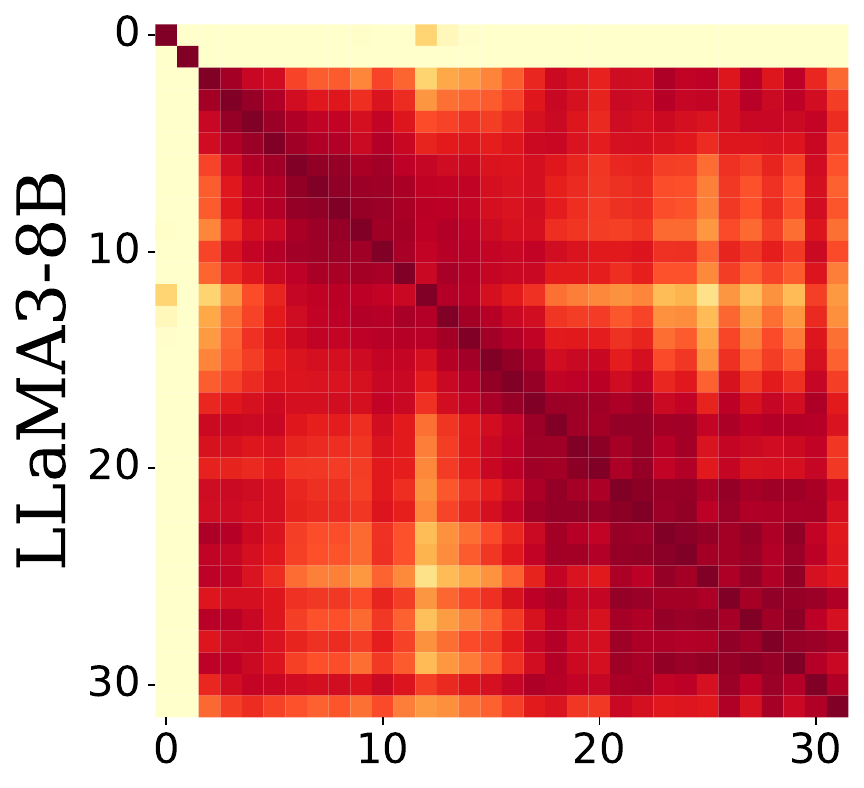}
\hfill
\includegraphics[width = 0.22\textwidth]{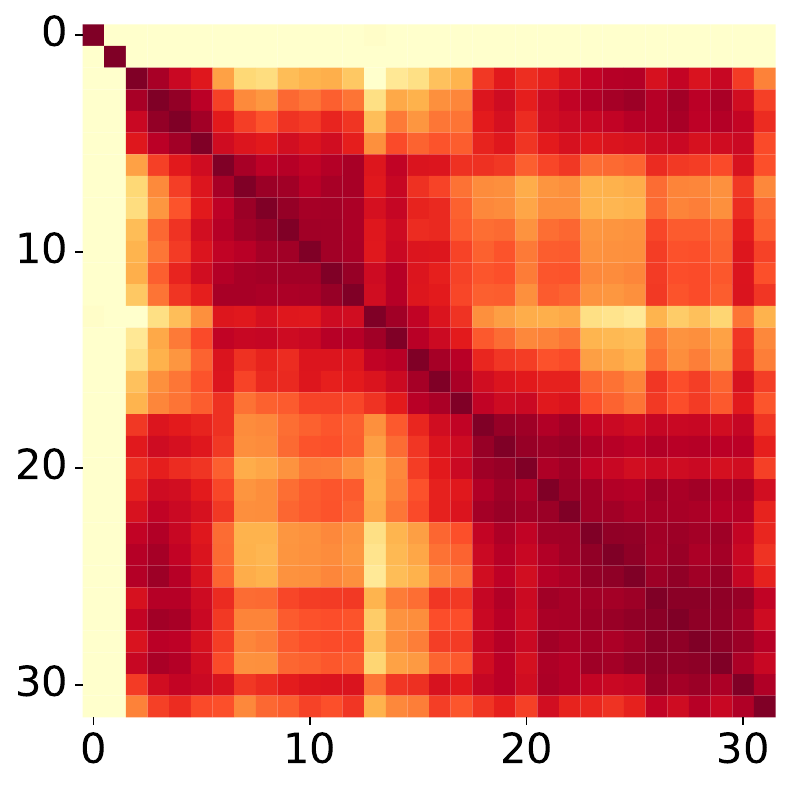}
\hfill
\includegraphics[width = 0.22\textwidth]{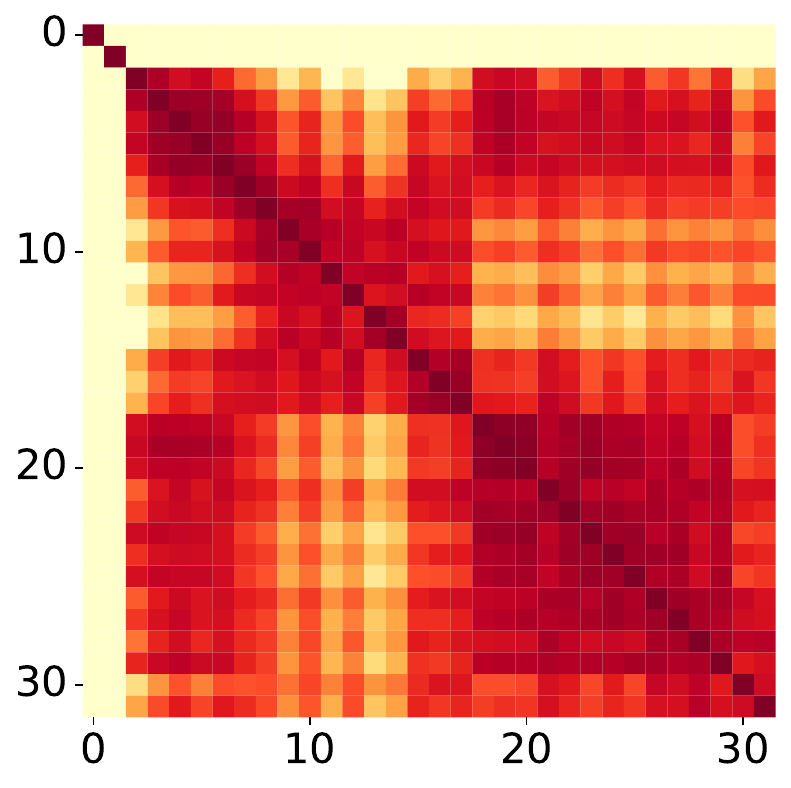}
\hfill
\includegraphics[width = 0.27\textwidth]{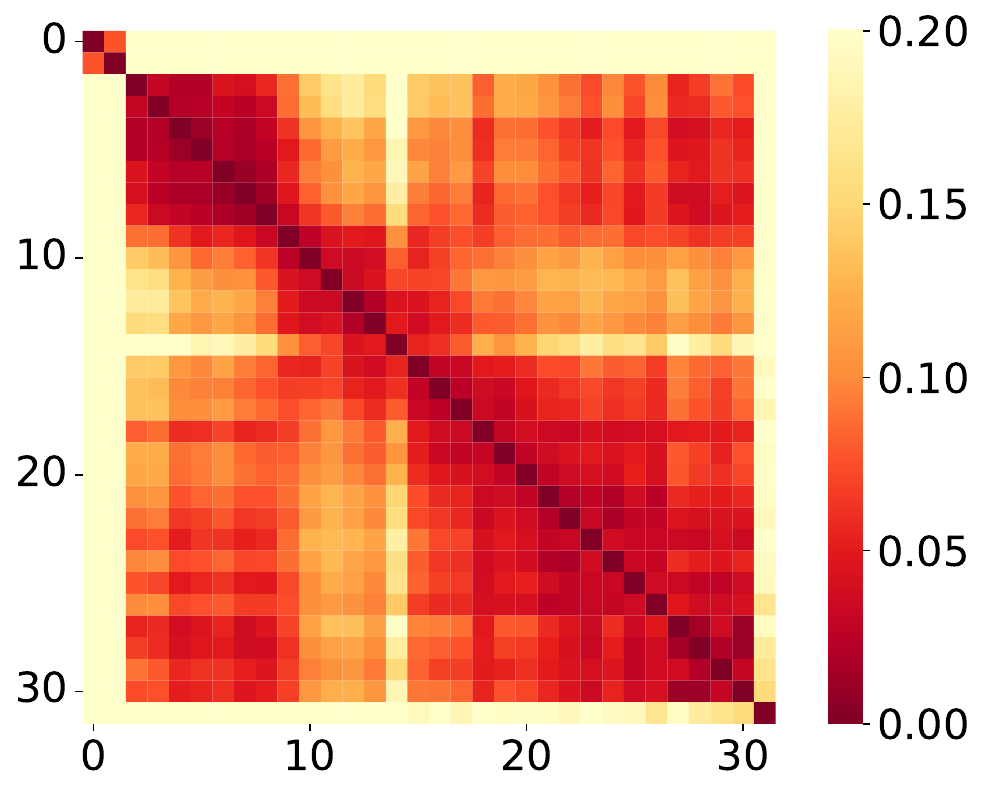}
\subfloat[Physical Commonsense QA]{\includegraphics[width = 0.24\textwidth]{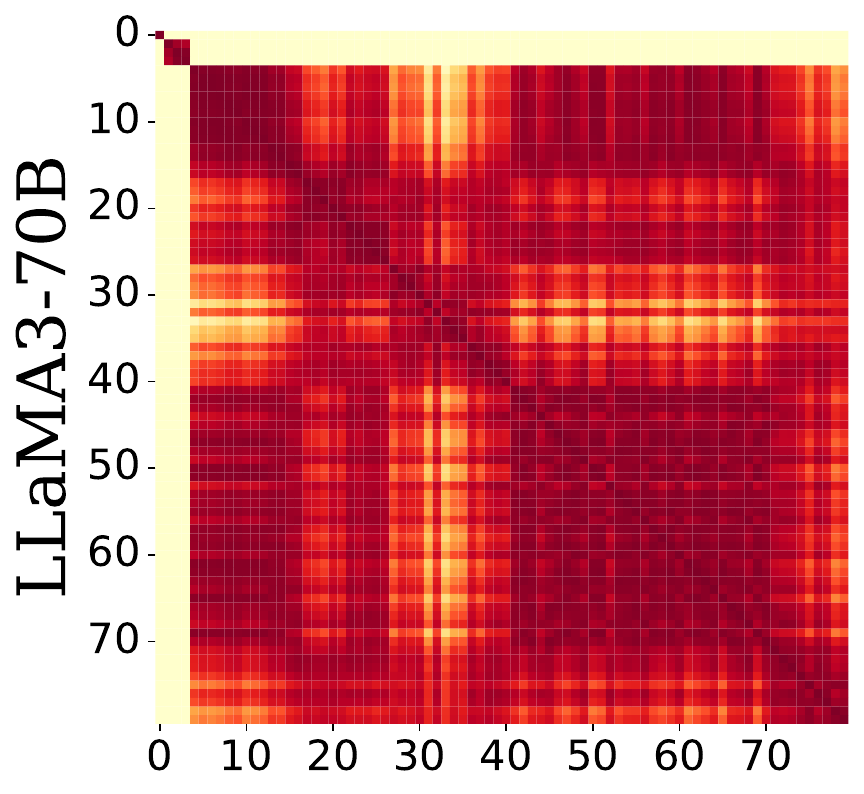}}
\hfill
\subfloat[Short Sentence Translation]{\includegraphics[width = 0.22\textwidth]{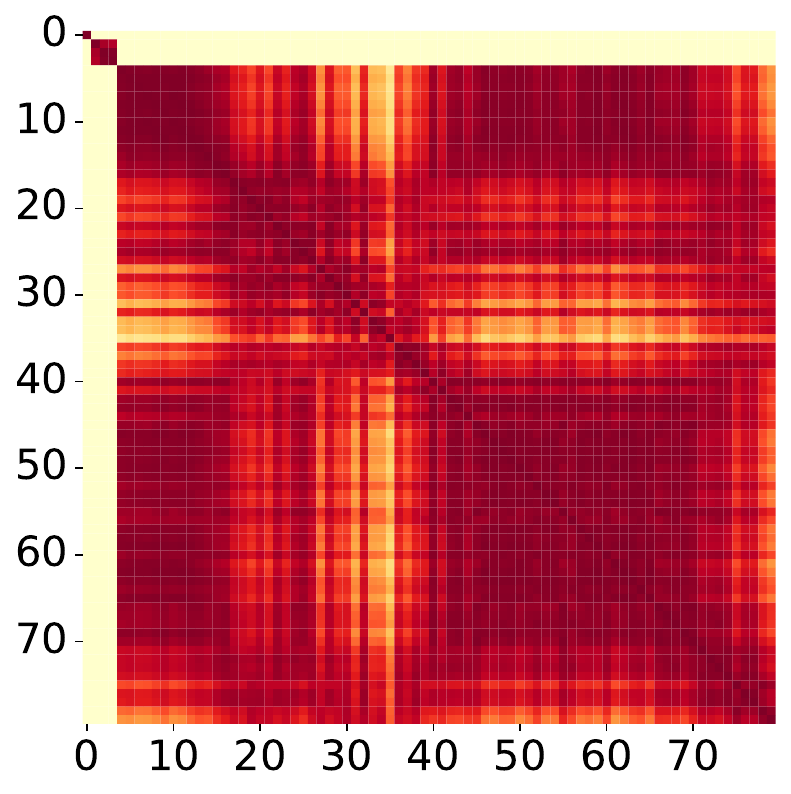}}
\hfill
\subfloat[Coreference Resolution]{\includegraphics[width = 0.22\textwidth]{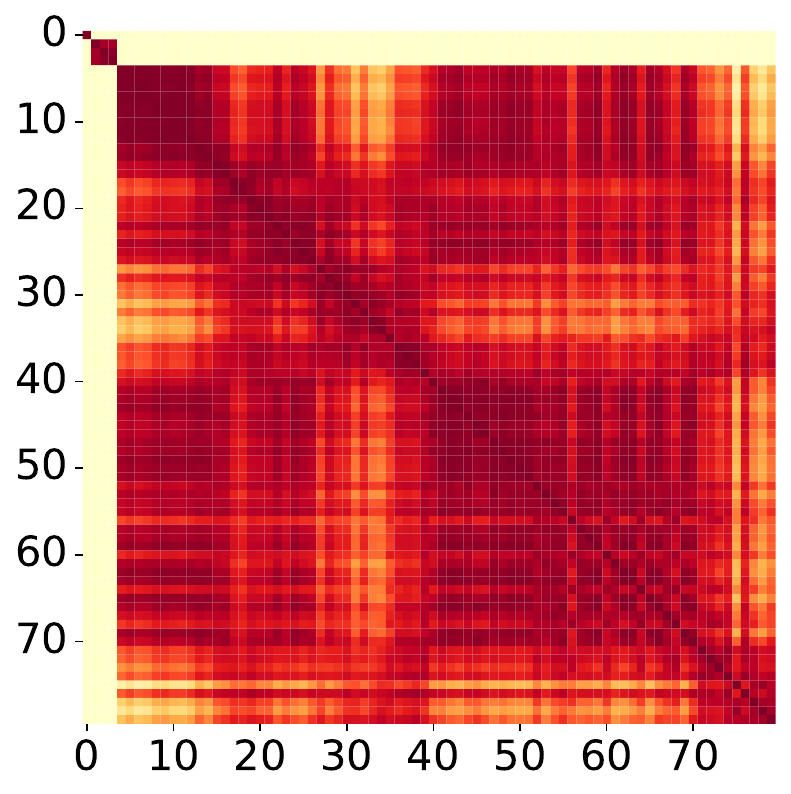}}
\hfill
\subfloat[Mathematical Reasoning]{\includegraphics[width = 0.27\textwidth]{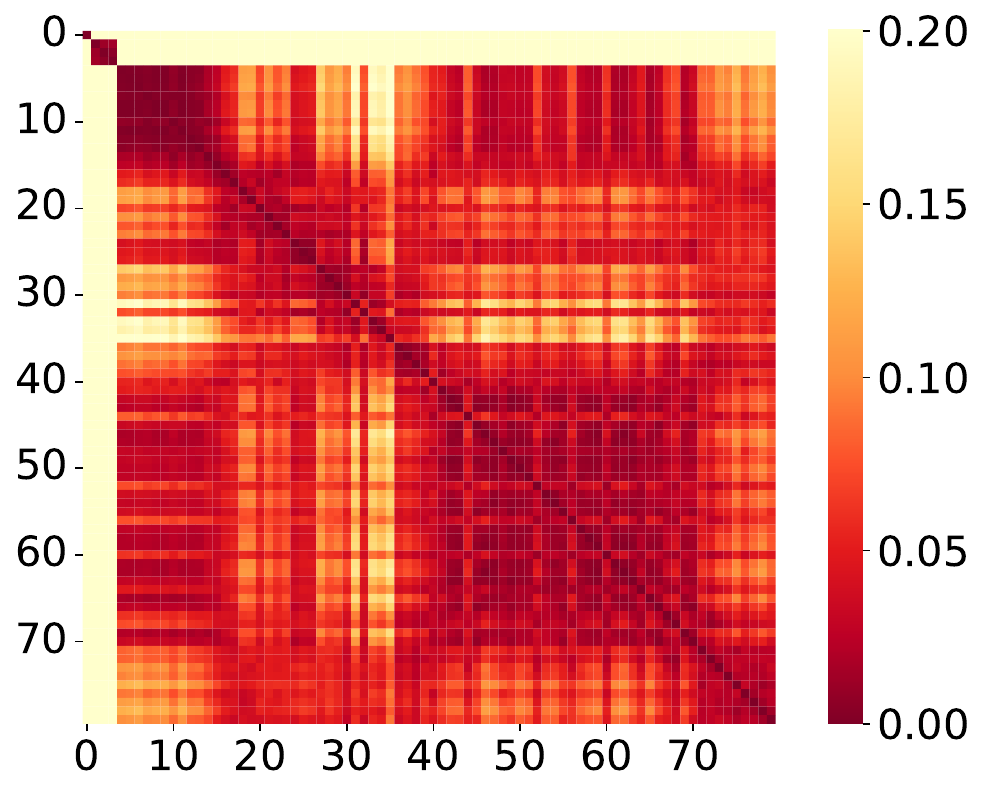}}
\caption{The JS divergence scores for the attention weights between every pair of layers, calculated under setting \textbf{S1}. For all figures, both the horizontal and vertical coordinates stand for layer indices. A deeper red color indicates a lower JS divergence score, corresponding to higher similarity. For instance, the cell located at the third row and fourth column of the top-left figure indicates that the JS divergence between the third and fourth layers of LLaMA3-8B is less than $0.05$. See Figure \ref{app:heatmap2} for results of LLaMA2-7B and Gemma-7B.}
\label{fig:heatmap}
\end{figure*}

\section{Layer-wise Similarity of Attention Weights}
\label{sec:Layer-wise_Similarity_of_Attention_Weights}
Self-attention in transformer models is essentially a procedure that fuses the information from the context to facilitate better understanding \cite{DBLP:journals/corr/abs-2311-17633}. We envision that the attention mechanism of most layers in LLMs may consistently highlight several fixed tokens and assign similar weights to them. To investigate this, \textit{we thoroughly analyze the similarity of attention weights across different layers of LLMs}. Since in MHA the attention mechanism computes separate attention weights for each head, we measure the similarity under the following two settings, as illustrated in Figure \ref{fig:s1_s2}.

\paragraph{S1: Average attention heads first, then compute similarity.} To analyze the similarity of the overall attention scores across layers, we first average the weights of all attention heads within each layer and then compare these weights across all layers.

\paragraph{S2: Align attention heads, compute similarity, and then average the similarity scores.} To measure the similarity while considering the diversity of attention heads, one should align heads from two layers ahead, and then compute the average similarity scores. Specifically, we employ three strategies: (1) Position-based alignment matches attention heads according to their respective positions within the attention weight matrices; for example, the head at dimension 0 in one layer is matched with the head at dimension 0 in another layer. (2) Random-pair alignment matches heads from two layers randomly. (3) Similarity-based alignment pairs each head with the most similar counterpart in another layer without enforcing a strict one-to-one correspondence, which reflects the oracle similarity. We envision that similarity-based aligning heads is necessary for achieving high inter-layer similarity, whereas position-based or random-pair alignment hinders the preservation of similarity.

\subsection{Experiment Settings}
\paragraph{Models and datasets.} We conducted comprehensive experiments on 4 LLMs, specifically LLaMA2-7B \cite{DBLP:journals/corr/abs-2307-09288}, Gemma-7B \cite{DBLP:journals/corr/abs-2403-08295}, LLaMA3-8B \cite{LLaMA3modelcard}, and LLaMA3-70B. Our analysis focused on the behavior of the attention mechanism across various tasks, including physical commonsense QA, short sentence translation, coreference resolution, and mathematical reasoning. The corresponding datasets are PIQA \cite{DBLP:conf/aaai/BiskZLGC20}, WMT \cite{DBLP:conf/wmt/KocmiABBDFFFGGH24}, WEC \cite{DBLP:conf/naacl/EirewCD21}, and GSM8K \cite{DBLP:journals/corr/abs-2110-14168}. Among them, WMT consists of 100 human-selected short sentences, while the others contain 100 randomly sampled instances. The first two tasks involve short inputs where capturing local dependencies is sufficient, while the latter two require handling long-range dependencies\footnote{Each input in PIQA and WMT averages approximately 40 tokens, while WEC contains 120 tokens and GSM8K has 450 tokens per instance.}.

\begin{figure}[!t]
  \centering
  \captionsetup[subfloat]{labelfont=scriptsize, textfont=scriptsize}
  \subfloat[]{
    \includegraphics[width=0.47\columnwidth]{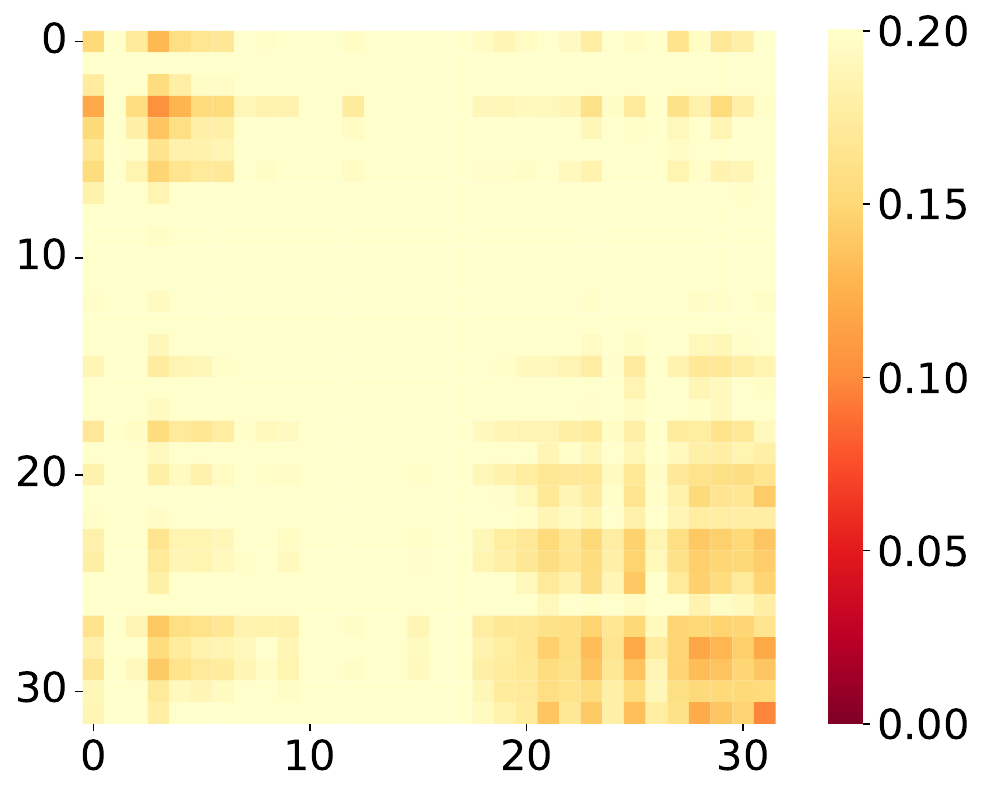}
  }
  \subfloat[]{
    \includegraphics[width=0.47\columnwidth]{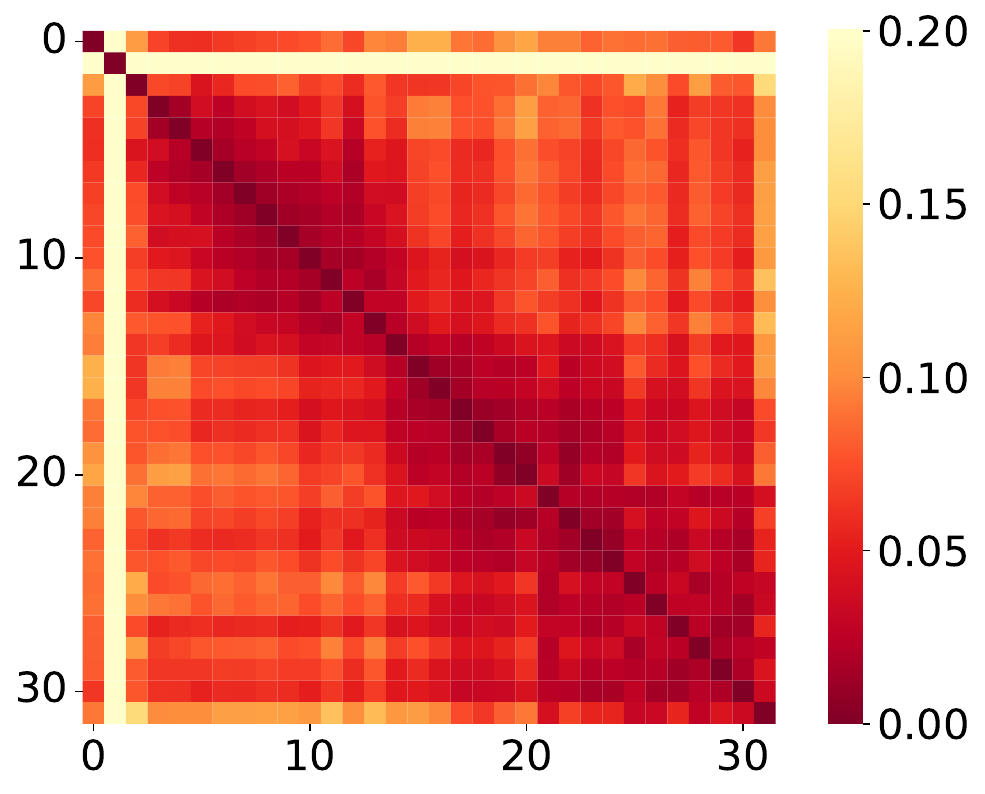}
  }\hfill
  \caption{Figure (a) presents the JS divergence between the attention distributions of two distinct sentences, each with an equal number of tokens, across all pairs of layers in LLaMA3-8B. Figure (b) shows the JS divergence of attention weights for LLaMA3-8B on the PIQA dataset, excluding the first token, which is a special token that receives the majority of the attention.}
  \label{fig:heap_map_rm0_baseline}
\end{figure}

\begin{figure*}
    \centering
    \input{figures/similarity}
    \caption{Figure (a) displays the cosine similarity scores for sub-modules within the attention mechanism across each pair of adjacent layers. Figures (b), (c), and (d) present the average JS divergence of attention weights between adjacent layers under three different alignment strategies in setting \textbf{S2}: position-based, random-pair, and similarity-based, respectively. Lines are added to improve the visual clarity of trends between discrete layers, even though the x-axis represents discrete layer indices.}
    \label{fig:trend_of_similarity}
\end{figure*}

\paragraph{Details of assessing attention similarity.} We primarily focus on evaluating the similarity of attention weights from the last input token to all other tokens. Specifically, for each benchmark, we feed 100 samples into the model, extract and aggregate attention weight distributions across heads (i.e., \textbf{S1} and \textbf{S2} illustrated in Figure \ref{fig:s1_s2}), and compute the Jensen–Shannon (JS) divergence averaged over 100 samples for each pair of layers. For baselines, we report JS scores between the attention weights of two distinct sentences, as shown in Figure \ref{fig:heap_map_rm0_baseline} (a). Moreover, although our main analysis considers the attention weights from the last input token to all other tokens, the experimental results in Table \ref{app_tab:topk_tokens} indicate that the attention distributions from the token generated at step 1024 to all preceding tokens also exhibit strong inter-layer similarity.

\subsection{Results}
The similarity scores calculated under setting \textbf{S1} are shown in Figure \ref{fig:heatmap}. Besides, Figure \ref{fig:trend_of_similarity} (a) records the similarity scores for sub-modules within the attention mechanism. From these results, we get the following observations:

\paragraph{Attention weights are remarkably similar across transformer layers, especially the ones in adjacent layers.} We see, first of all, most JS divergence scores sustained at a degree lower around $0.1$, indicating that most layers prefer a similar attention pattern regardless of models and tasks. Another interesting finding is that the JS divergence score near the diagonal line remains below $0.05$\footnote{See Figure \ref{fig:visualization_of_the_attention_probability_distribution} for the visualization of attention probability distribution pairs with various JS divergence scores.}, demonstrating an extremely similar attention pattern in adjacent layers. This is reasonable because adjacent layers' representations are more similar than non-adjacent ones in deep transformer models \cite{DBLP:conf/blackboxnlp/PhangLB21}. In addition to JS divergence scores, further evidence, such as the top-5 tokens receiving the highest attention weights per layer, presented in Appendix \ref{app:Additional_Evidence_Supporting_Attention_Similarity}, also supports the high similarity of attention weights across layers.

\paragraph{Few layers maintain different attention patterns.} Although most layer pairs in LLaMA3-8B exhibit highly similar attention patterns, some layers compute notably different ones. As shown by the white lines in Figure \ref{fig:heatmap}, attention weights in layers 1–2 and 11–14 differ from those in other layers. However, the red square at the intersection of the white cross indicates that the attention patterns among these adjacent layers remain internally consistent. Notably, the distribution of such distinct layers varies across models. For example, Figure \ref{app:heatmap2} presents results for Gemma-9B, where the first four layers differ from the rest, while the remaining 24 layers maintain highly similar attention patterns.

\paragraph{The similarity of inter-layer attention weights is independent of the task and reflects an inherent property of the model.} Taking LLaMA3-8B as an instance, we observe that the similarity between the attention weights of the first layer and the other layers consistently remains low across different tasks. In contrast, the similarity between the fifth and sixth layers is consistently high. This finding is particularly valuable as it allows for the reuse of attention patterns across specific layers, regardless of the task.

\begin{figure*}
    \centering
    \begin{tikzpicture}[scale=0.5]
\tikzstyle{every node}=[font=\Large]

\node [rectangle,draw=black,minimum height=1.0em,minimum width=33.88em,font=\small,anchor=west,align=center] (final) at (-0.03em,17em){};

\node [anchor=center,color=black!50] at (2em,17.0em){\pgfuseplotmark{*}};
\node [anchor=west,scale=0.9] at (2.5em,17.0em){\scriptsize Poor Perf.};

\node [anchor=center,scale=1.5,color=black!50] at (10.0em,16.9em){\pgfuseplotmark{triangle*}};
\node [anchor=west,scale=0.9] at (10.5em,16.9em){\scriptsize Satisfactory Perf.};

\draw [very thick, color=color2]  (21.0em,17.0em) -- (24.0em,17.0em) coordinate(a);
\node [anchor=west,scale=0.9] at (24em,17.0em){\scriptsize Standard Attention};

\draw [very thick, color=color5]  (35.0em,17.0em) -- (38.0em,17.0em) coordinate(a);
\node [anchor=west,scale=0.9] at (38em,17.0em){\scriptsize Average Attention};

\draw [very thick, color=color7]  (49.0em,17.0em) -- (52.0em,17.0em) coordinate(a);
\node [anchor=west,scale=0.9] at (52em,17.0em){\scriptsize Directly Sharing Attention};

\begin{axis}[
   at={(0,0)},
   ymajorgrids=true,
   xmajorgrids=true,
   grid style={draw=black!20, dashed},
   xtick pos=left,
   xmin=0,xmax=35,
   xlabel={\LARGE Layer Index},
   ylabel={\LARGE Accuracy},
   xlabel style={anchor=center, xshift=0em, yshift=-1.8em},
   title=\LARGE(a) PIQA, every axis title/.style={at={(9em,-8em)}}
   ]
\addplot[line width=2pt,color=color5] plot coordinates {
(3,57.02)
(5,62.19)
(7,70.02)
(9,75.30)
(11,77.42)
(13,77.75)
(15,77.20)
(17,78.29)
(19,76.17)
(21,76.99)
(23,77.20)
(25,78.18)
(27,78.24)
(29,78.84)
(31,78.67)
};\label{pgfplots:piqa_avg} 
\addplot[only marks,mark=*,mark size=3pt,draw=color5,color=color5] plot coordinates {
(3,57.02)
(5,62.19)
(7,70.02)
};\label{pgfplots:piqa_avg_low} 
\addplot[only marks,mark=triangle*,mark size=5pt, color=color5,color=color5] plot coordinates {
(9,75.30)
(11,77.42)
(13,77.75)
(15,77.20)
(17,78.29)
(19,76.17)
(21,76.99)
(23,77.20)
(25,78.18)
(27,78.24)
(29,78.84)
(31,78.67)
};\label{pgfplots:piqa_avg_high} 
\addplot[line width=2pt,color=color7] plot coordinates {
(3,53.86)
(5,80.96)
(7,75.57)
(9,78.18)
(11,79.27)
(13,78.94)
(15,78.84)
(17,79.16)
(19,80.03)
(21,79.87)
(23,79.60)
(25,80.85)
(27,80.20)
(29,79.98)
(31,79.65)
};\label{pgfplots:piqa_share} 
\addplot[only marks,mark=*,mark size=3pt,draw=color7,color=color7] plot coordinates {
(3,53.86)
};\label{pgfplots:piqa_share_low} 
\addplot[only marks,mark=triangle*,mark size=5pt, color=color7,color=color7] plot coordinates {
(5,80.96)
(7,75.57)
(9,78.18)
(11,79.27)
(13,78.94)
(15,78.84)
(17,79.16)
(19,80.03)
(21,79.87)
(23,79.60)
(25,80.85)
(27,80.20)
(29,79.98)
(31,79.65)
};\label{pgfplots:piqa_share_high} 
\addplot[line width=2pt,draw=color2!80,color=color2!80] plot coordinates {
(0,80.52)
(35,80.52)
};\label{pgfplots:piqa_ori}

\end{axis}

\begin{axis}[
   at={(25em,0)},
   ymajorgrids=true,
   xmajorgrids=true,
   grid style={draw=black!20, dashed},
   xtick pos=left,
   xmin=0,xmax=35,
   xlabel={\LARGE Layer Index},
   ylabel={\LARGE Accuracy},
   xlabel style={anchor=center, xshift=0em, yshift=-1.8em},
   title=\LARGE(b) MMLU, every axis title/.style={at={(9em,-8em)}}
   ]
\addplot[line width=2pt,color=color5] plot coordinates {
(3,25.69)
(5,26.24)
(7,26.63)
(9,27.89)
(11,32.82)
(13,37.44)
(15,51.47)
(17,50.73)
(19,64.53)
(21,64.93)
(23,64.79)
(25,64.88)
(27,63.69)
(29,64.03)
(31,65.3)
};\label{pgfplots:mmlu_avg} 
\addplot[only marks,mark=*,mark size=3pt,draw=color5,color=color5] plot coordinates {
(3,25.69)
(5,26.24)
(7,26.63)
(9,27.89)
(11,32.82)
(13,37.44)
(15,51.47)
(17,50.73)
};\label{pgfplots:mmlu_avg_low} 
\addplot[only marks,mark=triangle*,mark size=5pt, color=color5,color=color5] plot coordinates {
(19,64.53)
(21,64.93)
(23,64.79)
(25,64.88)
(27,63.69)
(29,64.03)
(31,65.3)
};\label{pgfplots:mmlu_avg_high} 
\addplot[line width=2pt,color=color7] plot coordinates {
(3,24.68)
(5,51.85)
(7,41.14)
(9,46.05)
(11,50.46)
(13,51.72)
(15,48.77)
(17,47.58)
(19,65.08)
(21,65.44)
(23,65.24)
(25,64.96)
(27,64.92)
(29,65.25)
(31,61.57)
};\label{pgfplots:mmlu_share} 
\addplot[only marks,mark=*,mark size=3pt,draw=color7,color=color7] plot coordinates {
(3,24.68)
(5,51.85)
(7,41.14)
(9,46.05)
(11,50.46)
(13,51.72)
(15,48.77)
(17,47.58)
};\label{pgfplots:mmlu_share_low} 
\addplot[only marks,mark=triangle*,mark size=5pt, color=color7,color=color7] plot coordinates {
(19,65.08)
(21,65.44)
(23,65.24)
(25,64.96)
(27,64.92)
(29,65.25)
(31,61.57)
};\label{pgfplots:mmlu_share_high} 
\addplot[line width=2pt,draw=color2!80,color=color2!80] plot coordinates {
(0,65.24)
(35,65.24)
};\label{pgfplots:mmlu_ori}

\end{axis}

\begin{axis}[
   at={(50em,0em)},
   ymajorgrids=true,
   xmajorgrids=true,
   grid style={draw=black!20, dashed},
   xtick pos=left,
   xmin=0,xmax=35,
   xlabel={\LARGE Layer Index},
   ylabel={\LARGE Exact Match},
   xlabel style={anchor=center, xshift=0em, yshift=-1.8em},
   title=\LARGE(c) GSM8K, every axis title/.style={at={(9em,-8em)}}
   ]

\addplot[line width=2pt,color=color5] plot coordinates {
(3,1.82)
(5,1.36)
(7,2.2)
(9,2.05)
(11,1.52)
(13,2.43)
(15,4.09)
(17,10.39)
(19,23.12)
(21,34.19)
(23,35.86)
(25,40.71)
(27,41.09)
(29,48.9)
(31,48.67)
};\label{pgfplots:gsm8k_avg} 
\addplot[only marks,mark=*,mark size=3pt,draw=color5,color=color5] plot coordinates {
(3,1.82)
(5,1.36)
(7,2.2)
(9,2.05)
(11,1.52)
(13,2.43)
(15,4.09)
(17,10.39)
(19,23.12)
(21,34.19)
(23,35.86)
(25,40.71)
(27,41.09)
};\label{pgfplots:gsm8k_avg_low} 
\addplot[only marks,mark=triangle*,mark size=5pt, color=color5,color=color5] plot coordinates {
(29,48.9)
(31,48.67)
};\label{pgfplots:gsm8k_avg_high} 
\addplot[line width=2pt,color=color7] plot coordinates {
(3,1.44)
(5,24.94)
(7,6.29)
(9,2.65)
(11,7.05)
(13,2.2)
(15,6.67)
(17,18.27)
(19,39.42)
(21,44.96)
(23,48.82)
(25,48.14)
(27,48.22)
(29,47.01)
(31,43.67)
};\label{pgfplots:gsm8k_share} 
\addplot[only marks,mark=*,mark size=3pt,draw=color7,color=color7] plot coordinates {
(3,1.44)
(5,24.94)
(7,6.29)
(9,2.65)
(11,7.05)
(13,2.2)
(15,6.67)
(17,18.27)
(19,39.42)
};\label{pgfplots:gsm8k_share_low} 
\addplot[only marks,mark=triangle*,mark size=5pt, color=color7,color=color7] plot coordinates {
(21,44.96)
(23,48.82)
(25,48.14)
(27,48.22)
(29,47.01)
(31,43.67)
};\label{pgfplots:gsm8k_share_high} 
\addplot[line width=2pt,draw=color2!80,color=color2!80] plot coordinates {
(0,49.73)
(35,49.73)
};\label{pgfplots:gsm8k_ori} 

\end{axis}

\end{tikzpicture}
    \caption{The performance of LLaMA3-8B when applying the average and directly sharing attention strategies to successive pairs of adjacent layers (e.g., layers 3–4, 5–6, 7–8). For instance, the leftmost two points in Figure (a) show the performance when DS and average attention are applied to layers 3 and 4 of the model. The red line indicates the model's original performance, in which standard attention is retained across all layers. When the performance of an attention pattern achieves $90\%$ of the original model's performance, the point is marked with a triangle; otherwise, it is marked with a circle. See Figure \ref{fig:trend_of_sensitivity_of_LLaMA2_7B} for the results of LLaMA2-7B.}
    \label{fig:trend_of_sensitivity}
\end{figure*}

\paragraph{Ablation on the first token.} \citet{DBLP:conf/iclr/XiaoTCHL24} observe that the first token in the sequence often receives disproportionately high attention weights. They argue that the model allocates the excess attention mass to the initial token. We also assess the attention similarity by excluding the weights on the first token, re-softmaxing the remaining weights, and computing the JS divergence scores. The experimental results are shown in Figure \ref{fig:heap_map_rm0_baseline} (b), which is comparable to the first one in the upper left corner of Figure \ref{fig:heatmap}. We can see that the JS divergence scores near the diagonal are around $0.05$, consistently demonstrating significant similarity of the attention weights in adjacent layers.

\paragraph{Only attention weights exhibit cross-layer similarity.} We also measure the similarity of intermediate hidden states in the attention mechanism across each pair of adjacent layers by calculating the cosine similarity. Figure \ref{fig:trend_of_similarity} (a) shows that only the similarity of the attention scores $QK^T$ (the blue line) remains consistently close to $1$ across layers, while the cosine similarity of other intermediate hidden states stays around $0$. This indicates that, while most transformer layers maintain similar attention patterns, they still serve distinct functions, as their $Q$, $K$, and $V$ matrices capture different features. This reflects that these models learn implicit attention patterns across layers while maintaining distinct representations within each layer.

We further analyze the similarity of attention weight while considering the diversity of attention heads, i.e., calculating similarity scores under setting \textbf{S2}. Experimental results of LLaMA3-8B on GSM8K are shown in Figure \ref{fig:trend_of_similarity} (b), (c), and (d).

\paragraph{Similarity score falls when attention heads are matched based on positions.} As shown in Figure \ref{fig:trend_of_similarity} (b), the mean values of JS divergence are around $0.2$, indicating that an attention head in the current layer is not always similar to the one at the same position in the shared attention matrix. We attribute this to the fact that the parameters do not have an inherent positional relationship in neural networks. Thus, position-based alignment is equivalent to random-pair alignment, which is demonstrated by the similar results between Figure \ref{fig:trend_of_similarity} (b) and (c).

\paragraph{Aligning with the most similar head recovers the similarity.} We further measure the oracle similarity by aligning the most similar head for the one in the current layer and calculating the average similarity. From Figure \ref{fig:trend_of_similarity} (d), we see the similarity scores remain below $0.1$ in most layers, which indicates that most attention heads can be aligned with a highly similar one in other layers. It also implies that directly utilizing the shared attention weight matrix might be sub-optimal, and it is crucial to align attention heads beforehand.

\section{Sensitivity to Attention Weights}
\label{sec:Sensitivity to Attention Weights}
Although the attention weights across different LLM layers are highly similar, they are not identical, and sharing them still introduces minor deviations in the model. Previous studies have shown that even small distortions in parameters or embeddings during the inference process of LLMs can lead to notable performance degradation \cite{DBLP:journals/corr/abs-2503-23924,DBLP:journals/corr/abs-2310-14928}. Therefore, the next step is to \textit{analyze how deviations in the attention weights affect performance}.

Here, we select two corrupted attention patterns to simulate deviations from the standard attention weights. The first pattern is the attention weight matrix of the front layer without alignment, i.e., directly sharing weights, as depicted in Figure \ref{fig:various_attentions} (c). Moreover, inspired by AAN \cite{DBLP:conf/acl/XiongZS18}, the second pattern assigns a uniform attention score across all token positions, i.e., the average weights $\frac{1}{l}$, illustrated in Figure \ref{fig:various_attentions} (b).

Subsequently, to assess sensitivity, we apply the two aforementioned patterns to successive pairs of adjacent layers (e.g., layers 3–4, 5–6, 7–8). Under this setup, a significant performance drop from the original model indicates that these layers are particularly sensitive to deviations in attention weights. The findings guide the design of the attention-sharing configurations detailed in Section \ref{sec:Methodology}.

\subsection{Results}
We conducted experiments on three datasets, including PIQA, MMLU \cite{DBLP:conf/iclr/HendrycksBBZMSS21}, and GSM8K. From Figures \ref{fig:trend_of_sensitivity} and \ref{fig:trend_of_sensitivity_of_LLaMA2_7B}, we draw the following conclusions.


\paragraph{Shallow layers are sensitive to the attention score while deep layers are not.} We can see that, in shallow layers, relatively small deviations in attention weights, like sharing attention weights, are more likely to cause a performance collapse. On the contrary, even significant changes happening in deep layers, like averaging attention weights, influence the performance inconspicuously. It indicates that the small deviations contain specific features unique to each layer, which are necessary for sharing attention weights.

\paragraph{The sensitivity of layers is task-dependent.} To retain $90\%$ of the original performance (represented by the points forming a triangular shape), models need to preserve standard attention in different layers depending on the dataset: the shallow layers for PIQA, the first half of the layers for MMLU, and most layers for GSM8K. Upon further analysis of the datasets, we attribute this phenomenon to varying task difficulty. Specifically, PIQA is a relatively simple two-choice benchmark requiring only basic physical commonsense reasoning. MMLU, by contrast, is a more challenging four-choice benchmark comprising 57 knowledge-intensive tasks. Unlike these two, GSM8K demands step-by-step mathematical reasoning to arrive at the final answer, indicating a significantly higher level of complexity. This trend aligns with findings from early-exit studies, which show that the optimal layer for exiting generation correlates with input difficulty—simpler inputs can be processed by early layers, while more complex inputs necessitate deeper-layer computation \cite{DBLP:journals/csur/MatsubaraLR23}.

\begin{figure*}
\centering
\includegraphics[width=0.9\textwidth]{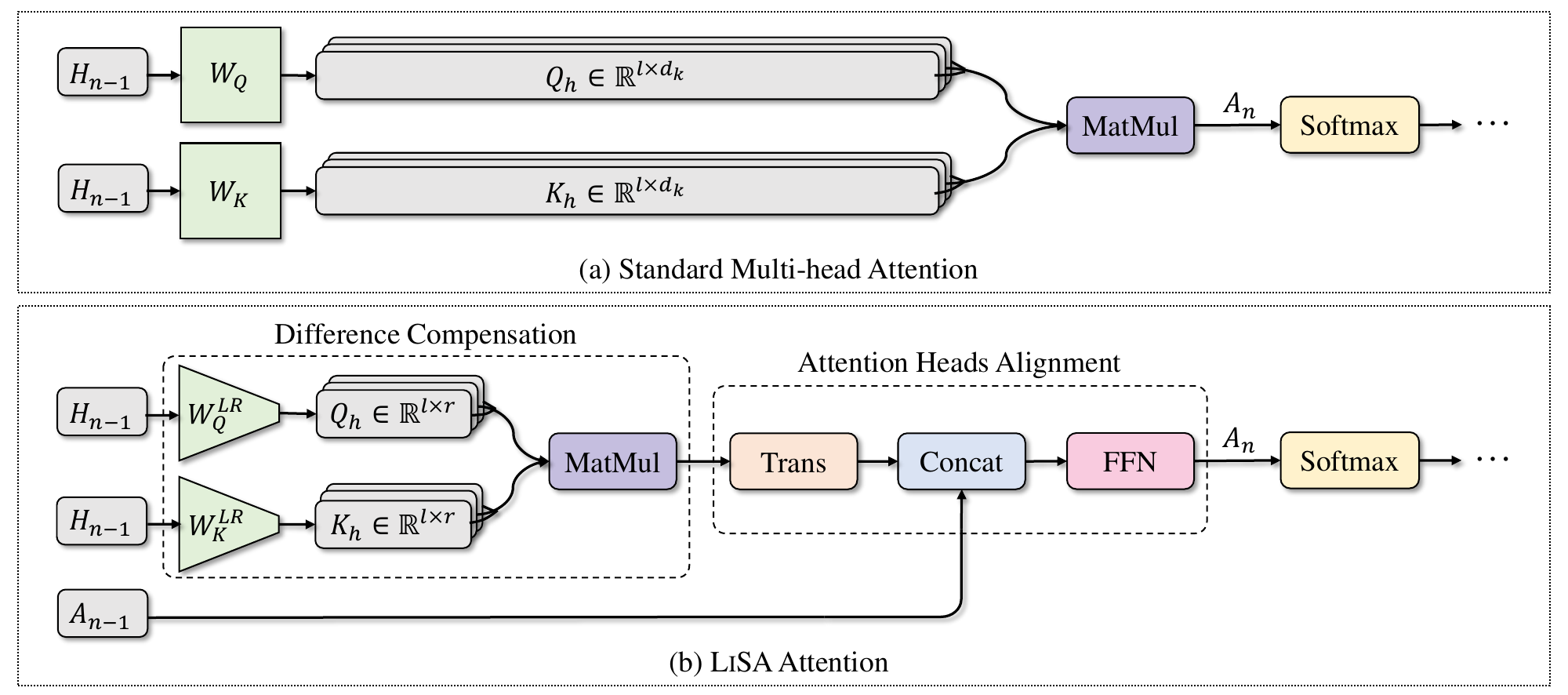}
\caption{A comparison of \textsc{LiSA} with the standard multi-head attention.}
\label{fig:A summary of LiSA}
\end{figure*}

\section{Reducing the Inter-layer Redundancy}
\label{sec:Reducing_the_Inter-layer_Redundancy}
Since the attention scores are similar across transformer layers, it's a natural step to reuse these results across multiple layers, making the inference more efficient. However, this utopia faces two challenges:
\begin{itemize}
    \item \textit{An alignment of attention heads in two layers is crucial for maintaining high similarity.}
    \item \textit{For sensitive layers, minor attention weight deviations cause performance collapses.}
\end{itemize}

We address these challenges by introducing two lightweight components: an \textit{attention heads alignment} module and a \textit{difference compensation} module. The main idea is that we not only align the most similar heads in the shared weights matrix for each head but also compensate for deviations by approximating the difference between the shared weights matrix and the original one. Bring it all together, we present \textsc{LiSA}, which significantly reduces the inter-layer redundancy of attention in well-trained LLMs with minimal loss. Moreover, we theoretically analyze the efficiency of \textsc{LiSA} in Section \ref{sec:Theoretical Analysis of Efficiency}.

\subsection{Methodology}
\label{sec:Methodology}
As shown in Figure \ref{fig:A summary of LiSA}, \textsc{LiSA} reconstructs the calculation steps prior to $\text{Softmax}(\cdot)$ in the self-attention mechanism for a better utilization of the shared attention weights.

\paragraph{Learn to share attention weights.} Let $A$ denote the attention score matrix, i.e., the intermediate output computed prior to applying $\text{Softmax}(\cdot)$. We use $A_n$ to represent the attention score matrix at layer $n$; for example, $A_0$ corresponds to the first layer of an LLM. When the layer $n$ arms with \textsc{LiSA}, the attention heads alignment module accepts a matrix $A_{n-1}$ from the adjacent front layer $n-1$ and produces an aligned one $A^{align}_{n-1}$. Specifically, given a matrix $A \in \mathbb{R}^{h \times l \times l}$, we first transpose it to $A^\top \in \mathbb{R}^{l \times l \times h}$, and then use feed-forward networks (FFNs) to rearrange attention heads and produce the aligned matrix $A^{align}$.

\begin{figure}
\centering
\includegraphics[width=0.5\textwidth]{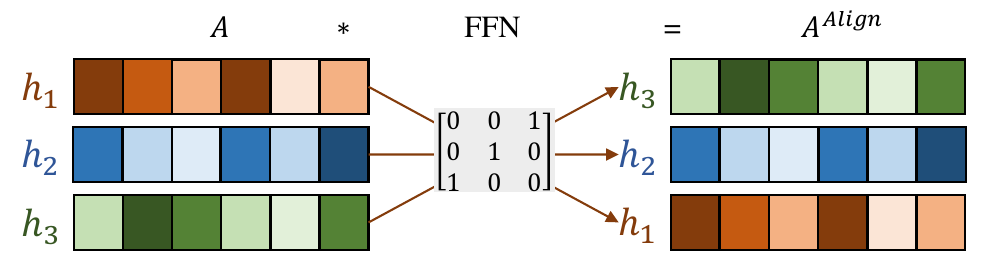}
\caption{An illustration of how FFNs rearrange the attention heads.}
\label{fig:FFN}
\end{figure}

As illustrated in Figure \ref{fig:FFN}, we take an example to explain how FFNs can align attention heads. For simplicity, we start with a one-layer FFN. Supposing that $h=3$ and we need to achieve such alignment: $1\rightarrow3$, $2\rightarrow2$, and $3\rightarrow1$. The shared weight matrix can be aligned by multiplying it with a permutation matrix. Moreover, since the weights of the FFN are consecutive, it also performs as fusing the weights from multiple attention heads.

\paragraph{Low-rank projection closes gaps.} For the difference compensation module, we first use two low-rank linear projections $W^Q_{LR}, W^K_{LR} \in \mathbb{R}^{d \times r}$ as substitutes for $W^Q$ and $W^K$. Given the input hidden state $H$, these linear projections are promoted to capture specific features for the current layer. The resulting $Q_{LR}$ and $K_{LR}$ matrices are then processed by the scaled dot-product mechanism to derive the difference $A_{\Delta} \in \mathbb{R}^{h \times l \times l}$, which is subsequently integrated into the shared attention weight matrix through addition or linear fusion. The whole process is shown as follows:
\begin{align}
& A_{\Delta} = \frac{HW^Q_{LR}(HW^K_{LR})^T} {\sqrt{r}}, \\
& A = \text{Integrate}(A^{align}_{n-1}, A_{\Delta}).
\end{align}
Note that if a tiny dimension $r$ is used, such that $r \ll d$, the representation of $Q$ and $K$ matrices is significantly compressed, thus we can save the memory consumption.

\paragraph{An overview of \textsc{LiSA}.} Complete \textsc{LiSA} is shown in Figure \ref{fig:A summary of LiSA}. To facilitate more precise alignment, we extend the input of the attention heads alignment module by concatenating $A^\top_{n-1} \in \mathbb{R}^{l \times l \times h}$ with $A^\top_{\Delta} \in \mathbb{R}^{l \times l \times h}$. Then the attention heads alignment module fuses two matrices and outputs the final attention score matrix $A_n$. Surprisingly, a super lightweight two-layer or single-layer FFN performs effectively in this module. See Figure \ref{fig:alignment_FFN} for a well-trained FFN.

\begin{figure}
    \centering
    \includegraphics[width=0.48\textwidth]{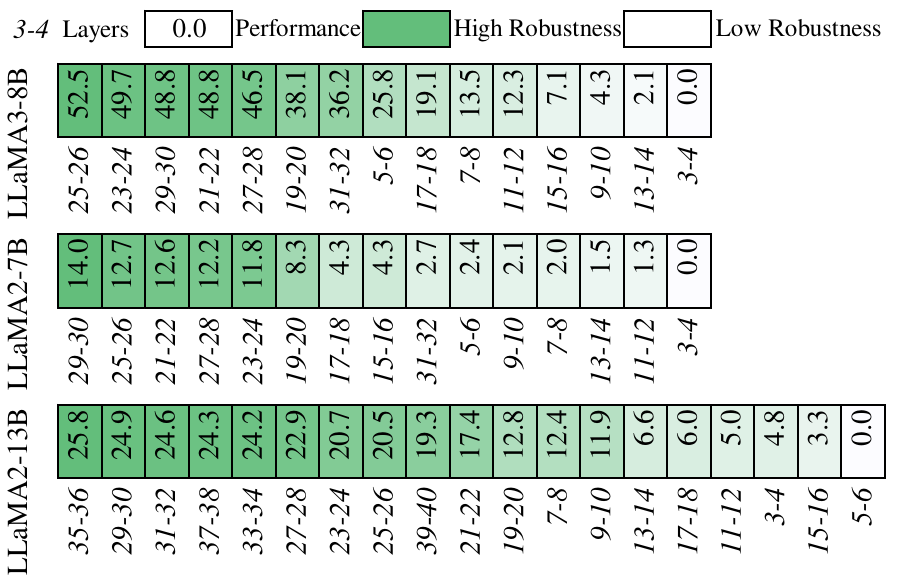}
    \caption{The descending order ranked by each layer's robustness. We evaluated models on the 100 items randomly sampled from the GSM8K training set.}
    \label{fig:priority_robust}
\end{figure}

\paragraph{Selecting layers to implement \textsc{LiSA}.} To optimally preserve the performance, we consider the robustness of a layer to variations in attention weights and the similarity of layer-wise attention scores when implementing \textsc{LiSA}.
\begin{itemize}
    \item Selecting robust layers. Given an LLM, we assess each layer's robustness by evaluating its performance when armed with the same attention-sharing strategy described in Section \ref{sec:Sensitivity to Attention Weights}, focusing on challenging mathematical reasoning tasks. For example, Figure \ref{fig:priority_robust} suggests layers 21-30 of LLaMA3-8B are preferred to implement \textsc{LiSA}.
    \item Excluding the first and last few layers. The attention weights in these layers differ from those in adjacent layers, suggesting that attention sharing may not be appropriate for them. For instance, in the upper one of Figure \ref{fig:heatmap} (d), low similarity values (i.e., yellow blocks) are located in the first two rows and columns, as well as the last row and column.
    \item Implementing \textsc{LiSA} at intervals. For layers beyond the first two categories, we recommend alternating \textsc{LiSA} with standard attention—for instance, applying \textsc{LiSA} for two consecutive layers, followed by a standard attention layer, and repeating this pattern. This periodic application of \textsc{LiSA} prevents continuous sharing of the same attention weights, which could limit performance.
\end{itemize}

\begin{table*}[htbp]
  \centering
  \LARGE
 \resizebox{1.0\linewidth}{!}{
    \begin{tabular}{cclccl}
    \toprule
    \multirow{2}{*}{\textbf{Base Model}} & \textbf{\#Total} & \multirow{2}{*}{\textbf{Model Name}} & \textbf{\#Sharing } & \multirow{2}{*}{\textbf{Proportion}} & \multirow{2}{*}{\textbf{Specific Layers}} \\
     & \textbf{Layers} & & \textbf{Layers} & & \\
    \midrule
    \multirow{7}{*}{LLaMA3-8B} & \multirow{7}{*}{32} & DS (17) & 17 & 53.13\% & \textcolor{color7}{\textit{5,6,17,18,19,20,21,22,23,24,25,26,27,28,29,30,31}} \\
     &  & DS (21) & 21 & 65.63\% & \textcolor{color7}{\textit{4,5,7,8,10,11,13,14,16,17,19,20,22,23,24,25,26,27,28,29,30}} \\
     &  & DS (27) & 27 & 84.38\% & \textcolor{color7}{\textit{4,5,6,7,8,9,10,11,12,13,14,15,16,17,18,19,20,21,22,23,24,25,26,27,28,29,30}} \\
     &  & \textsc{LiSA} (17) & 17 & 53.13\% & \textcolor{color2}{5,6,17,18,19,20,21,22,23,24,25,26,27,28,29,30,31} \\
     &  & \textsc{LiSA}$_{SL}$ (7+10) & 17 & 53.13\% & \textcolor{color2}{5,6,17,18,19,20,21,}\textcolor{color7}{\textit{22,23,24,25,26,27,28,29,30,31}} \\
     &  & \textsc{LiSA} (21) & 21 & 65.63\% &  \textcolor{color2}{4,5,7,8,10,11,13,14,16,17,19,20,22,23,24,25,26,27,28,29,30} \\
     &  & \textsc{LiSA} (27) & 27 & 84.38\% &  \textcolor{color2}{4,5,6,7,8,9,10,11,12,13,14,15,16,17,18,19,20,21,22,23,24,25,26,27,28,29,30} \\
    \midrule
    \multirow{5}{*}{LLaMA2-7B} & \multirow{5}{*}{32} & DS (17) & 17 & 53.13\% & \textcolor{color7}{\textit{5,6,17,18,19,20,21,22,23,24,25,26,27,28,29,30,31}} \\
     &  & DS (21) & 21 & 65.63\% & \textcolor{color7}{\textit{4,5,7,8,10,11,13,14,16,17,19,20,22,23,24,25,26,27,28,29,30}} \\
     &  & \textsc{LiSA} (17) & 17 & 53.13\% &  \textcolor{color2}{5,6,17,18,19,20,21,22,23,24,25,26,27,28,29,30,31} \\
     &  & \textsc{LiSA}$_{SL}$ (7+10) & 17 & 53.13\% &  \textcolor{color2}{5,6,17,18,19,20,21,}\textcolor{color7}{\textit{22,23,24,25,26,27,28,29,30,31}} \\
     &  & \textsc{LiSA} (21) & 21 & 65.63\% &  \textcolor{color2}{4,5,7,8,10,11,13,14,16,17,19,20,22,23,24,25,26,27,28,29,30} \\
     \midrule
    \multirow{5}{*}{LLaMA2-13B} & \multirow{5}{*}{40} & DS (22) & 22 & 55.00\% & \textcolor{color7}{\textit{7,8,19,20,21,22,23,24,25,26,27,28,29,30,31,32,33,34,35,36,37,38}} \\
     &  & DS (26) & 26 & 65.00\% & \textcolor{color7}{\textit{7,8,9,10,18,19,20,21,22,23,24,25,26,27,28,29,30,31,32,33,34,35,36,37,38,39}} \\
     &  & \textsc{LiSA} (22) & 22 & 55.00\% &  \textcolor{color2}{7,8,19,20,21,22,23,24,25,26,27,28,29,30,31,32,33,34,35,36,37,38} \\
     &  & \textsc{LiSA}$_{SL}$ (10+12) & 22 & 55.00\% &  \textcolor{color2}{7,8,19,20,21,22,23,24,25,26,}\textcolor{color7}{\textit{27,28,29,30,31,32,33,34,35,36,37,38}} \\
     &  & \textsc{LiSA} (26) & 26 & 65.00\% &  \textcolor{color2}{7,8,9,10,18,19,20,21,22,23,24,25,26,27,28,29,30,31,32,33,34,35,36,37,38,39} \\
    \bottomrule
    \end{tabular}
}

  \caption{The configurations of the directly sharing attention (DS) and \textsc{LiSA} models. We report the proportion of layers employing DS or \textsc{LiSA} attention mechanisms relative to the total number of layers. For instance, LLaMA3-8B+\textsc{LiSA} (27) reduces redundant attention calculations within $84\%$ of the total layers. Layers applied DS and \textsc{LiSA} are indicated in \textcolor{color7}{\textit{blue}} (italicized) and \textcolor{color2}{red}, respectively. Layer numbering starts from 1.}
  \label{app tab:LiSA_pattern}
\end{table*}

\paragraph{Training strategy.} We only train the newly involved parameters, i.e., those of attention heads alignment and difference compensation modules, which further reduce the training cost of \textsc{LiSA}. For instance, only 56 million parameters in LLaMA3-8B ($0.7\%$ of total) are trained to apply \textsc{LiSA} to more than half of the layers. Moreover, to achieve efficient uptraining, we leverage the knowledge distillation technique. Aligning with feature-based knowledge methods \cite{DBLP:journals/corr/RomeroBKCGB14,DBLP:conf/eccv/PassalisT18,DBLP:conf/nips/KimPK18}, we regard the original model as a teacher and use its attention scores $A_n^*$ as a supervisory signal. Let $S$ be the set of indices corresponding to the layers that are equipped with \textsc{LiSA}. Then our regression loss function is formulated as follows:
\begin{align}
\mathcal{L}_{\text{KD}} = \frac{1}{|S|} \sum_{n \in S} \mathcal{L}_\delta (A_n, A_n^*)
\end{align}
where $|S|$ denotes the number of layers in the set $S$ and $\mathcal{L}_\delta (\cdot)$ stands for the Huber loss\footnote{See Appendix \ref{app:training_setups} for the complete formulation.} \cite{huber1992robust}. We also uptrain models on the language modeling task. Given the prefix $x_{<t}=\{x_1, x_2, ..., x_{t-1}\}$, the corresponding loss function can be expressed by:
\begin{align}
\mathcal{L}_{\text{LM}} = -\frac{1}{l} \sum_{t=1}^l \log p(x_t | x_{<t}).
\end{align}
Integrating these optimizing goals by a predefined weight $\beta$, then our overall loss function is
\begin{align}
\mathcal{L} = \beta \mathcal{L}_{\text{KD}} + (1 - \beta) \mathcal{L}_{\text{LM}}.
\end{align}
According to the ablation study shown in Table \ref{tab:ablation_study_of_each_sub-module_in_LiSA}, we set $\beta$ to $0.25$ for all experiments.

\subsection{Theoretical Efficiency Analysis}
\label{sec:Theoretical Analysis of Efficiency}
The efficiency of \textsc{LiSA} primarily arises from significantly compressing the K cache, which allows for larger batch sizes and longer sequences under reduced memory pressure during generation, thereby enhancing throughput. Here, we theoretically analyze the memory usage during inference. Its generation process comprises two stages for an LLM armed with the KV cache technique. In the \textbf{prefilling} stage, the memory saved by compressing K cache in $|S|$ layers with \textsc{LiSA} is $|S| \times h \times l \times (d_k-r) \times 2$ bytes, while storing an attention weight matrix requires $h \times l \times l \times 2$ bytes. Therefore, the total memory reduced by \textsc{LiSA} is $h \times l \times (|S| \times (d_k-r) - l) \times 2$ bytes. When the input sequence length $l$ exceeds $|S| \times (d_k-r)$, more memory is consumed. In the \textbf{decoding} stage, \textsc{LiSA} continues to compress the K cache as before and introduces a small weight matrix occupying $h \times l \times 2$ bytes. Consequently, the memory reduction by \textsc{LiSA} is $h \times l \times (|S| \times (d_k-r) - 1) \times 2$ bytes. Given that $|S| \times (d_k-r) \gg 1$, \textsc{LiSA} consistently saves memory in this stage.

Indeed, we can avoid additional memory consumption in the prefilling stage by leveraging the original attention mechanism for the initial inference step. To utilize \textsc{LiSA} in subsequent inference steps, one should calculate and store $K_{LR}$ instead of $K$ in the KV cache. The only difference between this decoding strategy and using \textsc{LiSA} throughout all inference steps is that the original attention weights are used in the first inference step. Thus, this approach is lossless, which is also empirically demonstrated in Table \ref{app tab:normal_forward}. We denote this decoding strategy as NF and do not apply it unless stated.

\begin{table*}[!htb]
\Large
  \centering
  
 \resizebox{1.0\linewidth}{!}{
    \begin{tabular}{lccccccccc}
    \toprule
    \multirow{2.5}{*}{\textbf{Model}} & \multirow{2.5}{*}{\makecell{\textbf{Trained} \\ \textbf{Param. (\%)}}} & \multirow{2.5}{*}{\makecell{\textbf{Saved} \\ \textbf{Param. (\%)}}} & \multirow{2.5}{*}{\makecell{\textbf{Compressing} \\ \textbf{Q (times)}}} & \multirow{2.5}{*}{\makecell{\textbf{Compressing} \\ \textbf{K (times)}}} & \multicolumn{5}{c}{\textbf{Commonsense \& Reading Comprehension}} \\
    \cmidrule(lr){6-10}
    &  &  &  &  & PIQA  & BoolQ  & WinoGrande  & CoQA  & OBQA (5)  \\
    \midrule
    LLaMA3-8B & - & - & - & $4\times$ & 80.69 & 81.13 & 73.40 & 67.40 & 46.60 \\
    DS (17) & - & \enspace 4.44 & all & all & 68.61 & 75.72 & 65.19 & 12.67 & 30.20 \\
    DS$_{LoRA}$ (17) & 0.70 & \enspace 4.44 & all & all & 79.82 & 80.15 & 72.06 & \enspace 0.00 & 45.00 \\
    DS$_{LoRA}$ (21) & 0.86 & \enspace 5.48 & all & all & 77.80 & 72.94 & 63.69 & \enspace 0.00 & 43.60 \\
    DS$_{LoRA}$ (27) & 1.11 & \enspace 7.05 & all & all & 73.83 & 58.13 & 53.35 & \enspace 0.00 & 36.20 \\
    \rowcolor{black!10} \textsc{LiSA} (17) & 0.70 & \enspace 3.74 & $6\times$ & $24\times$ & 79.87 & 81.65 & 73.95 & 63.53 & 46.20 \\
    \rowcolor{black!10}\textsc{LiSA}$_{SL}$ (7+10) & 0.46 & \enspace 3.98 & [$4\times$, all] & [1$6\times$, all] & 80.63 & 79.17 & 73.32 & 64.90 & 43.80 \\
    \rowcolor{black!10}\textsc{LiSA} (21) & 0.86 & \enspace 4.62 & $6\times$ & $24\times$ & 80.14 & 78.78 & 72.14 & 61.52 & 46.20 \\
    \rowcolor{black!10}\textsc{LiSA} (27) & 1.11 & \enspace 5.94 & $6\times$ & $24\times$ & 80.69 & 77.86 & 70.17 & 60.23 & 46.80 \\
    \midrule
    LLaMA2-7B & - & - & - & - & 79.11 & 77.74 & 68.98 & 63.88 & 42.60 \\
    DS (17) & - & \enspace 8.47 & all & all & 62.08 & 64.89 & 60.69 & \enspace 1.00 & 26.60 \\
    DS$_{LoRA}$ (17) & 1.33 & \enspace 8.47 & all & all & 77.20 & 76.39 & 66.54 & \enspace 0.00 & 40.00 \\
    DS$_{LoRA}$ (21) & 1.64 & 10.46 & all & all & 75.57 & 67.92 & 59.75 & \enspace 0.00 & 41.40 \\
    \rowcolor{black!10}\textsc{LiSA} (17) & 1.33 & \enspace 7.14 & $6\times$ & $6\times$ & 78.84 & 76.79 & 74.51 & 60.58 & 45.80 \\
    \rowcolor{black!10}\textsc{LiSA}$_{SL}$ (7+10) & 0.87 & \enspace 6.37 & [$4\times$, all] & [$4\times$, all] & 78.02 & 76.67 & 68.11 & 61.33 & 41.00 \\
    \rowcolor{black!10}\textsc{LiSA} (21) & 1.64 & \enspace 8.82 & $6\times$ & $6\times$ & 78.62 & 73.24 & 68.27 & 52.33 & 41.40 \\
    \midrule
    LLaMA2-13B & - & - & - & - & 80.52 & 80.55 & 72.14 & 66.37 & 49.00 \\
    DS (22) & - & \enspace 8.77 & all & all & 64.53 & 70.46 & 60.14 & \enspace 3.17 & 30.60 \\
    DS$_{LoRA}$ (22) & 1.37 & \enspace 8.77 & all & all & 77.91 & 77.83 & 69.30 & \enspace 0.00 & 48.20 \\
    DS$_{LoRA}$ (26) & 1.62 & 10.37 & all & all & 78.45 & 76.12 & 68.11 & \enspace 0.00 & 44.00 \\
    \rowcolor{black!10}\textsc{LiSA} (22) & 1.37 & \enspace 7.40 & $6\times$ & $6\times$ & 79.54 & 80.31 & 71.74 & 66.97 & 50.40 \\
    \rowcolor{black!10}\textsc{LiSA}$_{SL}$ (10+12) & 0.99 & \enspace 8.15 & [$4\times$, all] & [$4\times$, all] & 79.38 & 80.67 & 71.74 & 65.42 & 50.60 \\
    \rowcolor{black!10}\textsc{LiSA} (26) & 1.62 & \enspace 8.74 & $6\times$ & $6\times$ & 79.82 & 78.96 & 72.14 & 63.13 & 48.60 \\
    \toprule
    \multirow{2.5}{*}{\textbf{Model}} & \multicolumn{3}{c}{\textbf{Continued}} & \multicolumn{3}{c}{\textbf{World Knowledge}} & \makecell{\textbf{Reasoning}} & \makecell{\textbf{LM}} & \multirow{2.5}{*}{\makecell{\textbf{Avg. Perf.} \\ \textbf{Preserved (\%)}}} \\ 
    \cmidrule(lr){2-4} \cmidrule(lr){5-7} \cmidrule(lr){8-8} \cmidrule(lr){9-9}
    & ARC-E  & ARC-C (25) & HellaSwag (10) & TriviaQA  & NQ (5) & MMLU (5) & GSM8K CoT (8) & LAMBADA \\
    \midrule
    LLaMA3-8B & 77.61 & 59.30 & 82.26 & 63.39 & 29.14 & 64.98 & 51.71 & 3.48 & - \\
    DS (17) & 41.04 & 29.86 & 50.00 & \enspace 0.73 & \enspace 1.11 & 23.96 &\enspace 1.74 & 936.10 & \textbf{46.67} \\
    DS$_{LoRA}$ (17) & 75.38 & 56.66 & 78.44 & \enspace 4.80 & \enspace 0.08 & 49.14 & 25.85 & 6.28 & \textbf{67.83} \\
    DS$_{LoRA}$ (21) & 67.97 & 47.27 & 73.37 & \enspace 4.41 & \enspace 0.91 & 35.16 & \enspace 4.85 & 7.86 & \textbf{58.06} \\
    DS$_{LoRA}$ (27) & 54.21 & 28.07 & 57.88 & \enspace 3.81 & \enspace 0.03 & 23.85 & \enspace 0.38 & 22.47 & \textbf{45.38} \\
    \rowcolor{black!10}\textsc{LiSA} (17) & 79.29 & 58.96 & 81.17 & 57.66 & 27.17 & 61.22 & 45.94 & 3.79 & \textbf{96.77} \\
    \rowcolor{black!10}\textsc{LiSA}$_{SL}$ (7+10) & 77.74 & 59.04 & 79.85 & 53.11 & 25.01 & 61.69 & 42.76 & 4.89 & \textbf{94.31} \\
    \rowcolor{black!10}\textsc{LiSA} (21) & 74.28 & 55.12 & 80.83 & 52.38 & 26.04 & 59.52 & 39.27 & 3.96 & \textbf{92.63} \\
    \rowcolor{black!10}\textsc{LiSA} (27) & 74.92 & 53.33 & 79.43 & 43.65 & 25.65 & 50.58 & 31.77 & 4.37 & \textbf{88.38} \\
    \midrule
    LLaMA2-7B & 74.58 & 53.24 & 78.59 & 52.54 & 26.01 & 45.94 & 14.18 & 3.76 & - \\
    DS (17) & 36.66 & 29.52 & 35.82 & \enspace 0.11 & \enspace 0.39 & \enspace 1.85 & \enspace 0.53 & 20594.64 & \textbf{39.47} \\
    DS$_{LoRA}$ (17) & 63.05 & 46.33 & 72.85 & \enspace 3.93 & \enspace 0.00 & 35.19 & \enspace 6.14 & 8.28 & \textbf{64.82} \\
    DS$_{LoRA}$ (21) & 62.63 & 42.32 & 67.90 & \enspace 3.64 & \enspace 0.91 & 25.32 & \enspace 2.05 & 8.91 & \textbf{58.05} \\
    \rowcolor{black!10}\textsc{LiSA} (17) & 71.09 & 51.62 & 76.96 & 50.93 & 21.94 & 43.83 & 12.96 & 4.26 & \textbf{97.26} \\
    \rowcolor{black!10}\textsc{LiSA}$_{SL}$ (7+10) & 71.04 & 51.19 & 76.03 & 39.10 & 17.89 & 42.05 & \enspace 8.26 & 5.20 & \textbf{89.11} \\
    \rowcolor{black!10}\textsc{LiSA} (21) & 71.17 & 50.26 & 75.49 & 39.01 & 17.51 & 35.37 & 10.24 & 4.71 & \textbf{87.36} \\
    \midrule
    LLaMA2-13B & 77.48 & 59.73 & 82.50 & 60.86 & 29.81 & 50.51 & 24.03 & 3.3692 & - \\
    DS (22) & 39.44 & 29.86 & 44.44 & \enspace 0.28 & \enspace 0.42 & 43.57 & \enspace 0.00 & 12877.5226 & \textbf{46.76} \\
    DS$_{LoRA}$ (22) & 69.95 & 51.79 & 77.93 & \enspace 0.36 & \enspace 1.50 & 47.16 & \enspace 7.96 & 4.8904 & \textbf{65.95} \\
    DS$_{LoRA}$ (26) & 67.51 & 49.23 & 75.96 & \enspace 0.30 & \enspace 0.71 & 42.99 & \enspace 6.14 & 5.7901 & \textbf{62.61} \\
    \rowcolor{black!10}\textsc{LiSA} (22) & 74.49 & 57.17 & 81.20 & 55.67 & 26.45 & 48.75 & 21.30 & 3.7399 & \textbf{96.44} \\
    \rowcolor{black!10}\textsc{LiSA}$_{SL}$ (10+12) & 72.85 & 56.48 & 80.77 & 39.41 & 24.63 & 50.05 & 19.03 & 3.9728 & \textbf{92.68} \\
    \rowcolor{black!10}\textsc{LiSA} (26) & 74.54 & 54.95 & 80.81 & 36.62 & 24.63 & 48.09 & 15.85 & 3.8679 & \textbf{90.13} \\
    \bottomrule
    \end{tabular}
}
  \caption{Performance on 13 typical benchmarks. Columns 2–5 present several attributes of each model. For instance, equipping LLaMA3-8B with \textsc{LiSA}$_{SL}$ (7+10) requires training $0.46\%$ of total parameters while saving $3.98\%$ of them. In this configuration, \textsc{LiSA} compresses the $Q$ and $K$ matrices in the first 7 layers by factors of $4 \times$ and $16 \times$, respectively, while the remaining 10 layers cut down the $Q$ and $K$ matrices via DS. In the last column, we report the average preserved performance across all benchmarks, excluding LAMBADA. We only report the performance of DS (17) and DS (21) because sharing more layers further impairs overall performance.}
  \label{tab:14benchmarks}
\end{table*}

\subsection{Experiment Settings}

\paragraph{Model configuration.} We selected LLaMA3-8B, LLaMA2-7B, and LLaMA2-13B as the base models. The first two models, LLaMA3-8B and LLaMA2-7B, each comprise 32 layers with 32 attention heads of 128 dimensions each. In contrast, LLaMA2-13B consists of 40 layers, also with 32 attention heads of 128 dimensions each. We designed several layer-wise sharing configurations, detailed in Table \ref{app tab:LiSA_pattern}. Specifically, \textsc{LiSA} denotes the default structure that the attention heads alignment module uses a two-layer FFN along with \texttt{ReLU} as the activation function. This model compresses $Q, K$ by $6\times$ (i.e., $r=20$ compared to $d_k=128$). While \textsc{LiSA}$_{SL}$ stands for another structure that leverages a one-layer FFN for alignment and compresses the $Q, K$ by $4\times$ (i.e., $r=32$ compared to $d_k=128$). Additionally, directly sharing attention is applied to deep layers. 

For the baselines, we report the performance of directly sharing attention (DS) and its uptrained version (DS$_{LoRA}$). Specifically, we use the LoRA training method \cite{DBLP:conf/iclr/HuSWALWWC22} to ensure the number of learnable parameters matches that of \textsc{LiSA}. All trainable models are trained on 4.2 billion tokens, which is a subset of RedPajama-1T\footnote{\url{https://huggingface.co/datasets/togethercomputer/RedPajama-Data-1T}}. Other setups are reported in Appendix \ref{app:training_setups}.

\paragraph{Performance Evaluation.} Following LLaMA2 and LLaMA3, we conducted extensive evaluations on 13 typical downstream benchmarks. We reported the 0-shot accuracy on PIQA, BoolQ \cite{DBLP:conf/naacl/ClarkLCK0T19}, WinoGrande \cite{DBLP:journals/cacm/SakaguchiBBC21}, ARC easy (ARC-E) \cite{DBLP:journals/corr/abs-2102-03315}, 5-shot accuracy on OBQA \cite{DBLP:conf/emnlp/MihaylovCKS18} and MMLU, 10-shot accuracy on HellaSwag \cite{DBLP:conf/acl/ZellersHBFC19}, 25-shot accuracy on ARC challenge (ARC-C). For the exact match score, we reported 0-shot performance on TriviaQA \cite{DBLP:conf/acl/JoshiCWZ17}, 8-shot chain-of-thought \cite{DBLP:conf/nips/Wei0SBIXCLZ22} performance on GSM8K, and 5-shot performance on Natural Questions (NQ) \cite{DBLP:journals/tacl/KwiatkowskiPRCP19}. Furthermore, we included 0-shot extract match score on CoQA \cite{DBLP:journals/tacl/ReddyCM19} and the perplexity on LAMBADA \cite{DBLP:conf/acl/PapernoKLPBPBBF16}. More details are provided in Appendix \ref{app:evaluation_setups}.

\begin{table*}[htb]
\Large
  \centering
  
 \resizebox{1.0\linewidth}{!}{
    \begin{tabular}{lllllllllc}
    \toprule
    \textbf{[Input, Output]} & \textbf{[128, 512]} & \textbf{[128, 1024]} & \textbf{[512, 128]} & \textbf{[512, 1024]} & \textbf{[512, 3072]} & \textbf{[1024, 1024]} & \textbf{[1024, 3072]} & \textbf{[2048, 512]} & \textbf{Avg. Improv.} \\
    \midrule
    LLaMA3-8B & 538 & 395 & 1597  & 408  & 201  & 416  & 194  & 684 & -\\
    \rowcolor{black!10}\textsc{LiSA} (17) & 562 {\small +4.4\%} & 427 {\small +8.1\%} & 1669 {\small +4.5\%} & 449 {\small +10.1\%} & 232 {\small +15.9\%} & 461 {\small +10.7\%} & 221 {\small +13.6\%} & 775 {\small +13.2\%} & \textbf{10.1\%} \\
    \rowcolor{black!10}\textsc{LiSA} (21) & 582 {\small +8.1\%} & 445 {\small +12.7\%} & 1736 {\small +8.7\%} & 463 {\small +13.5\%} & 241 {\small +20.3\%} & 472 {\small +14.7\%} & 233 {\small +19.9\%} & 789 {\small +15.3\%} & \textbf{14.2\%} \\
    \rowcolor{black!10}\textsc{LiSA} (27) & 596 {\small +10.8\%} & 455 {\small +15.3\%} & 1797 {\small +12.5\%} & 483 {\small +18.5\%} & 260 {\small +29.4\%} & 501 {\small +20.5\%} & 247 {\small +27.2\%} & 834 {\small +21.9\%} & \textbf{19.5\%} \\
    \rowcolor{black!10}\textsc{LiSA} $_{SL}$ (7+10) & 563 {\small +4.6\%} & 433 {\small +9.7\%} & 1733 {\small +8.6\%} & 455 {\small +11.7\%} & 231 {\small +15.2\%} & 459 {\small +12.6\%} & 225 {\small +12.0\%} & 774 {\small +13.1\%} & \textbf{10.9\%} \\
    \midrule
    LLaMA2-7B & 875 & 544 & 2256 & 544 & 200 & 506 & 193 & 727 & -\\
    \rowcolor{black!10}\textsc{LiSA} (17) & 1008 {\small +15.2\%} & 645 {\small +18.7\%} & 2520 {\small +11.7\%} & 673 {\small +23.8\%} & 248 {\small +23.9\%} & 553 {\small +9.1\%} & 233 {\small +20.8\%} & 862 {\small +18.6\%}  & \textbf{17.7\%} \\
    \rowcolor{black!10}\textsc{LiSA} (21) & 1396 {\small +59.5\%} & 683 {\small +25.6\%} & 3062 {\small +35.7\%} & 707 {\small +30.0\%} & 260 {\small +30.0\%} & 653 {\small +29.0\%} & 250 {\small +29.2\%} & 870 {\small +19.7\%}  & \textbf{32.3\%} \\
    \rowcolor{black!10}\textsc{LiSA} $_{SL}$ (7+10) & 1224 {\small +39.9\%} & 696 {\small +28.0\%} & 2751 {\small +21.9\%} & 605 {\small +11.2\%} & 224 {\small +12.0\%} & 549 {\small +8.4\%} & 209 {\small +8.2\%} & 803 {\small +10.5\%} & \textbf{17.5\%} \\
    \midrule
    LLaMA2-13B & 710 & 409 & 1679 & 352 & 140 & 320 & 127 & 460 & -\\
    \rowcolor{black!10}\textsc{LiSA} (22) & 997 {\small +40.5\%} & 560 {\small +36.9\%} & 2059 {\small +22.6\%} & 474 {\small +34.7\%} & 187 {\small +33.2\%} & 432 {\small +35.0\%} & 172 {\small +35.9\%} & 549 {\small +19.3\%}  & \textbf{32.2\%} \\
    \rowcolor{black!10}\textsc{LiSA} (26) & 1043 {\small +46.9\%} & 584 {\small +42.9\%} & 2155 {\small +28.3\%} & 504 {\small +43.2\%} & 198 {\small +41.2\%} & 460 {\small +43.7\%} & 185 {\small +46.0\%} & 591 {\small +28.5\%}  & \textbf{40.1\%} \\
    \rowcolor{black!10}\textsc{LiSA} $_{SL}$ (10+12) & 896 {\small +26.2\%} & 492 {\small +20.2\%} & 1902 {\small +13.3\%} & 408 {\small +16.0\%} & 158 {\small +12.4\%} & 374 {\small +6.3\%} & 146 {\small +4.2\%} & 532 {\small +15.5\%} & \textbf{14.3\%} \\
    \bottomrule
    \end{tabular}
  }
  \caption{Throughput (token/s) on a A800 80GB GPU with different systems. ``[128, 512]'' denotes a prompt length of 128 and a generation length of 512.}
  \label{tab:throughput}
\end{table*}

\begin{table}[!htp]
\Large
  \centering
  
 \resizebox{1.0\linewidth}{!}{
    \begin{tabular}{lllll}
    \toprule
    \textbf{Batch Size} & \makecell{\quad \textbf{8}} & \makecell{\quad \textbf{8}} & \makecell{\ \ \,\textbf{16}} & \makecell{\  \textbf{32}} \\
    \textbf{[Input, Output]} & \textbf{[2048, 512]} & \textbf{[512, 1024]} & \textbf{[128, 1024]} & \textbf{[128, 512]} \\
    \midrule
    LLaMA3-8B & 35.43 & 46.31 & 61.23 & 43.37 \\
    \rowcolor{black!10}\textsc{LiSA} (17) & 33.69 {\small 4.9\%} & 43.26 {\small 6.6\%} & 57.57 {\small 6.0\%} & 41.04 {\small 5.4\%} \\
    \rowcolor{black!10}\textsc{LiSA} (21) & 33.45 {\small 5.6\%} & 42.53 {\small 8.2\%} & 56.42 {\small 7.9\%} & 39.68 {\small 8.5\%} \\
    \rowcolor{black!10}\textsc{LiSA} (27) & 32.68 {\small 7.8\%} & 41.58 {\small 10.2\%} & 54.32 {\small 11.3\%} & 39.67 {\small 8.5\%} \\
    \rowcolor{black!10}\textsc{LiSA}$_{SL}$ (7+10) & 31.35 {\small 11.5\%} & 41.99 {\small 9.3\%} & 55.54 {\small 9.3\%} & 40.44 {\small 6.8\%} \\
    \midrule
    LLaMA2-7B & 32.49 & 37.72 & 45.15 & 27.8 \\
    \rowcolor{black!10}\textsc{LiSA} (17) & 28.37 {\small 12.7\%} & 34.12 {\small 9.5\%} & 40.37 {\small 10.6\%} & 25.06 {\small 9.9\%} \\
    \rowcolor{black!10}\textsc{LiSA} (21) & 27.43 {\small 15.6\%} & 32.33 {\small 14.3\%} & 39.66 {\small 12.2\%} & 24.22 {\small 12.9\%} \\
    \rowcolor{black!10}\textsc{LiSA}$_{SL}$ (7+10) & 29.29 {\small 9.8\%} & 34.21 {\small 9.3\%} & 40.99 {\small 9.2\%} & 21.52 {\small 22.6\%} \\
    \midrule
    LLaMA2-13B & 48.88  & 56.25  & 63.59  & 35.53  \\
    \rowcolor{black!10}\textsc{LiSA} (22) & 41.23 {\small 15.7\%} & 51.02 {\small 9.3\%} & 57.16 {\small 10.1\%} & 30.83 {\small 13.2\%}  \\
    \rowcolor{black!10}\textsc{LiSA} (26) & 39.59 {\small 19.0\%} & 50.62 {\small 10.0\%} & 56.74 {\small 10.8\%} & 29.82 {\small 16.1\%} \\
    \rowcolor{black!10}\textsc{LiSA}$_{SL}$ (10+12) & 43.36 {\small 11.3\%} & 50.96 {\small 9.4\%} & 57.67 {\small 9.3\%} & 31.43 {\small 11.5\%} \\
    \bottomrule
    \end{tabular}
  }
  \caption{Generation latency (sec) on a A800 80GB GPU with different systems.}
  \label{tab:generation_latency}
\end{table}

\paragraph{Efficiency Evaluation.} Aligning with \citet{DBLP:conf/nips/Zhang00CZC0TRBW23}, we evaluated the end-to-end throughput and latency of our system. Throughput is defined as the number of prompted and generated tokens per unit of time, calculated as (prompted tokens + generated tokens) / (prompt time + decoding time). Latency refers to the total time consumed by the whole generation process. We conducted each experiment 10 times and reported the averaged results to ensure reliability and consistency across evaluations. All evaluated models are equipped with the KV cache to speed up generation. Moreover, it is worth noting that LLaMA3-8B serves as a strong baseline, with the GQA technique \cite{DBLP:conf/emnlp/AinslieLJZLS23} compressing the KV cache by $4 \times$ compared to MHA.

\subsection{Main Results}

\paragraph{Performance on downstream tasks.} Table \ref{tab:14benchmarks} illustrates that employing \textsc{LiSA} to share attention weights across layers in existing LLMs results in minimal performance loss across various domains and sizes of models. Notably, our best-performing model, LLaMA3-8B+\textsc{LiSA} (17), which implements the \textsc{LiSA} structure in over half of its layers, maintains comparable performance as the original model while adding only a few trainable parameters. Furthermore, despite sharing weights across most layers, LLaMA3-8B+\textsc{LiSA} (27) shows minimal performance degradation on most benchmarks. In comparison, directly sharing attention (DS) leads to significant performance declines. For instance, applying DS to 17 layers in LLaMA3-8B results in the model retaining merely $46.67\%$ of its original performance. Even though uptraining the DS with the same number of parameters and an equivalent amount of data, the significant performance declines observed in the DS$_{LoRA}$ models highlight severe impairments to the original capabilities. These findings underscore \textsc{LiSA}'s effectiveness as a robust solution for sharing attention weights across layers in LLMs.

\paragraph{End-to-end inference efficiency evaluation.} We first examine the throughput \textit{under the limitation of 80GB memory} on the same A800 GPUs with variable batch sizes. To avoid extra memory consumption, we apply the NF decoding strategy when the length of the input sequence surpasses $2048$. Table \ref{tab:throughput} shows that \textsc{LiSA} achieves significant throughput improvements across a range of input-output scenarios, with increases ranging from $17.5\%$ to $32.3\%$ for LLaMA2-7B. Moreover, when \textsc{LiSA} is implemented on a larger model like LLaMA2-13B, a greater improvement in throughput is observed, highlighting that \textsc{LiSA}'s benefits grow as the model scales. It is important to note that LLaMA3-8B serves as a robust baseline, where the GQA technique has compressed the KV cache by $4\times$ compared to MHA. When equipped with \textsc{LiSA}, $10.1\%$ to $19.5\%$ improvements are still observed. Additionally, we report the generation latency \textit{under the same batch size settings} in Table \ref{tab:generation_latency}, which indicates that \textsc{LiSA} consistently reduces the latency compared to the baseline.

\begin{table}[htbp]
  \centering
 \LARGE
 \resizebox{1.0\linewidth}{!}{
    \begin{tabular}{lcccccc}
    \toprule
    \textbf{Model} & \textbf{OBQA} & \textbf{HellaSwag} & \textbf{PIQA} & \textbf{BoolQ} & \textbf{WinoGrande} & \textbf{ARC-E} \\
    \midrule
    MHA & 24.20 & 28.95 & 58.49 & 61.47 & 52.01 & 35.44 \\
    GQA & 24.40 & 28.29 & 59.03 & 57.58 & 50.91 & 35.65 \\
    DS (8) & 25.20 & 27.68 & 58.71 & 61.10 & 52.49 & 34.18 \\
    \textsc{LiSA}$_{plus}$ (8) & 26.20 & 27.92 & 58.65 & 62.02 & 50.20 & 35.14 \\
    \bottomrule
    \end{tabular}
}

  \caption{Performance of different attention models pre-trained from scratch. The original model consists of $12$ layers, each with $12$ attention heads, and an attention head dimension of $64$. The layer-wise sharing configuration for DS (8) and Plus (8) is ``2,3,4,6,7,8,10,11''. For the GQA model, we set the number of KV attention heads to 2.}
  \label{app tab:performance_of_different_attention_models_pre_trained_from_scratch}
\end{table}

\subsection{Pre-training From Scratch}
We argue that if heads are explicitly aligned by directly sharing when training an LLM from scratch, then the attention heads alignment module can be discarded, and the predicted difference can be directly added to the shared weight matrix, thus a more concise and efficient \textsc{LiSA}$_{plus}$ will be achieved. To investigate this, we pre-train LLaMA-like models with 12 layers and 164 million parameters on 10 billion tokens\footnote{This pre-training corpus takes up 40GB which aligns with GPT-2 \cite{radford2019language}.}. Performance shown in Table \ref{app tab:performance_of_different_attention_models_pre_trained_from_scratch} demonstrates that both directly sharing weights and applying \textsc{LiSA}$_{plus}$ across two-thirds of total layers are lossless. The evaluation losses are shown in Figure \ref{fig:train_from_scratch_loss}. These experimental results not only show the potential of \textsc{LiSA} in the pre-training LLMs from scratch but also mirror our observation of the redundancy within the inter-layer attention mechanism again.

\subsection{Task-specific \textsc{LiSA}}
For certain vertical-domain tasks, it is preferable to estimate \textsc{LiSA}'s performance under a given attention-sharing configuration in a training-free manner. Thus, we can iteratively refine the configuration based on its estimated effectiveness, thereby improving overall results. We show that DS can serve as a substitute to estimate \textsc{LiSA}'s performance without training, and then we provide refinement suggestions for different situations.

Specifically, we assess the performance loss of DS under the same attention-sharing configuration as \textsc{LiSA}. Here, DS serves as the lower bound of \textsc{LiSA}. If DS exhibits large performance degradation, \textsc{LiSA} must devote more effort to preserve performance, indicating that the corresponding task is more challenging. As shown in Table \ref{app_tab:trade_off}, we categorize 12 tasks into three levels based on DS's performance loss: Low-loss (green): loss within $0\%-40\%$; Moderate-loss (yellow): within $40\%-70\%$; and High-loss (red): loss within $70\%-100\%$.

After determining the level, we outline several criteria: For low-loss tasks, \textsc{LiSA} performs without loss and even slightly outperforms the original model, allowing us to further improve efficiency by applying attention sharing in more layers; For a moderate-loss task, \textsc{LiSA} can generally achieve near-lossless performance; For a high-loss task, \textsc{LiSA} can still recover most of the performance. However, if stricter performance requirements exist, we can reduce the number of layers employing \textsc{LiSA} to minimize performance loss.

\begin{figure}
    \centering
    \begin{tikzpicture}[scale=0.8]
    \begin{axis}[
        width =80mm,
        height=50mm,
        enlarge y limits=0.2,
        xbar stacked, 
        bar width=10mm,
        ytick=data,
        yticklabels={\footnotesize{\textsc{LiSA}$_{SL}$ (7+10)}, \textsc{LiSA} (21), \textsc{LiSA} (17)},
        xtick={0,20,40,60,80,100},
        xlabel={Percentage (\%)},
        xmin=0, xmax=100,
        area legend,
        y dir=reverse,
        nodes near coords,
        legend style={at={(0.5,1.3)},
        anchor=north,legend columns=-1},
        ]
    \addplot+[xbar] plot coordinates {(52.2,2)(51.1,1)(52.3,0)};
    \addplot+[xbar] plot coordinates {(47.8,2)(48.9,1)(47.7,0)};
    \legend{Win, Lose}
    \end{axis}
\end{tikzpicture}
    \caption{The win rate of \textsc{LiSA} models compared with LLaMA3-8B. All models have been fine-tuned using instruction data.}
    \label{fig:win_rate}
\end{figure}

\begin{table}[htp]
  \centering
  \LARGE
 \resizebox{1.0\linewidth}{!}{
    \begin{tabular}{lcccccc}
    \toprule
    \multirow{2.5}{*}{\textbf{Model}} & \multicolumn{2}{c}{\textbf{BoolQ}}  & \multicolumn{2}{c}{\textbf{PIQA}} & \multicolumn{2}{c}{\textbf{ARC-E}} \\
    \cmidrule(lr){2-3} \cmidrule(lr){4-5} \cmidrule(lr){6-7}
     & 5-shot & 10-shot & 5-shot & 10-shot & 5-shot & 10-shot \\
    \midrule
    LLaMA3-8B & 82.26 & 83.30 & 82.64 & 82.86 & 84.81 & 84.81 \\
    \textsc{LiSA} (17) & 83.52 & 84.31 & 82.21 & 82.43 & 84.30 & 84.05 \\
    \textsc{LiSA} (21) & 77.37 & 77.00 & 81.28 & 82.26 & 82.58 & 82.62 \\
    \textsc{LiSA} (27) & 76.15 & 74.86 & 81.61 & 82.21 & 80.89 & 82.07 \\
    \bottomrule
    \end{tabular}
}

  \caption{Ablation study of different numbers of shot.}
  \label{app tab:different_shots}
\end{table}

\subsection{Ablation Study}
We present ablation studies of \textsc{LiSA} on (1) different sequence lengths, (2) instruct-tuned models, and (3) dissecting the effectiveness of each component.

\begin{table}[t!]
\LARGE
    \centering
    \resizebox{1.0\linewidth}{!}{
    \begin{tabular}{cc|rrrrrrrr}
        \toprule
        \multicolumn{2}{l|}{\textbf{Model}} & \textbf{BoolQ} & \textbf{PIQA} & \textbf{CoQA} & \textbf{MMLU} & \textbf{GSM8K} \\
        \midrule
        \multicolumn{2}{l|}{LLaMA3-8B} & 81.13 & 80.69 & 67.40 & 65.24 & 51.71 \\
        \multicolumn{2}{l|}{DS (21)} & 40.46 & 56.86 & 0.11 & 0.00 & 2.65 \\
        \multicolumn{2}{l|}{\hspace{7mm}+Align} & 74.34 & 77.48 & 50.28 & 46.53 & 7.96 \\
        \multicolumn{2}{l|}{\hspace{7mm}+Diff.} & 37.86 & 52.99 & 0.00 & 0.61 & 0.68 \\
        \multicolumn{2}{l|}{\hspace{7mm}+Both} & 76.85 & 80.09 & 63.03 & 56.78 & 26.84 \\
        \bottomrule
    \end{tabular}
    }
    \caption{Ablation study of sub-modules in \textsc{LiSA}. ``+Align'' and ``+Diff.'' mean we individually enable the attention heads alignment and the difference compensation module, respectively. ``+Both'' denotes that we use both modules at the same time.}
    \label{tab:ablation_study_of_each_sub-module_in_LiSA}
\end{table}

\begin{table}[t!]
\LARGE
    \centering
    \resizebox{1.0\linewidth}{!}{
    \begin{tabular}{cccccccccc}
        \toprule
        \multicolumn{2}{c|}{\textbf{Model Configuration}} & \textbf{BoolQ} & \textbf{PIQA} & \textbf{CoQA} & \textbf{MMLU} & \textbf{GSM8K} \\
        \midrule
        \multirow{4}{*}{\makecell{\textbf{Alignment}\\ \textbf{Structure}}} 
        & \multicolumn{1}{l|}{Plus} & 68.72 & 78.24 & 48.73 & 38.38 & 6.14 \\
        & \multicolumn{1}{l|}{SL} & 72.81 & 78.89 & 56.82 & 61.58 & 10.31 \\
        & \multicolumn{1}{l|}{\textbf{DL + ReLU}} & \textbf{76.85} & \textbf{80.09} & \textbf{63.03} & \textbf{56.78} & \textbf{26.84} \\
        & \multicolumn{1}{l|}{DL + SiLU} & 75.63 & 79.38 & 62.55 & 57.05 & 22.30 \\
        \midrule
        \multirow{3}{*}{\makecell{\textbf{Hidden}\\ \textbf{Size}}} 
        & \multicolumn{1}{l|}{128} & 76.18 & 79.33 & 61.63 & 56.37 & 24.56 \\
        & \multicolumn{1}{l|}{\textbf{256}} & \textbf{76.85} & \textbf{80.09} & \textbf{63.03} & \textbf{56.78} & \textbf{26.84} \\
        & \multicolumn{1}{l|}{512} & 76.48 & 79.16 & 61.95 & 56.30 & 24.87 \\
        \midrule
        \multirow{5}{*}{\makecell{\textbf{Rank of}\\ $\boldsymbol{W_{LR}^Q,W_{LR}^K}$}} 
        & \multicolumn{1}{l|}{128} & 74.65 & 79.49 & 60.07 & 54.62 & 21.76 \\
        & \multicolumn{1}{l|}{192} & 76.57 & 79.49 & 61.58 & 56.99 & 23.65 \\
        & \multicolumn{1}{l|}{320} & 76.85 & 80.09 & 60.03 & 56.78 & 26.84 \\
        & \multicolumn{1}{l|}{\textbf{640}} & \textbf{76.61} & \textbf{80.20} & \textbf{62.53} & \textbf{57.25} & \textbf{27.67} \\
        & \multicolumn{1}{l|}{1024} & 77.49 & 79.54 & 63.07 & 57.21 & 30.10 \\
        \midrule
        \multirow{3}{*}{\makecell{$\boldsymbol{\beta}$}}
        & \multicolumn{1}{l|}{\textbf{0.25}} & \textbf{76.85} & \textbf{80.09} & \textbf{63.03} & \textbf{56.78} & \textbf{26.84} \\
        & \multicolumn{1}{l|}{0.50} & 77.09 & 79.43 & 62.78 & 61.79 & 24.64 \\
        & \multicolumn{1}{l|}{0.75} & 75.35 & 79.00 & 61.03 & 61.89 & 18.35 \\
        \bottomrule
    \end{tabular}
    }
    \caption{Ablation study of different configurations. Plus indicates the difference matrix is added to the shared attention weight matrix. SL and DL represent one-layer and two-layer FFNs are used in the attention heads alignment module, respectively. The hidden size stands for the intermediate size of the above two-layer FFN. The default configuration denoted as \textsc{LiSA} is \textbf{bolded}.}
    \label{tab:ablation_study_of_different_configures}
\end{table}

\paragraph{$Q_1$: Does increasing the number of shots during inference affect \textsc{LiSA}'s effectiveness? $A_1$: No.} Table \ref{app tab:different_shots} displays the results of incrementally increasing the number of shots. It indicates that \textsc{LiSA} maintains robust performance, effectively leveraging different numbers of shots, similar to the performance of the original model.

\paragraph{$Q_2$: Does \textsc{LiSA} affect the performance of instruction fine-tuning? $A_2$: No.} We first fine-tune LLaMA3 and our \textsc{LiSA} models on the Alpaca dataset \cite{alpaca}, and then leverage GPT-4 to judge pairs of responses. The win rate points in Figure \ref{fig:win_rate} show that \textsc{LiSA} models even slightly outperform the baseline.

\paragraph{$Q_3$: Whether all sub-modules have been empirically verified? $A_3$: Yes.} Table \ref{tab:ablation_study_of_each_sub-module_in_LiSA} presents the results of ablating every sub-module in \textsc{LiSA}, with each model trained on 1 billion tokens. The results highlight the critical roles of the attention heads alignment and the difference compensation modules in preserving performance. Preliminary experiments are detailed in Table \ref{tab:ablation_study_of_different_configures}, demonstrating the effectiveness of each setup in \textsc{LiSA}.

\section{Conclusion}
In this work, we first provide a comprehensive layer-wise redundancy analysis of the attention mechanism in LLMs. We find that: (1) Most transformer layers perform a highly similar attention pattern; (2) Individual attention heads hinder from directly sharing attention weight; (3) Shallow layers are sensitive to little deviations in attention weight, while deep layers are not. Driven by these insights, we propose a learnable sharing attention mechanism for existing well-trained LLMs. Comprehensive experiments demonstrate that our method significantly reduces the inter-layer redundancy of attention, achieving efficient throughput and memory with minimal loss. As far as we know, this is the first attempt to analyze and reduce inter-layer redundancy of attention weights within LLMs. In future work, we plan to investigate whether this problem occurs in large models of other modalities.

\section*{Acknowledgements}
This work was supported in part by the National Science Foundation of China (Nos. 62276056 and U24A20334), the Yunnan Fundamental Research Projects (No.202401BC070021), the Yunnan Science and Technology Major Project (No. 202502AD080014), and the Program of Introducing Talents of Discipline to Universities, Plan 111 (No.B16009). The authors thank Siming Wu and Peinan Feng for their valuable advice, and extend our sincere gratitude to action editor Xavier Carreras and the anonymous TACL reviewers for their insightful feedback and constructive suggestions.

\bibliography{tacl2021}
\bibliographystyle{acl_natbib}


\appendix

\begin{figure*}
\captionsetup[subfloat]{labelfont=scriptsize, textfont=scriptsize}
\includegraphics[width = 0.24\textwidth]{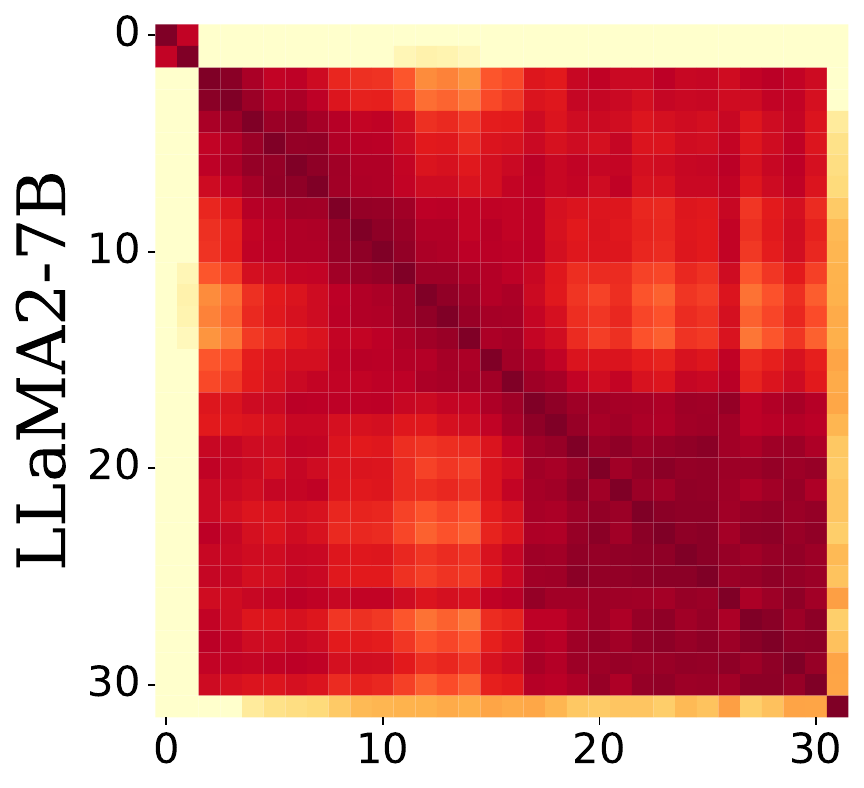}
\hfill
\includegraphics[width = 0.22\textwidth]{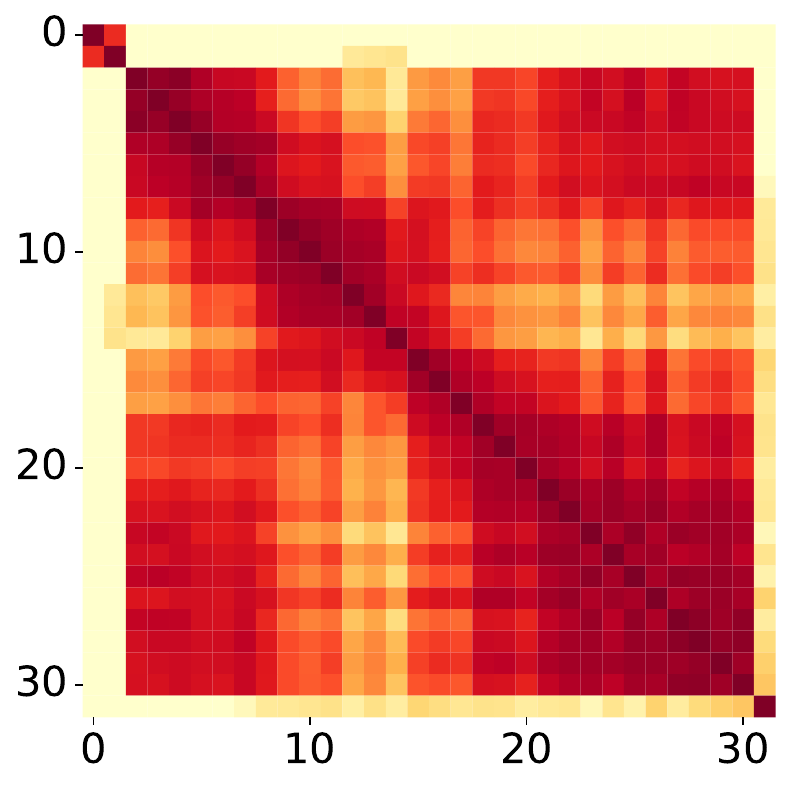}
\hfill
\includegraphics[width = 0.22\textwidth]{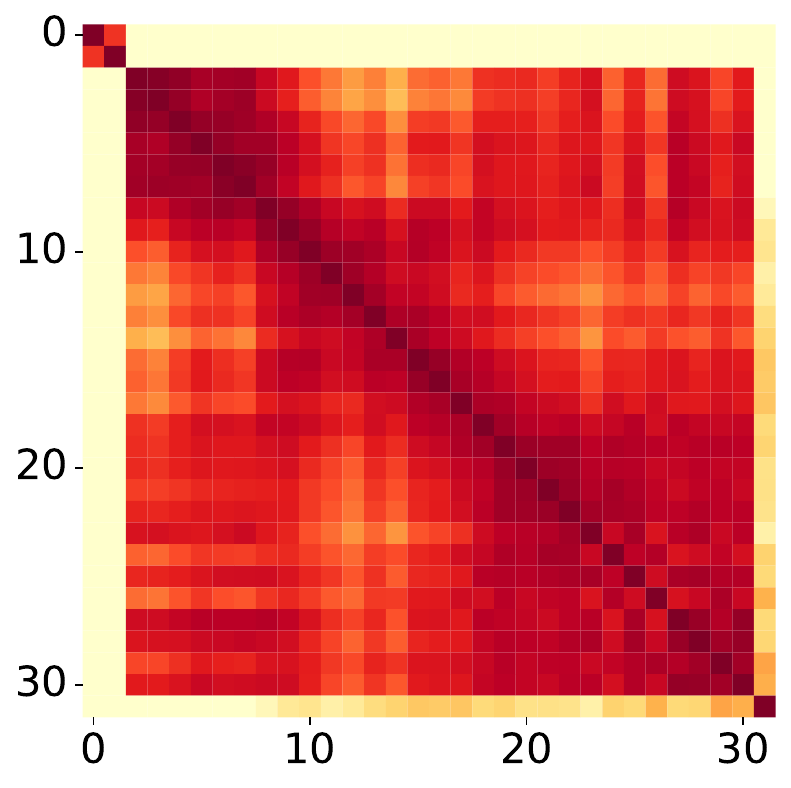}
\hfill
\includegraphics[width = 0.27\textwidth]{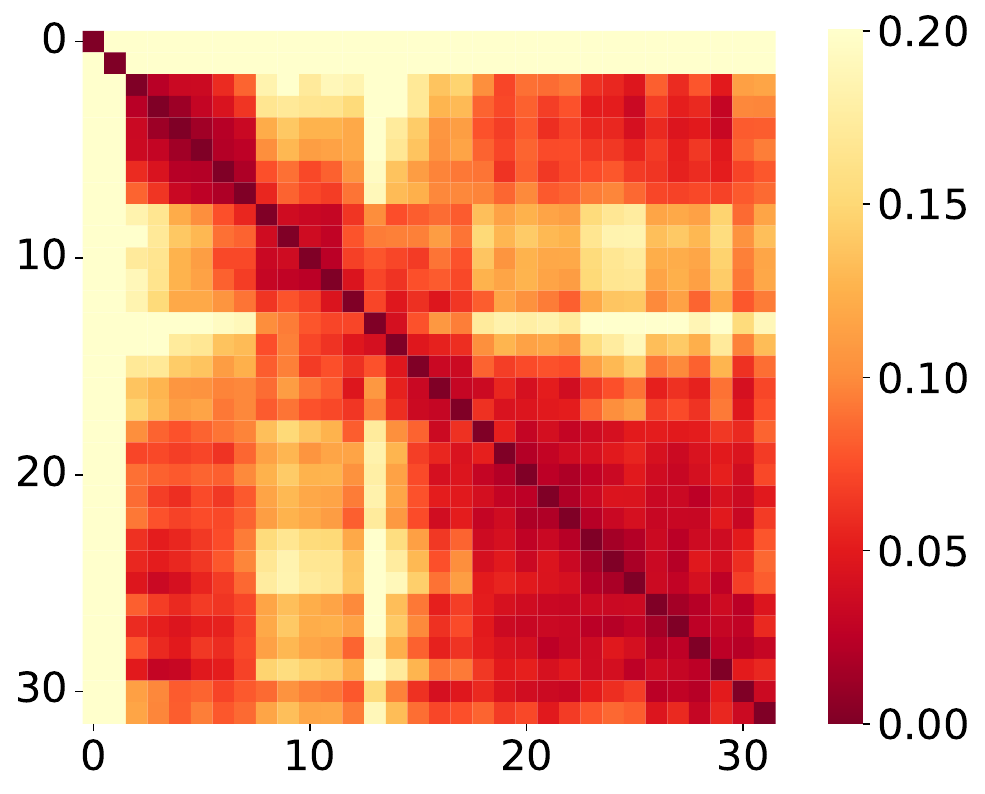}
\subfloat[Physical Commonsense QA]{\includegraphics[width = 0.24\textwidth]{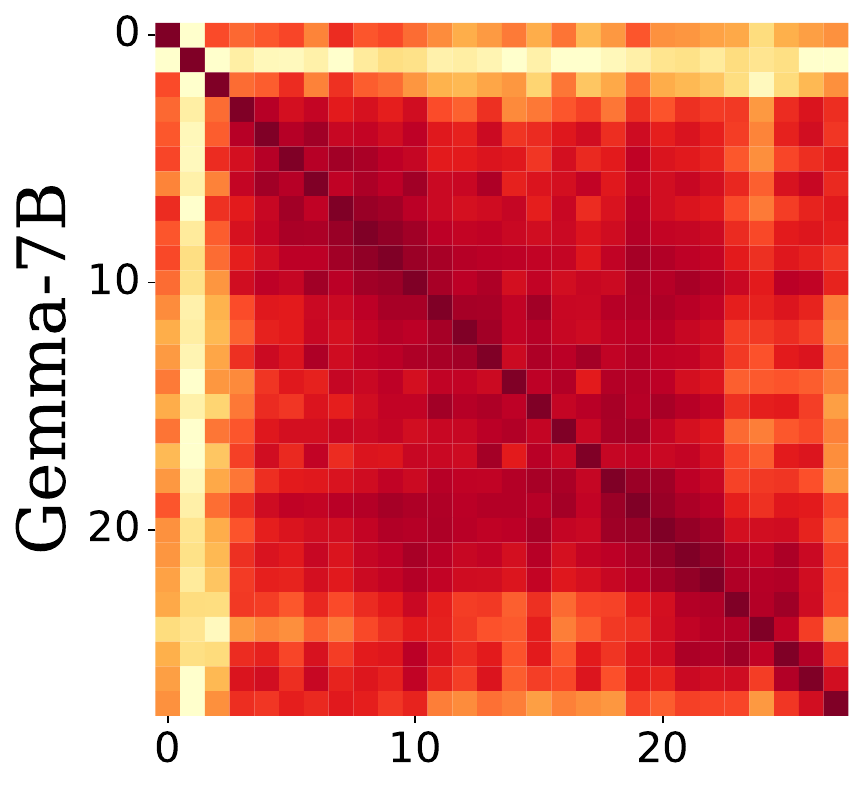}}
\hfill
\subfloat[Short Sentence Translation]{\includegraphics[width = 0.22\textwidth]{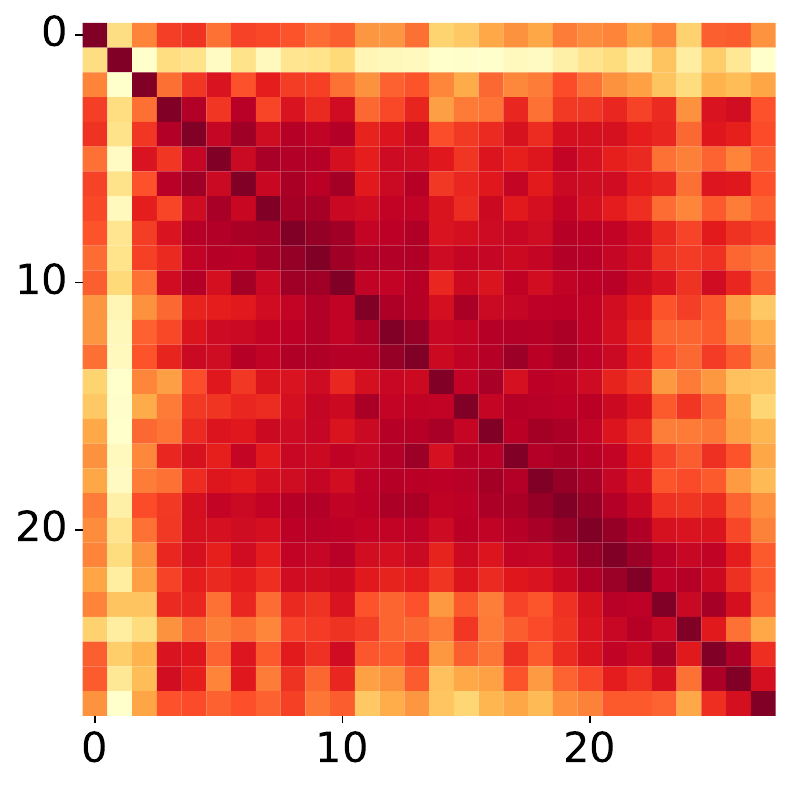}}
\hfill
\subfloat[Coreference Resolution]{\includegraphics[width = 0.22\textwidth]{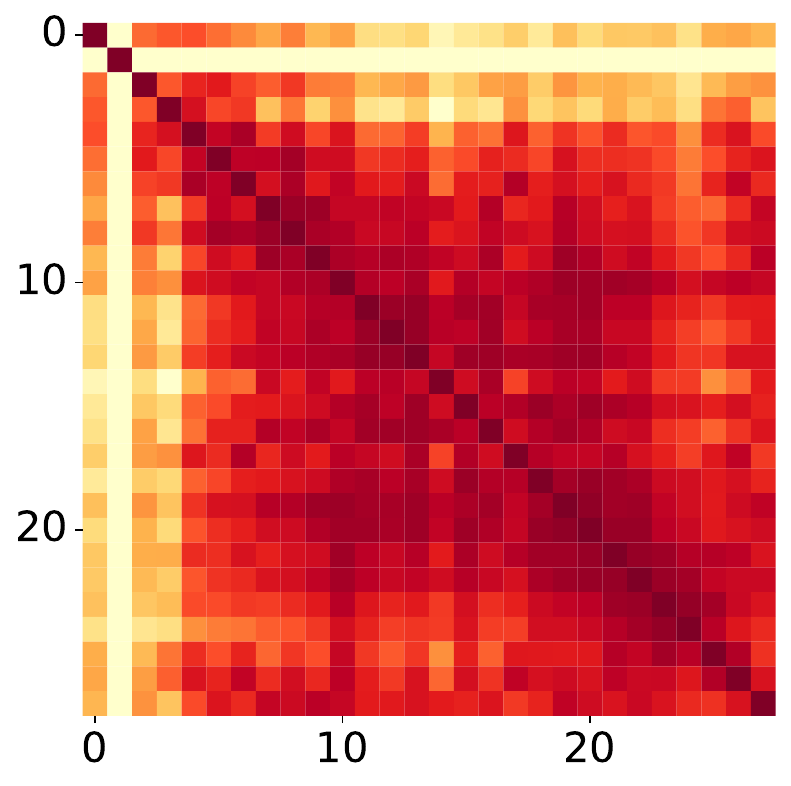}}
\hfill
\subfloat[Mathematical Reasoning]{\includegraphics[width = 0.27\textwidth]{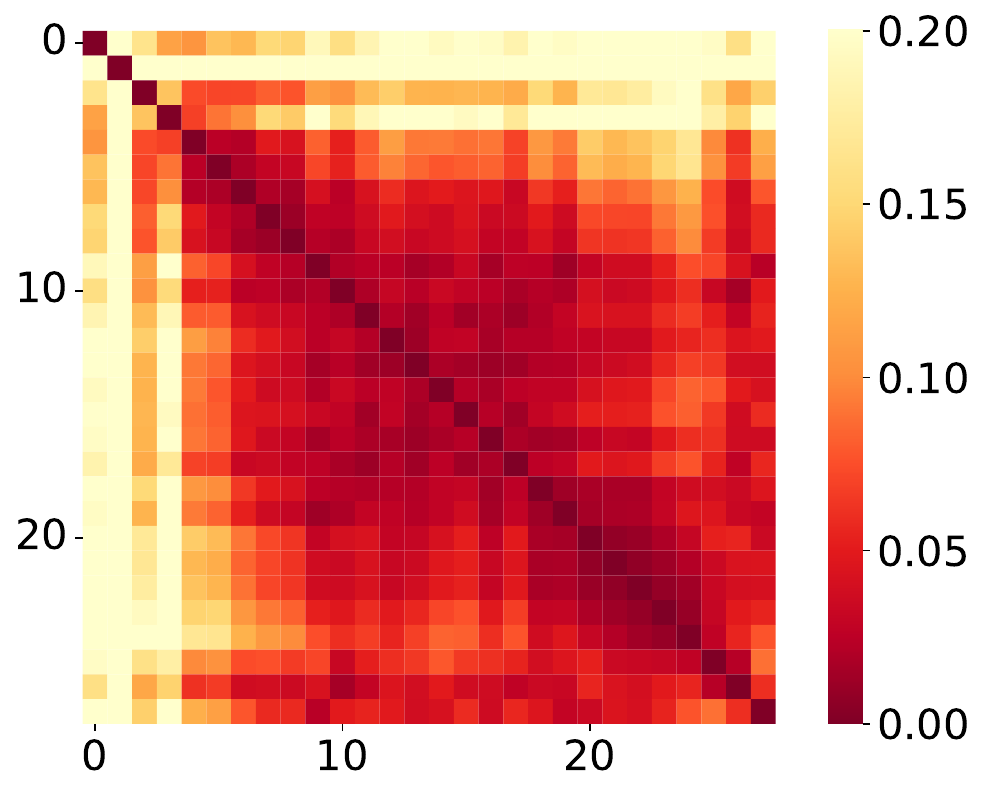}}
\caption{The JS divergence scores of the attention weights for every pair of layers (calculated under setting S1). The greater the redness, the higher the similarity.}
\label{app:heatmap2}
\end{figure*}

\begin{figure*}
    \centering
\begin{tikzpicture}[scale=1.0]
\begin{axis}[
   at={(0,0)},
   x=1.4em,
   y=5em,
   ymajorgrids=true,
   xmajorgrids=true,
   grid style={draw=black!10, dashed},
   xtick pos=left,
   legend entries={ Layer 1, Layer 3 },
   ymax=0.9,
   xmin=-1,xmax=26,
   ytick={0.0,0.4,0.8},
   xlabel={Token Index},
   legend style={
      fill,
      at={(0.98,3.5em)},
      legend columns=2,
      legend cell align=left,
      anchor=east
    },
   yticklabel style={/pgf/number format/fixed,/pgf/number format/fixed zerofill,/pgf/number format/precision=1},
   xlabel style={anchor=center, xshift=0em, yshift=-0.8em},
   title=(a) The corresponding JS divergence score is $0.3685$., every axis title/.style={at={(19em,-4em)}}
   ]              

\addplot[very thick,draw=color7,color=color7] plot coordinates {
(0,0.04333423)
(1,0.0241181)
(2,0.01158958)
(3,0.00844483)
(4,0.03378611)
(5,0.10672702)
(6,0.00988921)
(7,0.0373321)
(8,0.0119416)
(9,0.00764358)
(10,0.00803415)
(11,0.01697957)
(12,0.01182351)
(13,0.16863948)
(14,0.03148856)
(15,0.04666872)
(16,0.01260793)
(17,0.01684203)
(18,0.01312212)
(19,0.0094758)
(20,0.01288119)
(21,0.06644725)
(22,0.0120649)
(23,0.01874584)
(24,0.02856018)
(25,0.23081239)
};
\addplot[very thick,draw=color1,color=color1] plot coordinates {
(0,0.79458594)
(1,0.00365488)
(2,0.00364143)
(3,0.00461985)
(4,0.003868)
(5,0.01012645)
(6,0.0061455)
(7,0.0111398)
(8,0.00177438)
(9,0.00504788)
(10,0.00207178)
(11,0.00349443)
(12,0.00382762)
(13,0.02091016)
(14,0.02943448)
(15,0.02425419)
(16,0.00794599)
(17,0.00675072)
(18,0.00714393)
(19,0.00351513)
(20,0.00382834)
(21,0.00763459)
(22,0.00955388)
(23,0.0064291)
(24,0.00596229)
(25,0.01263922)
};
\end{axis}
\begin{axis}[
   at={(0,-10em)},
   x=1.4em,
   y=5em,   
   xmin=-1,xmax=26,
   ytick={0.0,0.4,0.8},
   ymajorgrids=true,
   xmajorgrids=true,
   grid style={draw=black!10, dashed},
   xtick pos=left,
   legend entries={ Layer 6, Layer 13},
   ymax=0.9,
   xmin=-1,xmax=26,
   ytick={0.0,0.4,0.8},
   ylabel={Attention Probability Distribution},
   legend style={
      fill,
      at={(0.98,3.5em)},
      legend columns=2,
      legend cell align=left,
      anchor=east
    },
   yticklabel style={/pgf/number format/fixed,/pgf/number format/fixed zerofill,/pgf/number format/precision=1},
   xlabel={Token Index},
   xlabel style={anchor=center, xshift=0em, yshift=-0.8em},
   title=(b) The corresponding JS divergence score is $0.0333$., every axis title/.style={at={(19em,-4em)}}
   ]              

\addplot[very thick,draw=color7,color=color7] plot coordinates {
(0,0.57800567)
(1,0.00626915)
(2,0.00692314)
(3,0.00729354)
(4,0.01379545)
(5,0.01656018)
(6,0.00506486)
(7,0.0102253)
(8,0.00578242)
(9,0.0076143)
(10,0.00354935)
(11,0.00658913)
(12,0.01477851)
(13,0.06616887)
(14,0.02725535)
(15,0.05496554)
(16,0.0157124)
(17,0.00855429)
(18,0.01386539)
(19,0.00484728)
(20,0.01092117)
(21,0.0252343)
(22,0.0109332)
(23,0.01669364)
(24,0.01896309)
(25,0.04343446)
};
\addplot[very thick,draw=color1,color=color1] plot coordinates {
(0,0.34873909)
(1,0.01620835)
(2,0.00806497)
(3,0.01044355)
(4,0.02361915)
(5,0.02840216)
(6,0.01536767)
(7,0.03128122)
(8,0.01391755)
(9,0.01290382)
(10,0.00634051)
(11,0.01369789)
(12,0.01650783)
(13,0.0646522)
(14,0.04711771)
(15,0.06770527)
(16,0.01588366)
(17,0.0170988)
(18,0.01483296)
(19,0.01118835)
(20,0.02174012)
(21,0.03129602)
(22,0.01969075)
(23,0.01982701)
(24,0.03529382)
(25,0.08817956)
};
\end{axis}
\begin{axis}[
   at={(0,-20em)},
   x=1.4em,
   y=5em,
   ymajorgrids=true,
   xmajorgrids=true,
   grid style={draw=black!10, dashed},
   xtick pos=left,
   legend entries={Layer 22, Layer 23},
   ymax=0.9,
   xmin=-1,xmax=26,
   ytick={0.0,0.4,0.8},
   legend style={
      fill,
      at={(0.98,3.5em)},
      legend columns=2,
      legend cell align=left,
      anchor=east
    },
   yticklabel style={/pgf/number format/fixed,/pgf/number format/fixed zerofill,/pgf/number format/precision=1},
   xlabel={Token Index},
   xlabel style={anchor=center, xshift=0em, yshift=-0.8em},
   title=(c) The corresponding JS divergence score is $0.0036$., every axis title/.style={at={(19em,-4em)}}
   ]

\addplot[very thick,draw=color7,color=color7] plot coordinates {
(0,0.70274949)
(1,0.01420072)
(2,0.00461283)
(3,0.00581601)
(4,0.0100854)
(5,0.01553237)
(6,0.00962346)
(7,0.03607196)
(8,0.00347256)
(9,0.00359858)
(10,0.00115282)
(11,0.0012078)
(12,0.00198127)
(13,0.03535202)
(14,0.03024685)
(15,0.00981544)
(16,0.00279651)
(17,0.00131053)
(18,0.00225326)
(19,0.00119111)
(20,0.00265124)
(21,0.00212572)
(22,0.00205437)
(23,0.00398802)
(24,0.00925512)
(25,0.08685457)
};
\addplot[very thick,draw=color1,color=color1] plot coordinates {
(0,0.65680921)
(1,0.0157594)
(2,0.00519477)
(3,0.01038994)
(4,0.01415171)
(5,0.02734223)
(6,0.01472745)
(7,0.03805634)
(8,0.00393894)
(9,0.00540681)
(10,0.0014547)
(11,0.00183819)
(12,0.00374719)
(13,0.03018799)
(14,0.03647182)
(15,0.01761552)
(16,0.00301028)
(17,0.00183743)
(18,0.00152287)
(19,0.00071919)
(20,0.00272236)
(21,0.00248076)
(22,0.00253787)
(23,0.00237141)
(24,0.00586467)
(25,0.09384093)
};
\end{axis}
\end{tikzpicture}
    \caption{A visualization of the attention probability distribution in two layers. We also report the corresponding JS divergence score. The horizontal coordinates stand for tokens with different positions.}
    \label{fig:visualization_of_the_attention_probability_distribution}
\end{figure*}

\begin{figure*}
    \centering
    \begin{tikzpicture}[scale=0.5]
\tikzstyle{every node}=[font=\Large]

\node [rectangle,draw=black,minimum height=1.0em,minimum width=33.88em,font=\small,anchor=west,align=center] (final) at (-0.03em,17em){};

\node [anchor=center,color=black!50] at (2em,17.0em){\pgfuseplotmark{*}};
\node [anchor=west,scale=0.9] at (2.5em,17.0em){\scriptsize Poor Perf.};

\node [anchor=center,scale=1.5,color=black!50] at (10.0em,16.9em){\pgfuseplotmark{triangle*}};
\node [anchor=west,scale=0.9] at (10.5em,16.9em){\scriptsize Satisfactory Perf.};

\draw [very thick, color=color2]  (21.0em,17.0em) -- (24.0em,17.0em) coordinate(a);
\node [anchor=west,scale=0.9] at (24em,17.0em){\scriptsize Standard Attention};

\draw [very thick, color=color5]  (35.0em,17.0em) -- (38.0em,17.0em) coordinate(a);
\node [anchor=west,scale=0.9] at (38em,17.0em){\scriptsize Average Attention};

\draw [very thick, color=color7]  (49.0em,17.0em) -- (52.0em,17.0em) coordinate(a);
\node [anchor=west,scale=0.9] at (52em,17.0em){\scriptsize Directly Sharing Attention};

\begin{axis}[
   at={(0,0)},
   ymajorgrids=true,
   xmajorgrids=true,
   grid style={draw=black!20, dashed},
   xtick pos=left,
   xmin=0,xmax=35,
   xlabel={\LARGE Layer Index},
   ylabel={\LARGE Accuracy},
   xlabel style={anchor=center, xshift=0em, yshift=-1.8em},
   title=\LARGE(a) PIQA, every axis title/.style={at={(9em,-8em)}}
   ]
\addplot[line width=2pt,color=color5] plot coordinates {
(3,50.71)
(5,57.4)
(7,60.99)
(9,71.49)
(11,75.95)
(13,75.63)
(15,73.94)
(17,73.07)
(19,73.61)
(21,75.63)
(23,78.07)
(25,77.42)
(27,77.58)
(29,77.86)
(31,73.18)
}; 
\addplot[only marks,mark=*,mark size=3pt,draw=color5,color=color5] plot coordinates {
(3,50.71)
(5,57.4)
(7,60.99)
(9,71.49)
(11,75.95)
(13,75.63)
(15,73.94)
(17,73.07)
(19,73.61)
(31,73.18)
};
\addplot[only marks,mark=triangle*,mark size=5pt, color=color5,color=color5] plot coordinates {
(21,75.63)
(23,78.07)
(25,77.42)
(27,77.58)
(29,77.86)
};
\addplot[line width=2pt,color=color7] plot coordinates {
(3,49.78)
(5,77.75)
(7,77.09)
(9,77.48)
(11,78.24)
(13,77.2)
(15,78.13)
(17,77.15)
(19,79.98)
(21,79.27)
(23,80.09)
(25,80.63)
(27,80.25)
(29,80.41)
(31,77.53)
};
\addplot[only marks,mark=*,mark size=3pt,draw=color7,color=color7] plot coordinates {
(3,49.78)
};
\addplot[only marks,mark=triangle*,mark size=5pt, color=color7,color=color7] plot coordinates {
(5,77.75)
(7,77.09)
(9,77.48)
(11,78.24)
(13,77.2)
(15,78.13)
(17,77.15)
(19,79.98)
(21,79.27)
(23,80.09)
(25,80.63)
(27,80.25)
(29,80.41)
(31,77.53)
};
\addplot[line width=2pt,draw=color2!80,color=color2!80] plot coordinates {
(0,79.11)
(35,79.11)
};

\end{axis}

\begin{axis}[
   at={(25em,0)},
   ymajorgrids=true,
   xmajorgrids=true,
   grid style={draw=black!20, dashed},
   xtick pos=left,
   xmin=0,xmax=35,
   xlabel={\LARGE Layer Index},
   ylabel={\LARGE Accuracy},
   xlabel style={anchor=center, xshift=0em, yshift=-1.8em},
   title=\LARGE(b) MMLU, every axis title/.style={at={(9em,-8em)}}
   ]
\addplot[line width=2pt,color=color5] plot coordinates {
(3,0)
(5,8.13)
(7,25.55)
(9,24.88)
(11,26.06)
(13,27.22)
(15,30.74)
(17,44.17)
(19,42.62)
(21,42.95)
(23,43.06)
(25,39.45)
(27,43.96)
(29,46.13)
(31,45.51)
};
\addplot[only marks,mark=*,mark size=3pt,draw=color5,color=color5] plot coordinates {
(3,0)
(5,8.13)
(7,25.55)
(9,24.88)
(11,26.06)
(13,27.22)
(15,30.74)
};
\addplot[only marks,mark=triangle*,mark size=5pt, color=color5,color=color5] plot coordinates {
(17,44.17)
(19,42.62)
(21,42.95)
(23,43.06)
(25,39.45)
(27,43.96)
(29,46.13)
(31,45.51)
};
\addplot[line width=2pt,color=color7] plot coordinates {
(3,0.06)
(5,28.49)
(7,25.26)
(9,30.25)
(11,28.22)
(13,33.64)
(15,37.49)
(17,46.32)
(19,45.56)
(21,45.79)
(23,45.5)
(25,45.88)
(27,45.77)
(29,45.66)
(31,45.56)
};
\addplot[only marks,mark=*,mark size=3pt,draw=color7,color=color7] plot coordinates {
(3,0.06)
(5,28.49)
(7,25.26)
(9,30.25)
(11,28.22)
(13,33.64)
(15,37.49)
};
\addplot[only marks,mark=triangle*,mark size=5pt, color=color7,color=color7] plot coordinates {
(17,46.32)
(19,45.56)
(21,45.79)
(23,45.5)
(25,45.88)
(27,45.77)
(29,45.66)
(31,45.56)
};
\addplot[line width=2pt,draw=color2!80,color=color2!80] plot coordinates {
(0,45.77)
(35,45.77)
};

\end{axis}

\begin{axis}[
   at={(50em,0em)},
   ymajorgrids=true,
   xmajorgrids=true,
   grid style={draw=black!20, dashed},
   xtick pos=left,
   xmin=0,xmax=35,
   xlabel={\LARGE Layer Index},
   ylabel={\LARGE Exact Match},
   xlabel style={anchor=center, xshift=0em, yshift=-1.8em},
   title=\LARGE(c) GSM8K, every axis title/.style={at={(9em,-8em)}}
   ]

\addplot[line width=2pt,color=color5] plot coordinates {
(3,1.52)
(5,0.53)
(7,0.61)
(9,1.74)
(11,1.52)
(13,2.05)
(15,1.59)
(17,3.26)
(19,3.41)
(21,7.43)
(23,7.43)
(25,9.93)
(27,12.28)
(29,12.05)
(31,11.45)
};
\addplot[only marks,mark=*,mark size=3pt,draw=color5,color=color5] plot coordinates {
(3,1.52)
(5,0.53)
(7,0.61)
(9,1.74)
(11,1.52)
(13,2.05)
(15,1.59)
(17,3.26)
(19,3.41)
(21,7.43)
(23,7.43)
(25,9.93)
};
\addplot[only marks,mark=triangle*,mark size=5pt, color=color5,color=color5] plot coordinates {
(27,12.28)
(29,12.05)
(31,11.45)
};
\addplot[line width=2pt,color=color7] plot coordinates {
(3,2.5)
(5,2.88)
(7,1.59)
(9,1.44)
(11,1.44)
(13,1.9)
(15,3.11)
(17,5.38)
(19,8.79)
(21,13.27)
(23,13.12)
(25,14.25)
(27,13.8)
(29,14.56)
(31,7.81)
};
\addplot[only marks,mark=*,mark size=3pt,draw=color7,color=color7] plot coordinates {
(3,2.5)
(5,2.88)
(7,1.59)
(9,1.44)
(11,1.44)
(13,1.9)
(15,3.11)
(17,5.38)
(19,8.79)
(31,7.81)
};
\addplot[only marks,mark=triangle*,mark size=5pt, color=color7,color=color7] plot coordinates {
(21,13.27)
(23,13.12)
(25,14.25)
(27,13.8)
(29,14.56)

};
\addplot[line width=2pt,draw=color2!80,color=color2!80] plot coordinates {
(0,14.71)
(35,14.71)
};

\end{axis}

\end{tikzpicture}
    \caption{The performance of LLaMA2-7B when applying the average and directly sharing attention strategies in every two adjacent layers.}
    \label{fig:trend_of_sensitivity_of_LLaMA2_7B}
\end{figure*}

\section{Extension for Section \ref{sec:Layer-wise_Similarity_of_Attention_Weights}}

\subsection{Additional Evidence Supporting Attention Similarity}
\label{app:Additional_Evidence_Supporting_Attention_Similarity}

\paragraph{Results on more models.} Figure \ref{app:heatmap2} shows the extended results of Figure \ref{fig:heatmap}, i.e., the JS divergence score of LLaMA2-7B and Gemma-7B. It suggests that highly similar attention patterns persist across most layers of different large language models.

\paragraph{Detailed results of the top-5 most attended tokens.} To provide a more quantitative assessment, Table \ref{app_tab:topk_tokens} presents detailed results for the top-5 tokens with the highest attention weights across the 32 layers of LLaMA3-8B. The first two columns (``1'' and ``1024'') indicate that at generation steps 1 and 1024, the attention weights in adjacent layers are highly similar. Moreover, the ``PIQA'' and ``GSM8K'' columns show that while most layers exhibit similar attention patterns, layers 11 to 17 display more divergent behavior, which corresponds to the white cross lines observed in Figure \ref{fig:heatmap}.

\paragraph{JS divergence score instances.} Figure \ref{fig:visualization_of_the_attention_probability_distribution} shows instances of two attention probability distributions and their corresponding JS divergence scores. Our results indicate that the attention probability distributions are highly similar when the JS divergence falls below $0.05$.

\paragraph{JS divergence score baseline.} To facilitate a more objective comparison of JS divergence scores, we introduce a baseline that compares attention distributions across different sentences of equal length. As shown in Figure \ref{fig:heap_map_rm0_baseline} (b), the attention weights differ significantly between the two unrelated sentences, highlighting the high similarity observed in adjacent layers.

\section{Extension for Section \ref{sec:Sensitivity to Attention Weights}}
Figure \ref{fig:trend_of_sensitivity_of_LLaMA2_7B} shows the performance of LLaMA2-7B when applying the average and directly sharing attention strategies in every two adjacent layers.

\section{Extension for Section \ref{sec:Reducing_the_Inter-layer_Redundancy}}

\subsection{Training Setups}
\label{app:training_setups}

\begin{table*}[htb]
\Large
  \centering
  
 \resizebox{1.0\linewidth}{!}{
    \begin{tabular}{lcccccccccccc}
    \toprule
    \textbf{Benchmark} & \cellcolor{color6}\textbf{BoolQ} & \cellcolor{color6}\textbf{WinoGrande} & \cellcolor{color6}\textbf{PIQA} & \cellcolor{color6}\textbf{OBQA} & \cellcolor{color6}\textbf{HellaSwag}  & \cellcolor{color3}\textbf{ARC-E} & \cellcolor{color3}\textbf{ARC-C} & \cellcolor{color3}\textbf{MMLU} & \cellcolor{red!40}\textbf{CoQA} & \cellcolor{red!40}\textbf{NQ} & \cellcolor{red!40}\textbf{GSM8K} & \cellcolor{red!40}\textbf{TriviaQA} \\
    \midrule
    LLaMA3-8B & 81.13 & 73.40 & 80.69 & 46.60 & 82.26 & 77.61 & 59.30 & 64.98 & 67.40 & 29.14 & 51.71 & 63.39 \\
    DS (17) & 75.72 & 65.19 & 68.61 & 30.20 & 50.00 & 41.04 & 29.86 & 23.96 & 12.67 & \enspace 1.11 & \enspace 1.74 & \enspace 0.73 \\
    \enspace \enspace \textit{\textbf{Perf. Loss\dag}} & \textit{\textbf{6.67\%}} & \textit{\textbf{11.19\%}} & \textit{\textbf{14.97\%}} & \textit{\textbf{35.19\%}} & \textit{\textbf{39.22\%}} & \textit{\textbf{47.12\%}} & \textit{\textbf{49.65\%}} & \textit{\textbf{63.13\%}} & \textit{\textbf{81.20\%}} & \textit{\textbf{96.19\%}} & \textit{\textbf{96.64\%}} & \textit{\textbf{98.85\%}} \\
    \textsc{LiSA} (17) & 81.65 & 73.95 & 79.87 & 46.20 & 81.17 & 79.29 & 58.96 & 61.22 & 63.53 & 27.17 & 45.94 & 57.66 \\
    \enspace \enspace \textit{\textbf{Perf. Loss}} & \textit{\textbf{-0.64\%}} & \textit{\textbf{-0.75\%}} & \textit{\textbf{1.02\%}} & \textit{\textbf{0.86\%}} & \textit{\textbf{1.33\%}} & \textit{\textbf{-2.16\%}} & \textit{\textbf{0.57\%}} & \textit{\textbf{5.79\%}} & \textit{\textbf{5.74\%}} & \textit{\textbf{6.76\%}} & \textit{\textbf{11.16\%}} & \textit{\textbf{9.04\%}} \\
    \midrule
    \textbf{Benchmark} & \cellcolor{color6}\textbf{WinoGrande} & \cellcolor{color6}\textbf{PIQA} & \cellcolor{color3}\textbf{OBQA} & \cellcolor{color3}\textbf{BoolQ} & \cellcolor{color3}\textbf{ARC-E} & \cellcolor{color3}\textbf{ARC-C} & \cellcolor{color3}\textbf{HellaSwag} & \cellcolor{red!40}\textbf{GSM8K} & \cellcolor{red!40}\textbf{TriviaQA} & \cellcolor{red!40}\textbf{NQ} & \cellcolor{red!40}\textbf{MMLU} & \cellcolor{red!40}\textbf{CoQA} \\
    \midrule
    LLaMA3-8B & 73.40 & 80.69 & 46.60 & 81.13 & 77.61 & 59.30 & 82.26 & 51.71 & 63.39 & 29.14 & 64.98 & 67.40 \\
    DS (27) & 51.54 & 56.58 & 25.40 & 38.07 & 30.64 & 22.78 & 28.31 & \enspace 2.12 & \enspace 0.04 & \enspace 0.03 & \enspace 0.00 & \enspace 0.00 \\
    \enspace \enspace \textit{\textbf{Perf. Loss\dag}} & \textit{\textbf{29.78\%}} & \textit{\textbf{29.88\%}} & \textit{\textbf{45.49\%}} & \textit{\textbf{53.08\%}} & \textit{\textbf{60.52\%}} & \textit{\textbf{61.59\%}} & \textit{\textbf{65.58\%}} & \textit{\textbf{95.90\%}} & \textit{\textbf{99.94\%}} & \textit{\textbf{99.90\%}} & \textit{\textbf{100.00\%}} & \textit{\textbf{100.00\%}} \\
    \textsc{LiSA} (27) & 70.17 & 80.69 & 46.80 & 77.86 & 74.92 & 53.33 & 79.43 & 31.77 & 43.65 & 25.65 & 50.58 & 60.23 \\
    \enspace \enspace \textit{\textbf{Perf. Loss}} & \textit{\textbf{4.40\%}} & \textit{\textbf{0.00\%}} & \textit{\textbf{-0.43\%}} & \textit{\textbf{4.03\%}} & \textit{\textbf{3.47\%}} & \textit{\textbf{10.07\%}} & \textit{\textbf{3.44\%}} & \textit{\textbf{38.56\%}} & \textit{\textbf{31.14\%}} & \textit{\textbf{11.98\%}} & \textit{\textbf{22.16\%}} & \textit{\textbf{10.64\%}} \\
    \bottomrule
    \end{tabular}
}
  \caption{The performance loss of DS and \textsc{LiSA} on LLaMA3-8B. We report the performance loss as a percentage in the ``Perf. Loss'' lines. The symbol \dag represents that the 12 benchmarks are ordered by increasing the performance loss of DS models, from left to right. Green indicates that this is a low-loss task, with the performance loss of DS models below $40\%$. While yellow denotes that it is a moderate-loss task, with the performance loss of DS models in the range of $40\%$ to $70\%$. Additionally, red stands for high-loss tasks, with the performance loss of DS models above $70\%$.}
  \label{app_tab:trade_off}
\end{table*}

\paragraph{Detailed \textsc{LiSA} Configuration} Both the LLaMA2-7B and LLaMA3-8B models are equipped with 32 attention heads ($h=32$) per layer and have hidden state dimensions of $d=4096$. While LLaMA2-13B consists of 40 attention heads ($h=40$) per layer and has hidden state dimensions of $d=5120$.
\begin{itemize}
    \item When a layer is equipped with \textsc{LiSA}, it uses a two-layer FFN, which involves a $64 \times 256$ FFN, a \texttt{ReLU} activation function, and a $256 \times 32$ FFN. Additionally, \textsc{LiSA} includes two low-rank linear projections: for the LLaMA3-8B model, these projections are $W_{LR}^Q\in\mathbb{R}^{4096 \times 640}$ and $W_{LR}^K\in\mathbb{R}^{4096 \times 160}$; for the LLaMA2-7B, $W_{LR}^Q, W_{LR}^K\in\mathbb{R}^{4096 \times 640}$; and for the LLaMA2-13B, $W_{LR}^Q, W_{LR}^K\in\mathbb{R}^{5120 \times 800}$ each.
    \item Besides, \textsc{LiSA}$_{SL}$ uses a one-layer FFN sized $64 \times 32$, paired with two low-rank linear projections: for LLaMA3-8B, $W{LR}^Q\in\mathbb{R}^{4096 \times 1024}$ and $W_{LR}^K\in\mathbb{R}^{4096 \times 256}$; for LLaMA2-7B, both projections are $W_{LR}^Q, W_{LR}^K\in\mathbb{R}^{4096 \times 1024}$; and for LLaMA2-13B, $W_{LR}^Q, W_{LR}^K\in\mathbb{R}^{5120 \times 1280}$ each.
\end{itemize}

\paragraph{Huber loss function.} The standard function of Huber loss \cite{huber1992robust} can be expressed as follows:
\begin{align}
\mathcal{L}_\delta (a, b) = \begin{cases} 
\frac{1}{2} (a - b)^2 & \text{if } |a - b| \leq \delta \\
\delta (|a - b| - \frac{1}{2} \delta) & \text{otherwise}
\end{cases}
\end{align}
where $\delta$ is always set to $1$ in our experiments. Indeed, it is a combination of \textit{mean absolute error (MAE)} and \textit{mean squared error (MSE)} loss, which can make the training process more robust.

\paragraph{Datasets.} Since the trainable parameters introduced by \textsc{LiSA}, only account for $0.46\%$ to $1.64\%$ of the total parameters, we do not need a large training dataset. To obtain high-quality pre-training data, we applied different sampling proportions to subsets of \texttt{RedPajama-Data-1T} \cite{together_2023_redpajama}, including $10\%$ of ArXiv, $2\%$ of C4, $100\%$ of StackExchange, $100\%$ of Wikipedia, and $10\%$ of GitHub. The resulting dataset contains 20 billion tokens and we sampled 4.2 and 10 billion tokens from this dataset for the experiments of uptraining and pre-training from scratch.

\paragraph{Main experiment.} We trained all models using the \texttt{LLaMA-Factory}\footnote{https://github.com/hiyouga/LLaMA-Factory} package \cite{zheng2024llamafactory}. During the pre-training stage, we set the global batch size to $128$, $\beta$ to $0.25$, weight decay to $0.1$, number of training epochs to $1$, warmup steps to $1500$, maximum text length to $1024$, and the learning rate to $0.0003$. The training process consisted of 40,000 update steps. Additionally, we used DeepSpeed ZeRO-2 \cite{DBLP:journals/corr/abs-1910-02054}. All experiments were conducted on eight A800 GPUs.

\paragraph{Preliminary experiment.} To accelerate the training, we trained all models on 1 billion tokens from the \texttt{RedPajama-Data-1T-Sample} dataset. Other hyperparameters remain the same as in the main experiment, except for the global batch size, which is set to $16$.

\begin{figure}
\centering
\includegraphics[width=0.5\textwidth]{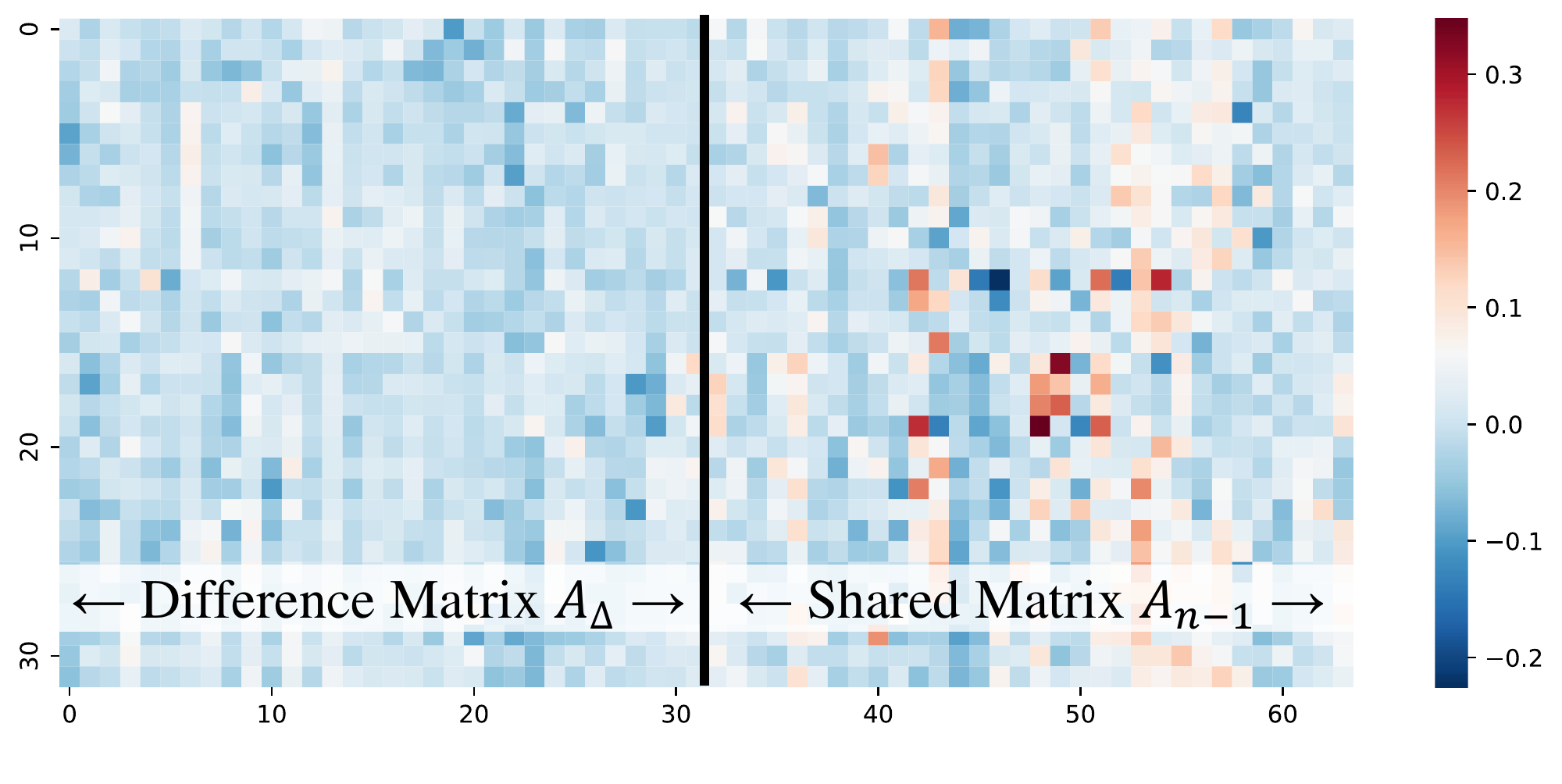}
\caption{A weight visualization of a well-trained single-layer FFN for aligning attention heads, whose shape is $64 \times 32$, i.e., $2h \times h$. Values in two square matrices represent the learned weights accounting for the difference matrix $A_\Delta$ and the shared attention weight matrix $A_{n-1}$, respectively.}
\label{fig:alignment_FFN}
\end{figure}

\subsection{Evaluation Setups}
\label{app:evaluation_setups}

\paragraph{Downstream tasks.} We used the \texttt{lm-evaluation-harness} package \cite{eval-harness} to evaluate the quality of outputs from different models. Except for the number of shots, which is set according to the configurations used by LLaMA2, LLaMA3, and \citet{DBLP:conf/iclr/XiaGZ024}, we kept other hyperparameters at their default settings in the \texttt{lm-evaluation-harness} package.

\paragraph{Supervised fine-tuning.} To evaluate the capabilities of following instructions, we first fine-tuned the models on the Alpaca dataset, which contains 52,000 instances. Then, we prompted the models to generate responses on the AlpacaEval \cite{alpaca_eval} data and leveraged GPT-4 (\texttt{gpt-4-0613}) to determine which of the two responses was better. Aligning with \citet{DBLP:conf/acl/WangZCLMXLZ24}, during the fine-tuning stage, we set the global batch size to $128$, weight decay to $0$, number of training epochs to $3$, warmup steps to $0$, maximal text length to $1024$, and the learning rate to $0.0001$. In the generation stage, the decoding temperature was set to $0.75$ and Top-p was set to $0.95$ to ensure the diversity of generated responses.

\begin{figure}
\centering
\includegraphics[width=0.5\textwidth]{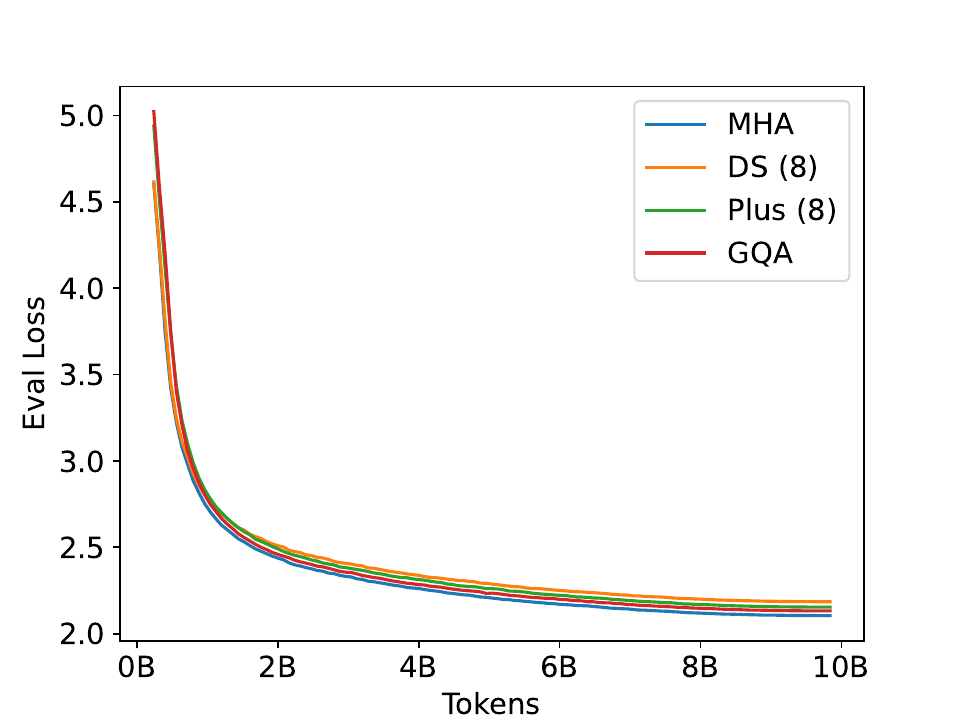}
\caption{Evaluation loss curves for pre-training LLaMA-like models with various attention mechanisms. The original model consists of $12$ layers, each with $12$ attention heads, and an attention head dimension of $64$. Plus (8) indicates that \textsc{LiSA}$_{plus}$ is applied to $8$ specific layers. The layer-wise sharing configuration for DS (8) and Plus (8) is ``2,3,4,6,7,8,10,11''. For the GQA model, we set the number of KV attention heads to 2.}
\label{fig:train_from_scratch_loss}
\end{figure}

\subsection{Experimental Results}
The experimental results presented in this section can be categorized as follows (aligning with the order in which they appear in the main text):

\paragraph{Visualization of the attention heads alignment module.} Figure \ref{fig:alignment_FFN} visualizes the weight of a well-trained single-layer FFN for aligning attention heads. Values in two square matrices represent the learned weights accounting for the difference matrix $A_\Delta$ and the shared attention weight matrix $A_{n-1}$, respectively. Cells with either deeper blue or deeper red indicate the larger absolute value of weights. We can see that FFN assigns significant weights to the $A_{n-1}$, which carries most of the attention information. Moreover, there are also some deep blue cells in the left matrix, demonstrating the necessity of $A_\Delta$.


\begin{table}[htp]
  \centering
  \LARGE
 \resizebox{0.65\linewidth}{!}{
    \begin{tabular}{lcc}
    \toprule
    \textbf{Model} & \textbf{GSM8K} & \textbf{MMLU}  \\
    \midrule
    \multicolumn{3}{c}{\textit{LLaMA3-8B}}  \\
    \midrule
    \textsc{LiSA} (17) & 45.94 & 61.22  \\
    \rowcolor{black!10}\quad+ NF & 47.31 & 61.22  \\
    \textsc{LiSA}$_{SL}$ (7+10) & 42.76 & 61.69  \\
    \rowcolor{black!10}\quad+ NF & 41.55 & 61.69  \\
    \textsc{LiSA} (21) & 39.27 & 59.52  \\
    \rowcolor{black!10}\quad+ NF & 42.99 & 59.52  \\
    \textsc{LiSA} (27) & 31.77 & 50.58  \\
    \rowcolor{black!10}\quad+ NF & 35.10 & 50.62  \\
    \midrule
    \multicolumn{3}{c}{\textit{LLaMA2-7B}}  \\
    \midrule
    \textsc{LiSA} (17) & 12.96 & 43.83  \\
    \rowcolor{black!10}\quad+ NF & 12.96 & 43.24  \\
    \textsc{LiSA}$_{SL}$ (7+10) & 8.26 & 42.05  \\
    \rowcolor{black!10}\quad+ NF & 7.96 & 43.18  \\
    \textsc{LiSA} (21) & 10.24 & 35.37  \\
    \rowcolor{black!10}\quad+ NF & 10.69 & 35.57  \\
    \bottomrule
    \end{tabular}
}

  \caption{Experiment of ablating the NF decoding strategy.}
  \label{app tab:normal_forward}
\end{table}

\paragraph{Normal forward.} Table \ref{app tab:normal_forward} compares the performance of enabling \textsc{LiSA} at all inference steps versus activating \textsc{LiSA} only after the first inference step, during which standard forward computation is used.


\paragraph{Pre-train from scratch.} Figure \ref{fig:train_from_scratch_loss} and Table \ref{app tab:performance_of_different_attention_models_pre_trained_from_scratch}  detail the training process and final performance of different attention models pre-trained from scratch.

\paragraph{Task-specific \textsc{LiSA}} Table \ref{app_tab:trade_off} summarizes the performance of DS and \textsc{LiSA}, as well as their respective degradation compared to the original model. Based on the performance drop introduced by DS, tasks can be categorized to guide the configuration of \textsc{LiSA}, enabling a better balance between inference efficiency and model effectiveness.


{
\setlength{\LTcapwidth}{\textwidth}
\onecolumn
\centering
\setlength{\tabcolsep}{3pt}
\begin{xltabular}{\textwidth}{
  c|*{3}{>{\centering\arraybackslash}X}|
    *{3}{>{\centering\arraybackslash}X}|
    *{3}{>{\centering\arraybackslash}X}|
    *{3}{>{\centering\arraybackslash}X}}
    \toprule
    \multirow{2.5}{*}{\textbf{Layer}} & \multicolumn{3}{c|}{\textbf{1}}  & \multicolumn{3}{c|}{\textbf{1024}} &  \multicolumn{3}{c|}{\textbf{PIQA}} &  \multicolumn{3}{c}{\textbf{GSM8K}} \\
    \cmidrule(lr){2-4}\cmidrule(lr){5-7}\cmidrule(lr){8-10}\cmidrule(lr){11-13}
     & Weight & Index & Token & Weight & Index & Token & Weight & Index & Token & Weight & Index & Token \\
    \midrule
    \endhead

    \midrule
    \multicolumn{13}{r}{{Continued on next page}} \\ 
    \bottomrule
    \endfoot

    \bottomrule 
    \multicolumn{2}{c}{}\\
    \caption{Top-5 tokens with the highest attention weights in each layer of LLaMA3-8B. ``1'' and ``1024'' refer to snapshots taken at generation steps 1 and 1024, respectively. ``PIQA'' and ``GSM8K'' show the results obtained by inputting a randomly selected sample from the PIQA and GSM8K benchmarks.}
    \label{app_tab:topk_tokens}\\
    \endlastfoot

    0 & 0.43  & 2 & a & 0.05  & 1025 & her & 0.17  & 36 & .$\backslash$n & 0.09  & 699 & ?$\backslash$n \\*
    0 & 0.19  & 3 & time & 0.05  & 1024 & in & 0.09  & 15 & ?$\backslash$n & 0.07  & 701 & : \\*
    0 & 0.14  & 4 & there & 0.03  & 1019 & . & 0.08  & 5 & $\backslash$n & 0.05  & 700 & A \\*
    0 & 0.13  & 0 & Once & 0.03  & 58 & , & 0.08  & 26 & , & 0.02  & 688 & . \\*
    0 & 0.11  & 1 & upon & 0.02  & 1023 & was & 0.07  & 22 & , & 0.02  & 694 & of \\*
    \midrule
    1 & 0.62  & 2 & a & 0.61  & 58 & , & 0.51  & 22 & , & 0.55  & 27 & , \\*
    1 & 0.19  & 1 & upon & 0.17  & 9 & . & 0.23  & 15 & ?$\backslash$n & 0.17  & 4 \\* 
    1 & 0.10  & 3 & time & 0.04  & 1025 & her & 0.03  & 36 & .$\backslash$n & 0.05  & 700 & A \\*
    1 & 0.06  & 4 & there & 0.02  & 1024 & in & 0.02  & 5 & $\backslash$n & 0.04  & 699 & ?$\backslash$n \\*
    1 & 0.04  & 0 & Once & 0.02  & 1026 & room & 0.02  & 4 & : & 0.02  & 701 & : \\*
    \midrule
    2 & 0.93  & 0 & Once & 0.79  & 0 & Once & 0.81  & 0 & The & 0.78  & 0 & Q \\*
    2 & 0.03  & 3 & time & 0.01  & 1024 & in & 0.02  & 16 & A & 0.02  & 699 & ?$\backslash$n \\*
    2 & 0.02  & 1 & upon & 0.01  & 1011 & and & 0.02  & 15 & ?$\backslash$n & 0.01  & 655 & Axel \\*
    2 & 0.01  & 2 & a & 0.01  & 1019 & . & 0.02  & 17 & : & 0.01  & 654 & : \\*
    2 & 0.01  & 4 & there & 0.01  & 1023 & was & 0.01  & 23 & drain & 0.01  & 600 & Olivia \\*
    \midrule
    3 & 0.89  & 0 & Once & 0.84  & 0 & Once & 0.76  & 0 & The & 0.80  & 0 & Q \\*
    3 & 0.06  & 3 & time & 0.02  & 1025 & her & 0.02  & 36 & .$\backslash$n & 0.02  & 701 & : \\*
    3 & 0.03  & 1 & upon & 0.02  & 1024 & in & 0.02  & 17 & : & 0.02  & 699 & ?$\backslash$n \\*
    3 & 0.02  & 4 & there & 0.01  & 1023 & was & 0.02  & 5 & $\backslash$n & 0.02  & 700 & A \\*
    3 & 0.01  & 2 & a & 0.01  & 1026 & room & 0.02  & 15 & ?$\backslash$n & 0.01  & 654 & : \\*
    \midrule
    4 & 0.90  & 0 & Once & 0.75  & 0 & Once & 0.66  & 0 & The & 0.72  & 0 & Q \\*
    4 & 0.05  & 3 & time & 0.03  & 1025 & her & 0.04  & 15 & ?$\backslash$n & 0.03  & 701 & : \\*
    4 & 0.03  & 4 & there & 0.02  & 1026 & room & 0.03  & 36 & .$\backslash$n & 0.03  & 699 & ?$\backslash$n \\*
    4 & 0.02  & 1 & upon & 0.02  & 1024 & in & 0.03  & 35 & cream & 0.02  & 700 & A \\*
    4 & 0.01  & 2 & a & 0.02  & 1022 & girl & 0.02  & 33 & and & 0.01  & 654 & : \\*
    \midrule
    5 & 0.90  & 0 & Once & 0.71  & 0 & Once & 0.62  & 0 & The & 0.70  & 0 & Q \\*
    5 & 0.04  & 1 & upon & 0.02  & 1024 & in & 0.05  & 15 & ?$\backslash$n & 0.03  & 699 & ?$\backslash$n \\*
    5 & 0.02  & 3 & time & 0.02  & 1025 & her & 0.04  & 17 & : & 0.02  & 701 & : \\*
    5 & 0.02  & 4 & there & 0.02  & 1019 & . & 0.03  & 36 & .$\backslash$n & 0.02  & 654 & : \\*
    5 & 0.01  & 2 & a & 0.02  & 1023 & was & 0.02  & 33 & and & 0.01  & 689 & What \\*
    \midrule
    6 & 0.88  & 0 & Once & 0.56  & 0 & Once & 0.53  & 0 & The & 0.62  & 0 & Q \\*
    6 & 0.05  & 3 & time & 0.06  & 1025 & her & 0.05  & 36 & .$\backslash$n & 0.04  & 701 & : \\*
    6 & 0.03  & 1 & upon & 0.03  & 1019 & . & 0.04  & 15 & ?$\backslash$n & 0.04  & 699 & ?$\backslash$n \\*
    6 & 0.03  & 4 & there & 0.03  & 1023 & was & 0.03  & 17 & : & 0.02  & 700 & A \\*
    6 & 0.02  & 2 & a & 0.03  & 1024 & in & 0.03  & 35 & cream & 0.01  & 688 & . \\*
    \midrule
    7 & 0.87  & 0 & Once & 0.66  & 0 & Once & 0.48  & 0 & The & 0.57  & 0 & Q \\*
    7 & 0.04  & 1 & upon & 0.03  & 1025 & her & 0.08  & 15 & ?$\backslash$n & 0.05  & 701 & : \\*
    7 & 0.04  & 4 & there & 0.02  & 1019 & . & 0.04  & 36 & .$\backslash$n & 0.04  & 699 & ?$\backslash$n \\*
    7 & 0.03  & 3 & time & 0.02  & 1023 & was & 0.03  & 17 & : & 0.03  & 654 & : \\*
    7 & 0.02  & 2 & a & 0.02  & 1024 & in & 0.03  & 35 & cream & 0.02  & 688 & . \\*
    \midrule
    8 & 0.83  & 0 & Once & 0.61  & 0 & Once & 0.46  & 0 & The & 0.30  & 0 & Q \\*
    8 & 0.07  & 1 & upon & 0.02  & 1026 & room & 0.07  & 36 & .$\backslash$n & 0.06  & 701 & : \\*
    8 & 0.04  & 4 & there & 0.02  & 1025 & her & 0.06  & 15 & ?$\backslash$n & 0.06  & 699 & ?$\backslash$n \\*
    8 & 0.03  & 3 & time & 0.01  & 1019 & . & 0.03  & 14 & cheese & 0.03  & 688 & . \\*
    8 & 0.03  & 2 & a & 0.01  & 1023 & was & 0.03  & 17 & : & 0.03  & 689 & What \\*
    \midrule
    9 & 0.87  & 0 & Once & 0.52  & 0 & Once & 0.41  & 0 & The & 0.30  & 0 & Q \\*
    9 & 0.05  & 1 & upon & 0.02  & 1025 & her & 0.06  & 36 & .$\backslash$n & 0.06  & 699 & ?$\backslash$n \\*
    9 & 0.03  & 3 & time & 0.02  & 1024 & in & 0.05  & 15 & ?$\backslash$n & 0.05  & 701 & : \\*
    9 & 0.03  & 2 & a & 0.02  & 1026 & room & 0.03  & 35 & cream & 0.04  & 688 & . \\*
    9 & 0.02  & 4 & there & 0.02  & 1023 & was & 0.03  & 4 & : & 0.03  & 689 & What \\*
    \midrule
    10 & 0.89  & 0 & Once & 0.59  & 0 & Once & 0.55  & 0 & The & 0.31  & 0 & Q \\*
    10 & 0.05  & 1 & upon & 0.03  & 1026 & room & 0.05  & 36 & .$\backslash$n & 0.05  & 701 & : \\*
    10 & 0.03  & 2 & a & 0.01  & 1024 & in & 0.04  & 15 & ?$\backslash$n & 0.04  & 699 & ?$\backslash$n \\*
    10 & 0.02  & 4 & there & 0.01  & 1025 & her & 0.03  & 4 & : & 0.03  & 688 & . \\*
    10 & 0.02  & 3 & time & 0.01  & 1019 & . & 0.03  & 35 & cream & 0.02  & 700 & A \\*
    \midrule
    11 & 0.88  & 0 & Once & 0.62  & 0 & Once & 0.49  & 0 & The & 0.30  & 0 & Q \\*
    11 & 0.06  & 1 & upon & 0.02  & 1019 & . & 0.06  & 15 & ?$\backslash$n & 0.06  & 701 & : \\*
    11 & 0.03  & 3 & time & 0.02  & 1024 & in & 0.04  & 36 & .$\backslash$n & 0.05  & 699 & ?$\backslash$n \\*
    11 & 0.02  & 2 & a & 0.01  & 1023 & was & 0.04  & 16 & A & 0.03  & 688 & . \\*
    11 & 0.01  & 4 & there & 0.01  & 1025 & her & 0.03  & 14 & cheese & 0.03  & 666 & . \\*
    \midrule
    12 & 0.83  & 0 & Once & 0.64  & 0 & Once & 0.30  & 0 & The & 0.36  & 0 & Q \\*
    12 & 0.07  & 1 & upon & 0.05  & 1026 & room & 0.08  & 36 & .$\backslash$n & 0.11  & 701 & : \\*
    12 & 0.05  & 3 & time & 0.03  & 1024 & in & 0.06  & 17 & : & 0.04  & 699 & ?$\backslash$n \\*
    12 & 0.03  & 4 & there & 0.03  & 1019 & . & 0.06  & 15 & ?$\backslash$n & 0.03  & 688 & . \\*
    12 & 0.02  & 2 & a & 0.03  & 1025 & her & 0.05  & 16 & A & 0.03  & 666 & . \\*
    \midrule
    13 & 0.87  & 0 & Once & 0.51  & 0 & Once & 0.40  & 0 & The & 0.17  & 0 & Q \\*
    13 & 0.06  & 1 & upon & 0.02  & 1026 & room & 0.07  & 16 & A & 0.08  & 666 & . \\*
    13 & 0.03  & 4 & there & 0.01  & 1019 & . & 0.06  & 15 & ?$\backslash$n & 0.06  & 701 & : \\*
    13 & 0.02  & 2 & a & 0.01  & 1024 & in & 0.04  & 36 & .$\backslash$n & 0.05  & 688 & . \\*
    13 & 0.02  & 3 & time & 0.01  & 1020 & The & 0.04  & 4 & : & 0.04  & 682 & and \\*
    \midrule
    14 & 0.85  & 0 & Once & 0.43  & 0 & Once & 0.37  & 0 & The & 0.24  & 0 & Q \\*
    14 & 0.06  & 3 & time & 0.04  & 1025 & her & 0.07  & 15 & ?$\backslash$n & 0.08  & 701 & : \\*
    14 & 0.04  & 1 & upon & 0.02  & 1019 & . & 0.07  & 16 & A & 0.07  & 666 & . \\*
    14 & 0.03  & 4 & there & 0.02  & 1024 & in & 0.05  & 36 & .$\backslash$n & 0.04  & 699 & ?$\backslash$n \\*
    14 & 0.02  & 2 & a & 0.02  & 1026 & room & 0.05  & 17 & : & 0.04  & 656 & has \\*
    \midrule
    15 & 0.75  & 0 & Once & 0.40  & 0 & Once & 0.37  & 0 & The & 0.32  & 0 & Q \\*
    15 & 0.10  & 3 & time & 0.04  & 1026 & room & 0.11  & 16 & A & 0.08  & 701 & : \\*
    15 & 0.07  & 1 & upon & 0.03  & 1024 & in & 0.06  & 36 & .$\backslash$n & 0.04  & 699 & ?$\backslash$n \\*
    15 & 0.05  & 4 & there & 0.02  & 1025 & her & 0.04  & 7 & : & 0.03  & 700 & A \\*
    15 & 0.03  & 2 & a & 0.02  & 1023 & was & 0.04  & 17 & : & 0.02  & 655 & Axel \\*
    \midrule
    16 & 0.74  & 0 & Once & 0.55  & 0 & Once & 0.44  & 0 & The & 0.40  & 0 & Q \\*
    16 & 0.11  & 3 & time & 0.04  & 1026 & room & 0.10  & 16 & A & 0.09  & 701 & : \\*
    16 & 0.07  & 4 & there & 0.03  & 1025 & her & 0.07  & 36 & .$\backslash$n & 0.05  & 655 & Axel \\*
    16 & 0.06  & 1 & upon & 0.03  & 1019 & . & 0.04  & 15 & ?$\backslash$n & 0.05  & 700 & A \\*
    16 & 0.03  & 2 & a & 0.02  & 1024 & in & 0.03  & 17 & : & 0.03  & 699 & ?$\backslash$n \\*
    \midrule
    17 & 0.80  & 0 & Once & 0.55  & 0 & Once & 0.54  & 0 & The & 0.43  & 0 & Q \\*
    17 & 0.11  & 3 & time & 0.04  & 1024 & in & 0.07  & 16 & A & 0.08  & 701 & : \\*
    17 & 0.04  & 4 & there & 0.03  & 1025 & her & 0.07  & 36 & .$\backslash$n & 0.03  & 700 & A \\*
    17 & 0.03  & 1 & upon & 0.02  & 1026 & room & 0.03  & 15 & ?$\backslash$n & 0.02  & 699 & ?$\backslash$n \\*
    17 & 0.01  & 2 & a & 0.02  & 1023 & was & 0.02  & 17 & : & 0.01  & 666 & . \\*
    \midrule
    18 & 0.86  & 0 & Once & 0.66  & 0 & Once & 0.65  & 0 & The & 0.55  & 0 & Q \\*
    18 & 0.05  & 3 & time & 0.02  & 1024 & in & 0.07  & 16 & A & 0.06  & 701 & : \\*
    18 & 0.04  & 1 & upon & 0.02  & 1023 & was & 0.04  & 36 & .$\backslash$n & 0.03  & 655 & Axel \\*
    18 & 0.03  & 4 & there & 0.02  & 1025 & her & 0.03  & 17 & : & 0.03  & 680 & he \\*
    18 & 0.01  & 2 & a & 0.02  & 1022 & girl & 0.02  & 15 & ?$\backslash$n & 0.03  & 696 & they \\*
    \midrule
    19 & 0.90  & 0 & Once & 0.69  & 0 & Once & 0.62  & 0 & The & 0.59  & 0 & Q \\*
    19 & 0.04  & 4 & there & 0.02  & 1026 & room & 0.07  & 16 & A & 0.04  & 701 & : \\*
    19 & 0.04  & 3 & time & 0.01  & 1024 & in & 0.06  & 7 & : & 0.03  & 655 & Axel \\*
    19 & 0.02  & 1 & upon & 0.01  & 1025 & her & 0.05  & 36 & .$\backslash$n & 0.02  & 699 & ?$\backslash$n \\*
    19 & 0.01  & 2 & a & 0.01  & 1019 & . & 0.02  & 15 & ?$\backslash$n & 0.01  & 700 & A \\*
    \midrule
    20 & 0.89  & 0 & Once & 0.61  & 0 & Once & 0.61  & 0 & The & 0.59  & 0 & Q \\*
    20 & 0.05  & 3 & time & 0.02  & 1024 & in & 0.07  & 16 & A & 0.06  & 655 & Axel \\*
    20 & 0.03  & 4 & there & 0.01  & 1026 & room & 0.06  & 36 & .$\backslash$n & 0.04  & 701 & : \\*
    20 & 0.02  & 1 & upon & 0.01  & 1023 & was & 0.05  & 7 & : & 0.01  & 700 & A \\*
    20 & 0.01  & 2 & a & 0.01  & 1019 & . & 0.03  & 15 & ?$\backslash$n & 0.01  & 699 & ?$\backslash$n \\*
    \midrule
    21 & 0.87  & 0 & Once & 0.68  & 0 & Once & 0.66  & 0 & The & 0.63  & 0 & Q \\*
    21 & 0.06  & 3 & time & 0.04  & 1025 & her & 0.08  & 36 & .$\backslash$n & 0.08  & 701 & : \\*
    21 & 0.05  & 4 & there & 0.03  & 1026 & room & 0.05  & 16 & A & 0.02  & 698 & together \\*
    21 & 0.01  & 1 & upon & 0.02  & 1024 & in & 0.04  & 15 & ?$\backslash$n & 0.02  & 699 & ?$\backslash$n \\*
    21 & 0.01  & 2 & a & 0.01  & 1019 & . & 0.02  & 35 & cream & 0.02  & 700 & A \\*
    \midrule
    22 & 0.89  & 0 & Once & 0.67  & 0 & Once & 0.66  & 0 & The & 0.63  & 0 & Q \\*
    22 & 0.04  & 3 & time & 0.02  & 1026 & room & 0.08  & 36 & .$\backslash$n & 0.05  & 701 & : \\*
    22 & 0.04  & 4 & there & 0.02  & 1025 & her & 0.07  & 16 & A & 0.03  & 655 & Axel \\*
    22 & 0.03  & 1 & upon & 0.01  & 1024 & in & 0.02  & 15 & ?$\backslash$n & 0.02  & 699 & ?$\backslash$n \\*
    22 & 0.01  & 2 & a & 0.01  & 1019 & . & 0.02  & 5 & $\backslash$n & 0.01  & 177 & Originally \\*
    \midrule
    23 & 0.93  & 0 & Once & 0.77  & 0 & Once & 0.75  & 0 & The & 0.71  & 0 & Q \\*
    23 & 0.04  & 3 & time & 0.02  & 1024 & in & 0.06  & 16 & A & 0.05  & 655 & Axel \\*
    23 & 0.01  & 4 & there & 0.02  & 1026 & room & 0.05  & 36 & .$\backslash$n & 0.03  & 701 & : \\*
    23 & 0.01  & 1 & upon & 0.01  & 1025 & her & 0.02  & 15 & ?$\backslash$n & 0.01  & 700 & A \\*
    23 & 0.00  & 2 & a & 0.01  & 1023 & was & 0.01  & 7 & : & 0.01  & 660 & pesos \\*
    \midrule
    24 & 0.94  & 0 & Once & 0.72  & 0 & Once & 0.76  & 0 & The & 0.73  & 0 & Q \\*
    24 & 0.03  & 4 & there & 0.02  & 1026 & room & 0.05  & 16 & A & 0.08  & 655 & Axel \\*
    24 & 0.02  & 3 & time & 0.02  & 1019 & . & 0.05  & 36 & .$\backslash$n & 0.03  & 701 & : \\*
    24 & 0.01  & 1 & upon & 0.01  & 1025 & her & 0.02  & 7 & : & 0.01  & 660 & pesos \\*
    24 & 0.01  & 2 & a & 0.01  & 1024 & in & 0.02  & 5 & $\backslash$n & 0.01  & 699 & ?$\backslash$n \\*
    \midrule
    25 & 0.92  & 0 & Once & 0.78  & 0 & Once & 0.82  & 0 & The & 0.79  & 0 & Q \\*
    25 & 0.05  & 3 & time & 0.04  & 1025 & her & 0.05  & 36 & .$\backslash$n & 0.04  & 701 & : \\*
    25 & 0.02  & 4 & there & 0.02  & 1026 & room & 0.04  & 16 & A & 0.04  & 655 & Axel \\*
    25 & 0.01  & 1 & upon & 0.01  & 1024 & in & 0.02  & 15 & ?$\backslash$n & 0.01  & 699 & ?$\backslash$n \\*
    25 & 0.00  & 2 & a & 0.01  & 1019 & . & 0.01  & 5 & $\backslash$n & 0.01  & 700 & A \\*
    \midrule
    26 & 0.90  & 0 & Once & 0.63  & 0 & Once & 0.67  & 0 & The & 0.59  & 0 & Q \\*
    26 & 0.04  & 3 & time & 0.02  & 1026 & room & 0.12  & 36 & .$\backslash$n & 0.10  & 655 & Axel \\*
    26 & 0.04  & 4 & there & 0.01  & 1019 & . & 0.05  & 16 & A & 0.07  & 701 & : \\*
    26 & 0.02  & 1 & upon & 0.01  & 1023 & was & 0.03  & 15 & ?$\backslash$n & 0.02  & 700 & A \\*
    26 & 0.01  & 2 & a & 0.01  & 1025 & her & 0.02  & 5 & $\backslash$n & 0.02  & 699 & ?$\backslash$n \\*
    \midrule
    27 & 0.91  & 0 & Once & 0.74  & 0 & Once & 0.80  & 0 & The & 0.64  & 0 & Q \\*
    27 & 0.03  & 4 & there & 0.02  & 1026 & room & 0.05  & 36 & .$\backslash$n & 0.08  & 655 & Axel \\*
    27 & 0.03  & 3 & time & 0.01  & 1025 & her & 0.03  & 16 & A & 0.04  & 701 & : \\*
    27 & 0.02  & 1 & upon & 0.01  & 724 & the & 0.02  & 35 & cream & 0.01  & 700 & A \\*
    27 & 0.01  & 2 & a & 0.01  & 1019 & . & 0.01  & 15 & ?$\backslash$n & 0.01  & 699 & ?$\backslash$n \\*
    \midrule
    28 & 0.84  & 0 & Once & 0.64  & 0 & Once & 0.68  & 0 & The & 0.64  & 0 & Q \\*
    28 & 0.07  & 4 & there & 0.05  & 1026 & room & 0.11  & 36 & .$\backslash$n & 0.10  & 701 & : \\*
    28 & 0.06  & 3 & time & 0.03  & 1025 & her & 0.03  & 16 & A & 0.03  & 700 & A \\*
    28 & 0.02  & 1 & upon & 0.01  & 1024 & in & 0.03  & 15 & ?$\backslash$n & 0.03  & 655 & Axel \\*
    28 & 0.01  & 2 & a & 0.00  & 1023 & was & 0.02  & 35 & cream & 0.02  & 699 & ?$\backslash$n \\*
    \midrule
    29 & 0.88  & 0 & Once & 0.67  & 0 & Once & 0.75  & 0 & The & 0.76  & 0 & Q \\*
    29 & 0.06  & 4 & there & 0.04  & 1026 & room & 0.06  & 36 & .$\backslash$n & 0.05  & 701 & : \\*
    29 & 0.05  & 3 & time & 0.02  & 1025 & her & 0.04  & 16 & A & 0.01  & 700 & A \\*
    29 & 0.01  & 1 & upon & 0.01  & 1019 & . & 0.03  & 15 & ?$\backslash$n & 0.01  & 699 & ?$\backslash$n \\*
    29 & 0.01  & 2 & a & 0.00  & 1018 & room & 0.01  & 5 & $\backslash$n & 0.01  & 695 & pesos \\*
    \midrule
    30 & 0.85  & 0 & Once & 0.49  & 0 & Once & 0.60  & 0 & The & 0.52  & 0 & Q \\*
    30 & 0.08  & 4 & there & 0.05  & 1026 & room & 0.11  & 36 & .$\backslash$n & 0.09  & 701 & : \\*
    30 & 0.04  & 3 & time & 0.02  & 1019 & . & 0.04  & 16 & A & 0.04  & 655 & Axel \\*
    30 & 0.03  & 1 & upon & 0.01  & 1025 & her & 0.03  & 15 & ?$\backslash$n & 0.02  & 700 & A \\*
    30 & 0.01  & 2 & a & 0.01  & 1024 & in & 0.02  & 6 & Q & 0.02  & 699 & ?$\backslash$n \\*
    \midrule
    31 & 0.78  & 0 & Once & 0.49  & 0 & Once & 0.61  & 0 & The & 0.51  & 0 & Q \\*
    31 & 0.16  & 4 & there & 0.16  & 1026 & room & 0.21  & 36 & .$\backslash$n & 0.22  & 701 & : \\*
    31 & 0.04  & 3 & time & 0.02  & 1025 & her & 0.02  & 15 & ?$\backslash$n & 0.02  & 699 & ?$\backslash$n \\*
    31 & 0.01  & 1 & upon & 0.02  & 1019 & . & 0.02  & 16 & A & 0.01  & 700 & A \\*
    31 & 0.01  & 2 & a & 0.01  & 1024 & in & 0.01  & 17 & : & 0.01  & 655 & Axe \\
\end{xltabular}
}

\iftaclpubformat

\fi

\end{document}